\documentclass[journal]{IEEEtran}

\usepackage{amsmath,amsfonts}
\usepackage{algorithmic}
\usepackage{algorithm}
\usepackage{array}
\usepackage{textcomp}
\usepackage{stfloats}
\usepackage{url}
\usepackage{verbatim}
\usepackage{graphicx}
\usepackage{soul}
\usepackage[table,dvipsnames]{xcolor}
\definecolor{lightyellow}{rgb}{1, 1, 0.93}
\usepackage[export]{adjustbox}

\IEEEoverridecommandlockouts
\DeclareTextFontCommand{\textbfit}{%
  \fontseries\bfdefault % change series without selecting the font yet
  \itshape
}
\usepackage{makecell}
\usepackage{multirow}
\usepackage{booktabs}% http://ctan.org/pkg/booktabs
\newcommand{\tabitem}{~~\llap{\textbullet}~~}
\usepackage{flushend}
\usepackage{cellspace}
\newcommand\cincludegraphics[2][]{\raisebox{-0.5\height}{\includegraphics[#1]{#2}}}
\newcommand{\STAB}[1]{\begin{tabular}{@{}c@{}}#1\end{tabular}}

\usepackage{doi}
\usepackage{hyperref}
\hypersetup{
    colorlinks=false,
    pdfborder={0 0 0},
}
\usepackage[numbers]{natbib}
\usepackage{subfigure}

\usepackage{relsize}
\newcommand{\subtitlerelsize}{4} %relative size: integer value
\newcommand{\subtitlelinesep}{0.1em} %line separation: a LaTeX length
\usepackage{titling}
\pretitle{\begin{flushleft}\Huge}
\posttitle{\par\end{flushleft}}
\preauthor{\begin{flushleft}\Large}
\postauthor{\end{flushleft}}
\predate{\begin{flushleft}}
\postdate{\end{flushleft}}
\usepackage{etoolbox}
\makeatletter
\patchcmd{\maketitle}
 {\def\@makefnmark}
 {\def\@makefnmark{}\def\useless@macro}
 {}{}
\makeatother

\usepackage{enumitem}
\newenvironment{phv}{\fontfamily{phv}\selectfont}{\par}

\begin{document}

\title{Unlocking the Emotional World of Visual Media:\break An Overview of the Science, Research, and Impact of  Understanding Emotion\\[\subtitlelinesep]%
    \smaller[\subtitlerelsize]{}\color{MidnightBlue}{\it Drawing insights from psychology, engineering, and the arts, this article provides a comprehensive overview of the field of emotion analysis in visual media, and discusses the latest research, systems, challenges, ethical implications, and potential impact of artificial emotional intelligence on society.}}

\author{By James Z. Wang,~\IEEEmembership{Senior Member,~IEEE,} Sicheng Zhao,~\IEEEmembership{Senior Member,~IEEE,} Chenyan Wu,
Reginald~B.~Adams,~Jr., Michelle~G. Newman, Tal Shafir, Rachelle Tsachor
        % <-this % stops a space
\thanks{Manuscript received 31 August 2022; revised 7 March 2023 and 15 April 2023; accepted 3 May 2023. 
The work of James Z. Wang and Chenyan Wu was supported in part by generous gifts from the
Amazon Research Awards Program. The work of James Z. Wang, Reginald B. Adams Jr., and Michelle G. Newman was supported in part by the National Science Foundation (NSF) under Grant IIS-1110970 and Grant CNS-1921783. The work of James Z. Wang, Reginald B. Adams Jr., Michelle G. Newman, Tal Shafir, and Rachelle Tsachor was supported in part by the NSF under Grant CNS-2234195 and Grant CNS-2234197. The work of James Z. Wang and Michelle G. Newman was supported in part by the NSF under Grant CIF-2205004. {\it (Corresponding author: James Z. Wang.)}\\
{\bf James Z. Wang} and {\bf Chenyan Wu} are with the Data Science and Artificial Intelligence Area and the Human-Computer Interaction Area of the College of Information Sciences and Technology, The Pennsylvania State University, University Park, PA 16802, USA (e-mail: jwang@psu.edu; czw390@psu.edu)\\
{\bf Reginald B. Adams, Jr.} and {\bf Michelle G. Newman} are with the Department of Psychology, The Pennsylvania State University, University Park, PA 16802, USA (e-mail: rba10@psu.edu; mgn1@psu.edu)\\
{\bf Sicheng Zhao} is with BNRist, Tsinghua University, Beijing 100084, China. (e-mail: schzhao@tsinghua.edu.cn)\\
{\bf Tal Shafir} is with the Emily Sagol Creative Arts
Therapies Research Center, University of Haifa, 3498838, Israel (e-mail: tshafir1@univ.haifa.ac.il)\\
{\bf Rachelle Tsachor} is with the School of Theatre and Music, University of Illinois, Chicago, IL 60607, USA (e-mail: rtsachor@uic.edu)}}%% <-this % stops a space

\markboth{Proceedings of the IEEE,~Vol.~XX, No.~X, February~2023}%
{Wang \MakeLowercase{\textit{et al.}}: Emotion in Images and Videos:\break Principles, Ideas, and Challenges}

\IEEEpubid{0000--0000/00\$00.00~\copyright~2023 IEEE}

\maketitle

\begin{abstract}
The emergence of artificial emotional intelligence technology is revolutionizing the fields of computers and robotics, allowing for a new level of communication and understanding of human behavior that was once thought impossible. Whereas recent advancements in deep learning have transformed the field of computer vision, automated understanding of evoked or expressed emotions in visual media remains in its infancy. This foundering stems from the absence of a universally accepted definition of ``emotion,'' coupled with the inherently subjective nature of emotions and their intricate nuances. In this paper, we provide a comprehensive, multidisciplinary overview of the field of emotion analysis in visual media, drawing on insights from psychology, engineering, and the arts. We begin by exploring the psychological foundations of emotion and the computational principles that underpin the understanding of emotions from images and videos. We then review the latest research and systems within the field, accentuating the most promising approaches. We also discuss the current technological challenges and limitations of emotion analysis, underscoring the necessity for continued investigation and innovation. We contend that this represents a ``Holy Grail'' research problem in computing and delineate pivotal directions for future inquiry. Finally, we examine the ethical ramifications of emotion-understanding technologies and contemplate their potential societal impacts. Overall, this article endeavors to equip readers with a deeper understanding of the domain of emotion analysis in visual media and to inspire further research and development in this captivating and rapidly evolving field.
\end{abstract}


\section{Introduction}\label{sec:intro}
% guide: https://www.scribbr.com/research-paper/research-paper-introduction/

As Artificial Intelligence (AI) technology becomes more prevalent and capable of performing a wide range of tasks, the need for effective communication between humans and AI systems is becoming increasingly important. 
The adoption of smart home products and services is projected to reach 400 million worldwide, with smart devices such as Alexa and Astro becoming increasingly common in households~\cite{url_stat_smart}. However, these devices are currently limited to executing specific commands and do not possess the capability to understand or respond to human emotions~\cite{krakovsky2018artificial}. This lack of emotional intelligence (EQ) limits their potential applications, and this constraint is particularly relevant for future robotic applications, such as personal assistant robots, social robots, service robots, factory/warehouse robots, and police robots, which require close collaboration and a comprehensive understanding of human behavior and emotions.

The ability to impart EQ to AI when dealing with visual information is a topic of growing interest. This article aims to address the fundamental question of {\it how to ``teach'' AI to understand and respond to human emotions based on images and videos.} The potential technical solutions to these questions have far-reaching implications for various application domains, including human-AI interaction, autonomous driving, social media, entertainment, information management and retrieval, design, industrial safety, and education.

To provide a comprehensive and well-balanced view of this complex subject, it is essential to draw on the expertise of various fields, including computer and information science and engineering, psychology, data science, movement analysis, and performing arts. The interdisciplinary nature of this topic highlights the need for collaboration and cooperation among researchers from different fields in order to achieve a deeper understanding of the subject.

In this article, we focus on the topic of affective visual information analysis as it represents a highly nuanced and complex area of study with strong connections to well-established scholarly fields such as computer vision, multimedia, and image and video processing. However, it is important to note that the techniques presented here can be integrated with other data modalities, such as speech, sensor-generated streaming data, and text, in order to enhance the performance of real-world applications.

The primary objective of this article is to introduce the technical communities to the emerging field of affective visual information analysis. Recognizing the breadth and dynamic nature of this field, we do not aim to provide a comprehensive survey of all sub-areas. Instead, our discussion focuses on the fundamental psychological and computational principles (Sections~\ref{sec:psych} and~\ref{sec:compu}), recent advancements and developments (Section~\ref{sec:survey}), core challenges and open issues (Section~\ref{sec:challenge}), connections to other areas of research and development (Section~\ref{sec:connect}), and ethical considerations related to this new technology (Section~\ref{sec:ethics}). We apologize in advance for any important publications that may have been omitted in our discussion.

{Recently, there have been some other surveys and reviews on artificial emotional intelligence, such as facial expression recognition~\cite{hassan2021automatic,li2020deep,jampour2022multiview,liu2022graph}, microexpression recognition~\cite{ben2022video,li2018towards,li2022deep}, textual sentiment classification~\cite{brauwers2022survey, nazir2022issues,deng2021survey}, music and speech emotion recognition~\cite{akccay2020speech,panda2020audio,latif2021survey}, affective image content analysis~\cite{zhao2021affective}, emotional body gesture recognition~\cite{noroozi2021survey}, bodily expressed emotion recognition~\cite{mahfoudi2022emotion}, emotion recognition from physiological signals~\cite{li2022eeg,saganowski2022emotion}, multimodal emotion recognition~\cite{zhang2020emotion,zhao2021emotion}, and affective theory use~\cite{smith2022lies}. These articles mainly focus on emotion and sentiment analysis for a specific modality from the perspective of machine learning and pattern recognition or focus on the psychological emotion theories. \citeauthor{cambria2016affective}~\cite{cambria2016affective} summarized the common tasks of affective computing and sentiment analysis and classified existing methods into three main categories: knowledge-based, statistical, and hybrid approaches. \citeauthor{ poria2017review}~\cite{ poria2017review} and \citeauthor{ wang2022systematic}~\cite{ wang2022systematic} reviewed both unimodal and multimodal emotion recognition before the year of 2017 and between the years of 2017 and 2020, respectively. As opposed to those reviews, the current paper aims to provide a comprehensive overview of emotion analysis from visual media (e.g., both images and videos) with insights drawn from multiple disciplines.}

\section{Emotion: The Psychological Foundation}\label{sec:psych}

How we define emotion largely descends from the theoretical framework used to study it. In this section, we provide an overview of the most prominent emotion theories, beginning with Darwin, and underscore how contemporary dimensional approaches to understanding emotion align with both the processing of emotion by the human brain and current computer vision approaches for  modeling emotion to make predictions about human perception (Section~\ref{sec:define}). In addition, we examine the intrinsic link between emotion and adaptive behavior, a contention that is largely shared across different emotion theories (Section~\ref{sec:behavior}).  

\subsection{Definitions and Models of Emotion}\label{sec:define}

\begin{figure*}[ht!]
\setlength{\tabcolsep}{5pt}
    \centering
    \begin{tabular}{ccc}
      \includegraphics[trim=20 210 620 85,clip,height=1.5in]{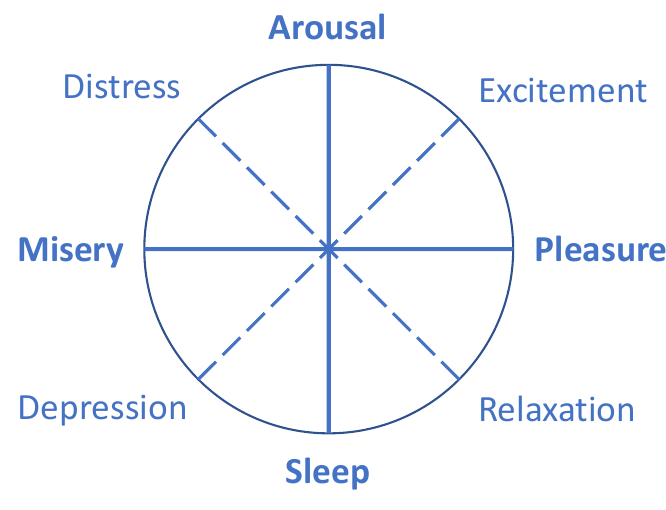}   &  
      \includegraphics[trim=15 210 530 85,clip,height=1.5in]{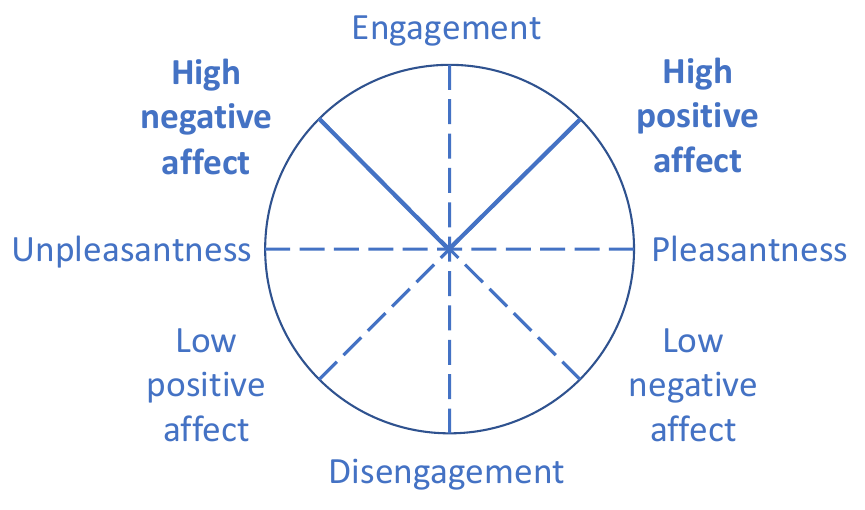}   & 
      \includegraphics[trim=15 210 635 85,clip,height=1.5in]{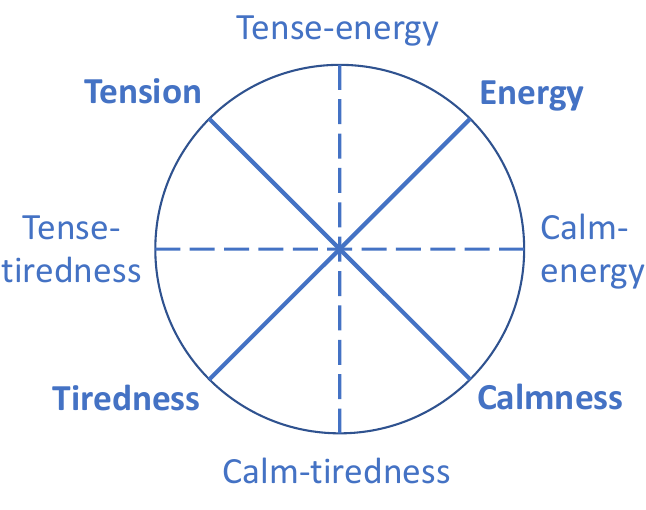} \\
       (a) Russell, 1980 & (b) Watson and Tellegen, 1985 & (c) Thayer, 1989
       \end{tabular}
       \vskip 0.1in
           \begin{tabular}{cc}
           \includegraphics[trim=45 210 560 85,clip,height=1.5in]{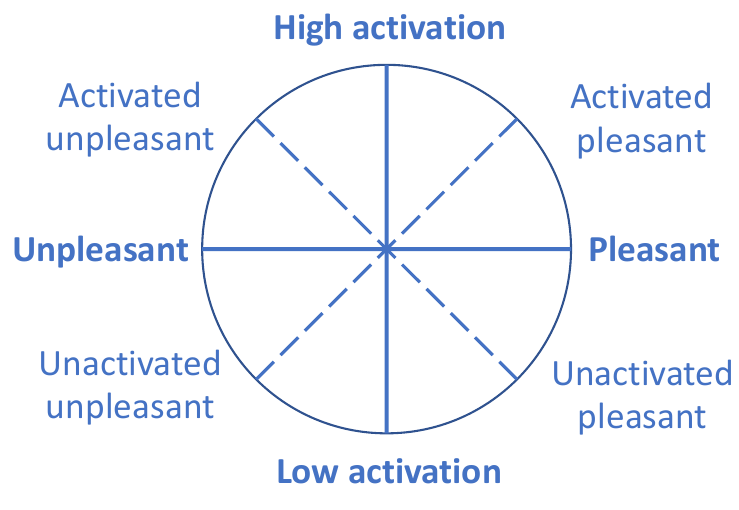}   &       \includegraphics[trim=15 210 600 85,clip,height=1.5in]{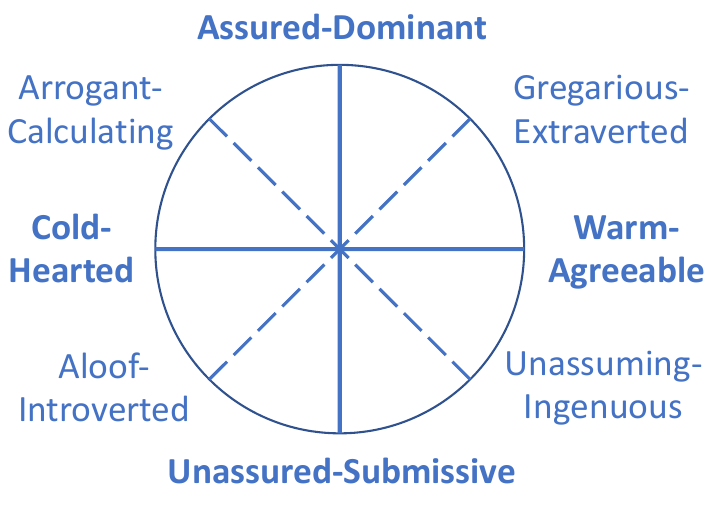}   \\
                (d) Larsen and Diener, 1992 & (e) Knutson, 1996
           \end{tabular}
    \caption{Five influential circumplex emotion models developed in the psychology literature~\cite{mehrabian1980basic,watson1985toward,thayer1989biopsychology,larsen1992promises,knutson1996facial}.}
    \label{fig:models}
\end{figure*}

One of the first emotion theories put forth was Charles Darwin's in his seminal book, ``On the Expression of the Emotions in Man and Animals''~\cite{darwin1998expression}. This book proposed that humans possess a finite set of biologically privileged emotions that evolved to confer upon us survival-related behavior. William James~\cite{james2013emotion} later added to this arguing that the experience of emotion is ultimately our experience of distinct patterns of physiological arousal and physical behaviors associated with each emotion~\cite{niedenthal2007embodying}. Building upon these assumptions, Ekman's Neurocultural Theory of Emotion~\cite{ekman1971universals} further perpetuated the notion that there exists a ``universal affect program'' that underlies the experience and expression of several discrete emotions, such as anger, fear, sadness, and happiness. According to this theory, basic emotional experiences and emotional displays evolved as adaptive responses to specific environmental contingencies, and thus felt, expressed, and recognized emotions are uniform across all people and cultures, and are marked by specific patterns of physiological and neural responsivity. 
 
Considerable research has since questioned the utility of this approach. This includes findings that people: 1) are often ill-equipped to describe their own emotions in discrete emotion terms, both in research and clinical settings~\cite{barrett2006solving}, 2) show low consensus in their ability to categorize both facial and vocal expressions of emotions in discrete emotion terms~\cite{posner2005circumplex}, and 3) show high intercorrelations across the emotional experiences they do report~\cite{russell1999bipolarity}. Such findings have prompted many researchers to explore alternative approaches to conceptualizing and measuring emotional experience that necessarily involve a cognitive component.
 
Magda Arnold's cognitive appraisal theory of emotion was the first to introduce the necessity of cognition in emotion elicitation~\cite{arnold1960emotion}. Although she did not disagree with Darwin and James that emotions are adaptive states spurring on survival-related behavior, she nonetheless took them to task for not considering vast individual variation in emotional experiences. Arnold rightly underscored the capacity for the same emotion-evoking events to lead to different emotional experiences in different people. This becomes readily apparent when considering one's own emotional experiences. For example, diving off a cliff may generate an aversive fear state in one person but an enjoyable thrill state in another. The difference in how an event is evaluated, therefore, shapes the emotion that results. Central to her theory was the importance of cognitive appraisal in initially eliciting an emotion. Once elicited, she largely agreed with Darwin's functional assumptions regarding the survival benefits of emotion-related behavior. Later, Schachter and Singer~\cite{schachter1962cognitive} drew on these insights to help resolve ongoing debates regarding James' theory of embodied emotional experience. Their research demonstrated that the emotion we experience when adrenaline is released in our body depends on our cognitive framing and context. Those injected with adrenaline reported feeling happier when in a fun context and more irritated when in an angering context. The only difference was the cognitive appraisal that framed the experience of that arousal. 

Building upon these ideas further, Mehrabian and Russell~\cite{mehrabian1974approach} proposed the  Pleasure, Arousal, and Dominance (PAD) Emotional State Model, which suggests a dimensional account of emotion, one in which PAD constitutes the fundamental dimensions that represent all emotions. Later, Russell dropped dominance as a key dimension and focused on what he refers to as ``core affect,'' suggesting that all emotions can be reduced to the fundamental psychological and biological dimensions of pleasantness and arousal~\cite{russell2003core}. Dimensional approaches offer a way of conceptualizing and assessing emotion that closely approximates how the human brain processes emotion (see Fig.~\ref{fig:models} for a comparison of different dimensional/circumplex models).

Notably, some early attempts to use computational methods to predict human emotions elicited by visual scenes employed the discrete emotion approach described at the outset of this section. For example, \citeauthor{mikels2005emotional}~\cite{mikels2005emotional} and \citeauthor{machajdik2010affective}~\cite{machajdik2010affective} used categorical approaches to assess the visual properties of stimuli taken from the International Affective Picture System (IAPS)~\cite{lang2005international}, a widely used set of emotionally evocative photographs in the emotion literature. However, such approaches resulted in high levels of multicollinearity between emotions, making it difficult to disentangle emotions using traditional regression models. In contrast, adopting a dimensional approach not only aligns well with emerging theoretical accounts of emotion, but has been validated by James Wang and his colleagues in the successful assessment of human aesthetics and emotions evoked by visual scenes, as well as bodily expressed emotion~\cite{lu2012shape,lu2016identifying,lu_adams_li_newman_wang_2017,ye2017probabilistic,kim2018development,luo2020arbee,datta2006studying}. This offers a methodological approach that is consistent with dimensional theories of emotion.  
\subsection{The Interplay Between Emotion and Behavior}\label{sec:behavior}

Fridlund's behavioral ecology perspective of emotion argues that emotional expression evolved primarily as a means of signaling behavioral intent~\cite{fridlund2014human}. The view that facial expression evolved specifically as a way to forecast behavioral intentions and consequences to others drew from Darwin's seminal writings on expression~\cite{darwin1998expression}, even though Darwin himself argued that expressions did not evolve for social communication per se. Fridlund's argument is based on the idea that perceiving behavioral intentions is adaptive. From this perspective, anger may primarily convey to an observer a readiness to attack, whereas fear may primarily convey a readiness to submit or retreat (see~\cite{yik1999interpretation}). From this perspective behavioral intentions are considered ``syndromes of correlated components''~\cite{fridlund2014human} (p. 151). Fridlund is not alone in these assumptions. Some researchers have gone so far as to suggest that feeling states associated with emotions are merely conscious perceptions of underlying behavioral intentions, or action tendencies, which implies that emotional feeling is simply the experience of behavioral intention, similar to William James's theory~\cite{frijda1997facial}.

It is worth noting that empirical research has provided support for the idea that behavioral intention is conveyed through emotional expression. For example, one study demonstrated that action tendencies and emotion labels are attributed to faces at comparable levels of consistency~\cite{frijda1997facial}. Similarly, in forced-choice paradigms~\cite{yik1999interpretation}, cross-cultural evidence indicates that participants assign behavioral intention descriptors with about equal consistency as they do with emotion descriptors.

A focus on approach-avoidance tendencies has been highlighted in most of the research conducted to date. The ability to detect another's intention to approach or avoid us is thought of as a principal factor governing social exchange. However, much of the work on approach-avoidance behavioral motivations has also tended to concentrate on the experience or response of an observer to a stimulus event~\cite{davidson1992emotion}. One common method of operationalizing approach and avoidance then stems from traditional behavioral learning paradigms that link behavioral motivation and emotion through reward versus punishment contingencies~\cite{miller1937analysis}. Approach motivation is defined by appetitive, reward-related behavior, while avoidance motivation is defined by aversive, punishment-related behavior, where appetitive behavior is movement toward a reward and aversive behavior is movement away from a punishment.  

Much research has focused on the relationship between approach and avoidance tendencies and emotional experience~\cite{harmon2003clarifying}. However, there has been less attention paid to whether approach and avoidance tendencies are fundamentally signaled by the external expression of emotion. It stands to reason that if the experience of emotion is associated with approach and avoidance tendencies, these tendencies should be signaled to others when expressed. This distinction is important as the approach-avoidance tendencies attributed to expressive faces may not always match the approach-avoidance reactions elicited by them. For example, the expression of joy arguably conveys a heightened likelihood of approach by the expressor and a reaction of approach from the observer~\cite{davidson1992emotion}. In contrast, anger expressions signal approach by the expressor, but tend to elicit avoidance by the observer.  

Recent insights from the embodiment literature also provide evidence that emotional experiences are grounded in specific action tendencies~\cite{niedenthal2007embodying}. This means that emotional experiences can be expressed through stored action tendencies in the body, rather than through semantic cues.  For example, studies have examined the coherence between emotional experience (positive or negative) and approach (arm flexion, i.e., pulling toward) versus avoidance behavior (arm extension, i.e., pushing away)~\cite{centerbar2008affective}. In one study, participants were randomly assigned to an arm flexion (approach behavior) or arm extension (avoidance behavior) condition, either during the reading of a story about a fictional character or during a positive versus negative semantic priming task before reading the story. Participants in the congruent conditions (happy prime and arm flexion, and sad prime and arm extension) were able to remember more items from the story.

Although important for explaining behavioral responding, these studies did not address whether basic tendencies to approach or avoid were also fundamentally signaled by emotional expressions. If they were, expressions coupled with approach and avoidant behaviors should impact the efficiency of emotion recognition. In one set of studies, anger expressions were found to facilitate judgments of approach versus withdrawing faces compared with fear expressions~\cite{adams2006emotional}. Similarly, perceived movement of a face toward or away from an observer likewise facilitated angry or fearful expression perception~\cite{nelson2013approach}. Thus, approach and avoidance movement are associated in a fundamental way with the recognition of anger and fear displays, respectively, supporting the conclusion that basic action tendencies are inherently associated with the perception of emotion.
 
In sum, despite widely debatable assumptions about the nature of emotion and emotional expression across various theories, most tend to agree that emotion expression conveys fundamental information regarding basic behavioral tendencies~\cite{adams2006emotional}. 

\section{Emotion: Computational Principles and Foundations}\label{sec:compu}

In this section, we aim to establish computational foundations for analyzing and recognizing emotions from visual media. Emotion recognition systems typically involve several fundamental data-related stages, including data collection (Section~\ref{sec:data}), data reliability assessment (Section~\ref{sec:reliability}), and data representation (Sections~\ref{sec:representation1},~\ref{sec:representation2}, and~\ref{sec:context} for general computer vision-based representation, movement coding, and context and function purpose detection, respectively). As we present specific examples at each stage, we will emphasize the underlying principles they adhere to.
We provide a list of representative datasets in Section~\ref{sec:datasets}. We will also introduce the factors of acted portrayals (Section~\ref{sec:acted}), cultural and gender dialects (Section~\ref{sec:culture}), structure (Section~\ref{sec:structure}), personality (Section~\ref{sec:personality}), and affective style (Section~\ref{sec:style}) in inferring emotion, based on prior research. 

\subsection{Data Collection}\label{sec:data}

Because the categories of emotions are not well-defined, it is not possible to program a computer to recognize all emotion categories based on a set of predefined logic rules, computational instructions, or procedures. Thus, researchers must take a data-driven approach in which computers learn from a large quantity of labeled, partially labeled, and/or unlabeled examples. To enable such research and subsequent real-world applications, it is essential to collect large-scale, high-quality, ecologically valid datasets. To highlight the complexity of the data collection problem as well as to introduce best practices, we describe a few data collection approaches that incorporate psychological principles in their design.

\begin{figure*}[ht!]
\begin{phv}
\centering\begin{tabular}{Sc | Sc | Sc | Sc}
    &  {Valence} & {Arousal} & {Likability}\\
    \hline
{\STAB{\rotatebox[origin=c]{90}{High Affect}}} &
    \cincludegraphics[width=1in]{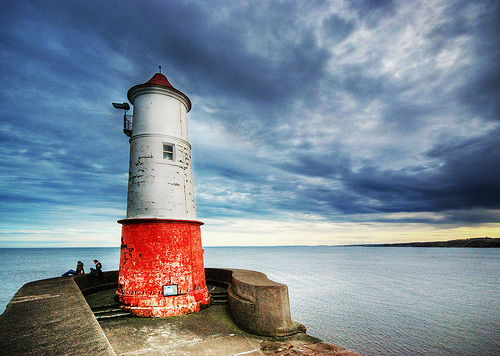} 
    \cincludegraphics[width=0.9in]{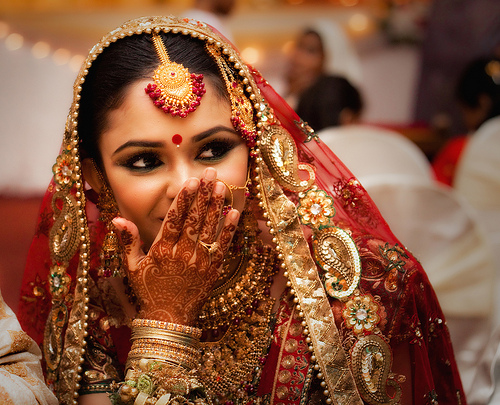}
    & 
    \cincludegraphics[width=0.8in]{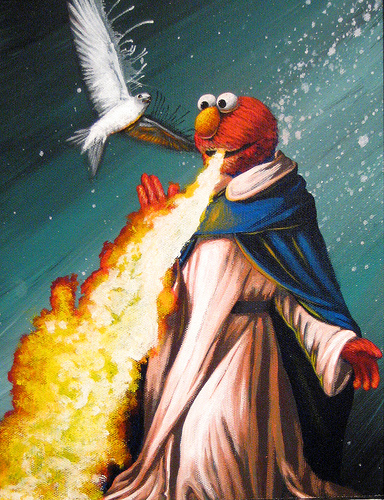}
    \cincludegraphics[width=1in]{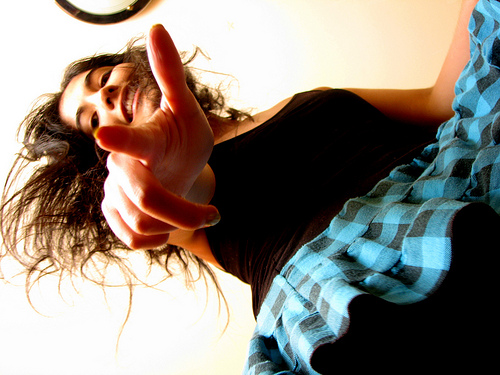}
    & 
    \cincludegraphics[width=0.9in]{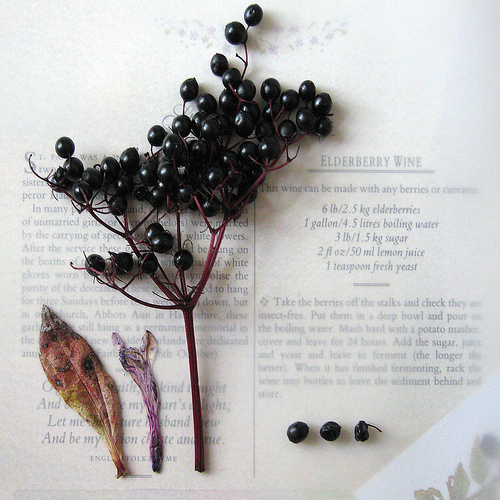}
    \cincludegraphics[width=1in]{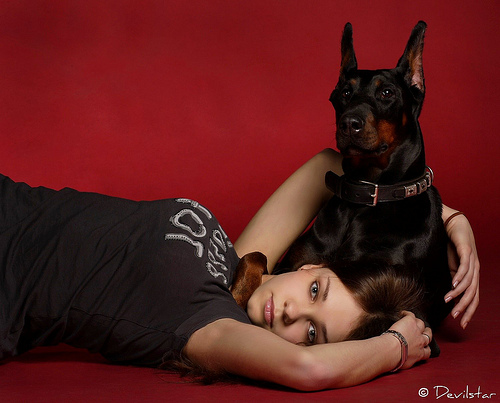}
\\ \hline
{\STAB{\rotatebox[origin=c]{90}{Neutral Affect}}} & \cincludegraphics[width=0.8in]{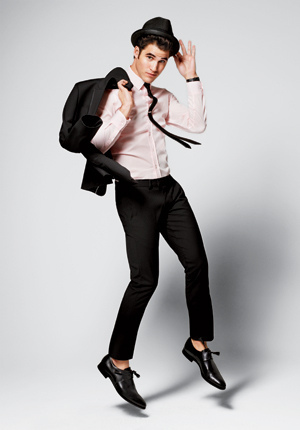}
    \cincludegraphics[width=1.1in]{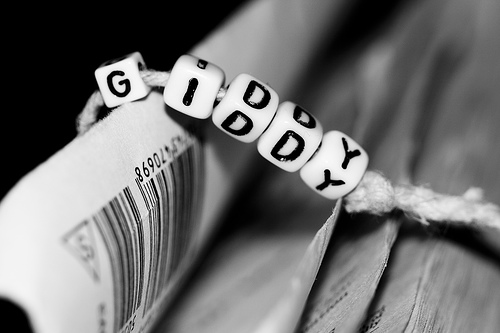}
    &
 \fcolorbox{gray}{white}{\cincludegraphics[width=0.7in]{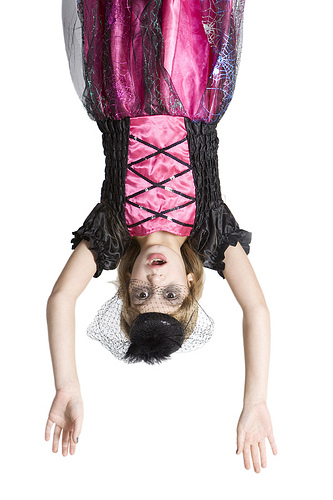}}
    \cincludegraphics[width=1in]{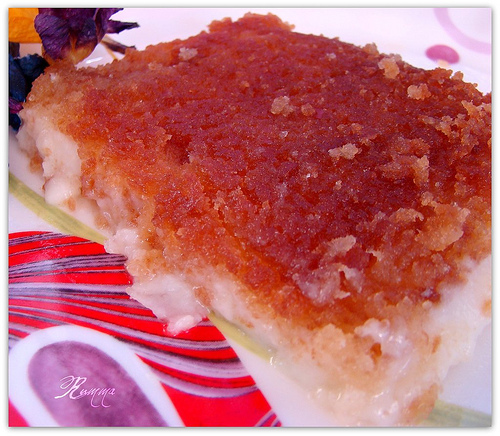}
    &
    \cincludegraphics[width=0.8in]{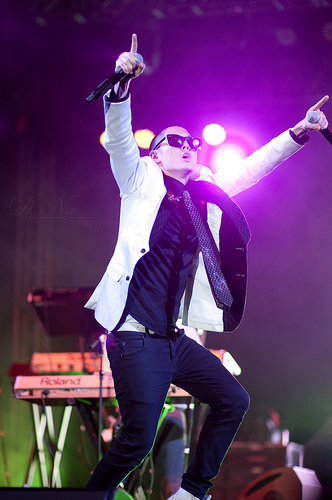}
    \cincludegraphics[width=1.1in]{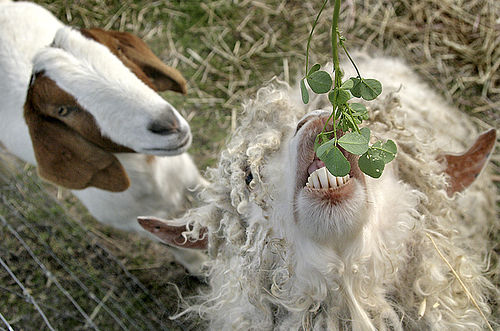}

\\ \hline
{\STAB{\rotatebox[origin=c]{90}{Low Affect}}} & \cincludegraphics[width=1in]{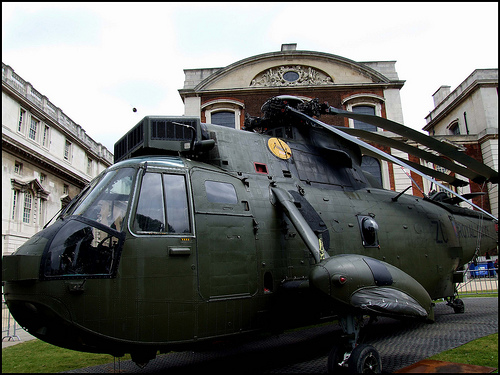}
    \cincludegraphics[width=0.9in]{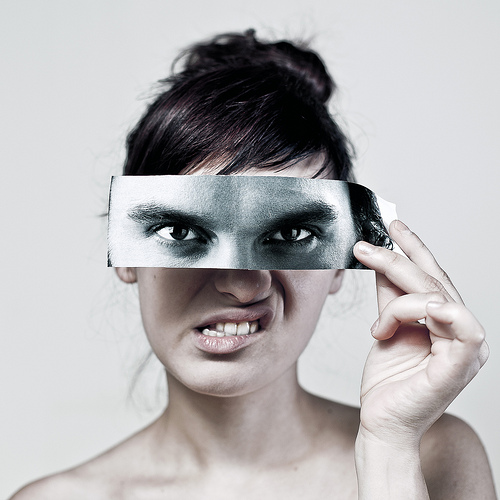}
    &
    \cincludegraphics[width=1in]{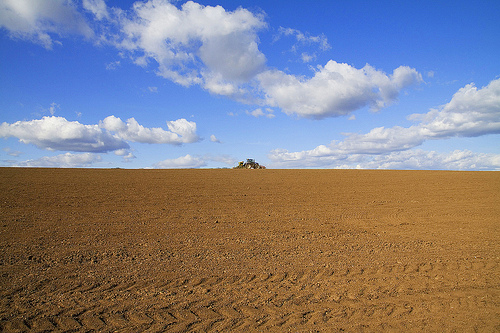}
    \cincludegraphics[width=0.8in]{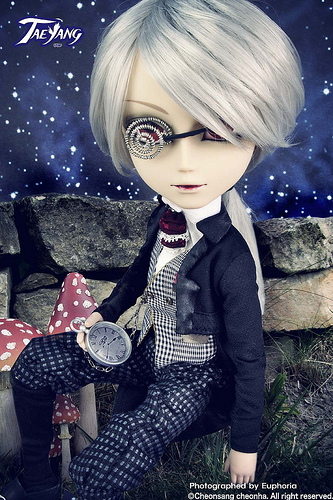}
    &
    \cincludegraphics[trim=13 0 13 0,clip,width=0.8in]{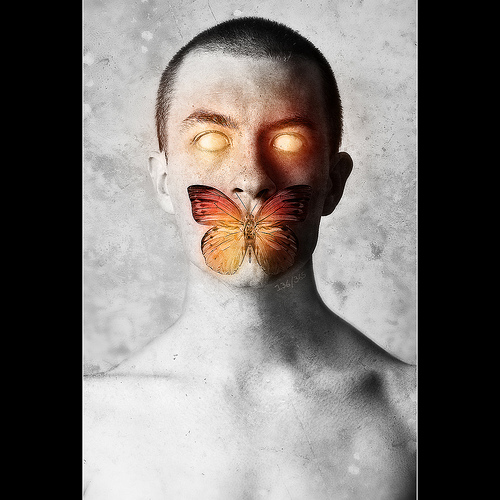}
    \cincludegraphics[width=1.1in]{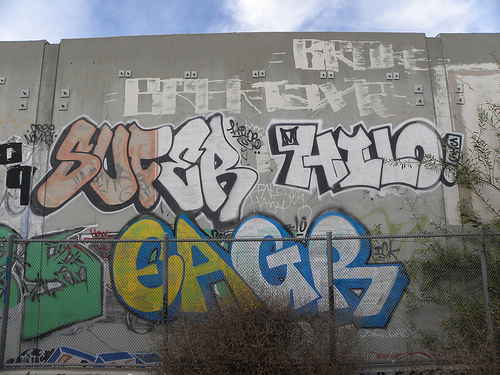}

\\ \hline
\end{tabular}
\end{phv}
\caption{Example images in the ISEE dataset~\cite{kim2018development}. Images were selected after a thorough test-retest reliability study.}\label{fig:ISEE}
\end{figure*}

\subsubsection{Evoked Emotion -- Immediate Response}

In the field of modeling evoked emotion, earlier researchers utilized the IAPS dataset, which consisted of only 1,082 images rated for evoked emotional response~\cite{lu2012shape}. In 2017, \citeauthor{lu_adams_li_newman_wang_2017}~\cite{lu_adams_li_newman_wang_2017} introduced one of the first large-scale datasets, the EmoSet, utilizing a human subject study. The EmoSet dataset is much larger, and all images are complex scenes that humans regularly encounter in daily life.

To create a diverse image collection, the researchers employed a data-crawling approach to gather nearly 44,000 images from social media, and obtained emotion labels (both dimensional and categorical) using crowdsourcing via the Amazon Mechanical Turk (AMT) platform. They used the VAD (Valence, Arousal, and Dominance) dimensional model~\cite{mehrabian1980basic}, which is similar to the PAD model. The researchers followed strict psychological subject study procedures and validation approaches. The images were collected from more than 1,000 users' Web albums on Flickr using 558 emotional words as search terms. These words were summarized by Averill~\cite{averill1975semantic}.

The researchers carefully designed their online crowdsourcing human subject study to ensure the quality of the data. For example, each image was presented to a subject for exactly six seconds. This design differed from conventional object recognition data annotation tasks, where the subject was often given no restrictions on the amount of time to view an image. This design followed psychological convention, as the intention was to collect the subject's {\it immediate} affective response to the visual stimuli. If subjects were given varying amounts of time to view an image before rating it, the data would not be a reliable capture of their immediate affective response. To accommodate for this, subjects were given the option to click a ``Reshow Image'' button if they needed to refer back to the image. In addition, recognizing that categorical emotions may not cover all feelings, this method allowed the subject to enter other feelings they may have had.

\subsubsection{Evoked Emotion -- Test-Retest Reliability}

The data collection method proposed by \citeauthor{lu_adams_li_newman_wang_2017}~\cite{lu_adams_li_newman_wang_2017} aimed to understand immediate affective responses to visual content, but it did not ensure retest reliability of affective picture stimuli over time and across a population. Many psychological studies, from behavioral to neuroimaging studies, have used visual stimuli that consistently elicited specific emotions in human subject. While the IAPS and other pictorial datasets have validated their data, they have not examined the retest reliability or agreement over time of their picture stimuli. 

To address this issue, \citeauthor{kim2018development}~\cite{kim2018development} developed the Image Stimuli for Emotion Elicitation (ISEE) as the first set of stimuli for which there was an unbiased initial selection method and with images specifically selected for high retest correlation coefficients and high within-person agreement across time. The ISEE dataset used a subset of 10,696 images from the Flickr-crawled EmoSet. In the initial screening study, study participants rated stimuli twice for emotion elicitation across a one-week interval, resulting in the selection of 1,620 images based on the number of ratings and retest reliability of each picture. Using this set of stimuli, a second phase of the study was conducted, again having participants rate images twice with a one-week interval, in which the researchers found a total of 158 unique images that elicited various levels of emotionality with both good reliability and good agreement over time. Fig.~\ref{fig:ISEE} shows 18 example images in the ISEE dataset.

\subsubsection{Expressed Emotion -- Body}\label{sec:bold}

In the field of expressed emotion recognition, the collection of data on bodily expressed emotions has received less attention compared to the more widely studied areas of facial expression and microexpression data collection. In addition, whereas earlier studies often relied on data collected in controlled laboratory environments, recent advancements in technology have made it possible to collect data in more naturalistic, real-world settings. These ``in-the-wild'' datasets are more challenging to collect, but they
offer the opportunity to capture a more diverse range of emotions and expressions. Whereas laboratory environments provide the advantage of advanced sensors such as Motion Capture (MoCap), body temperature, and brain electroencephalogram (EEG) for collecting data, and it is possible to capture self-
identified rather than perceived emotional expression, it is impossible to accurately replicate the vast array of diverse real-world scenarios within a controlled laboratory setting.

Using video clips from movies, TV shows, sporting, and wedding events as a source of data for emotion recognition has several advantages. These videos provide a wide range of scenarios, environments, and situations that can be used to train computer systems to understand human behavior, expression, and movement. For instance, these videos have recorded scenes during natural and man-made disasters, providing valuable information for understanding human emotions under extreme conditions. In addition, a large proportion of video shots in movies is of outdoor human activities, providing a diverse range of contexts for training.

However, it is important to note that using publicly available video clips as a source of data has its limitations. One such limitation is that this approach can only capture perceived emotions, as opposed to self-identified emotions. In many applications, perceived emotions are a sufficient proxy for actual emotions, particularly when the goal is for robots to ``perceive'' or ``express'' emotions in a way that is similar to humans for efficient communication with humans. {A further constraint is that the videos mainly feature staged or user-selected scenes, rather than depicting natural everyday interactions. This topic will be further explored in Section~\ref{sec:acted}.}

\begin{figure*}[ht!]
\centering
\includegraphics[trim=60 105 280 100,clip,width=\linewidth]{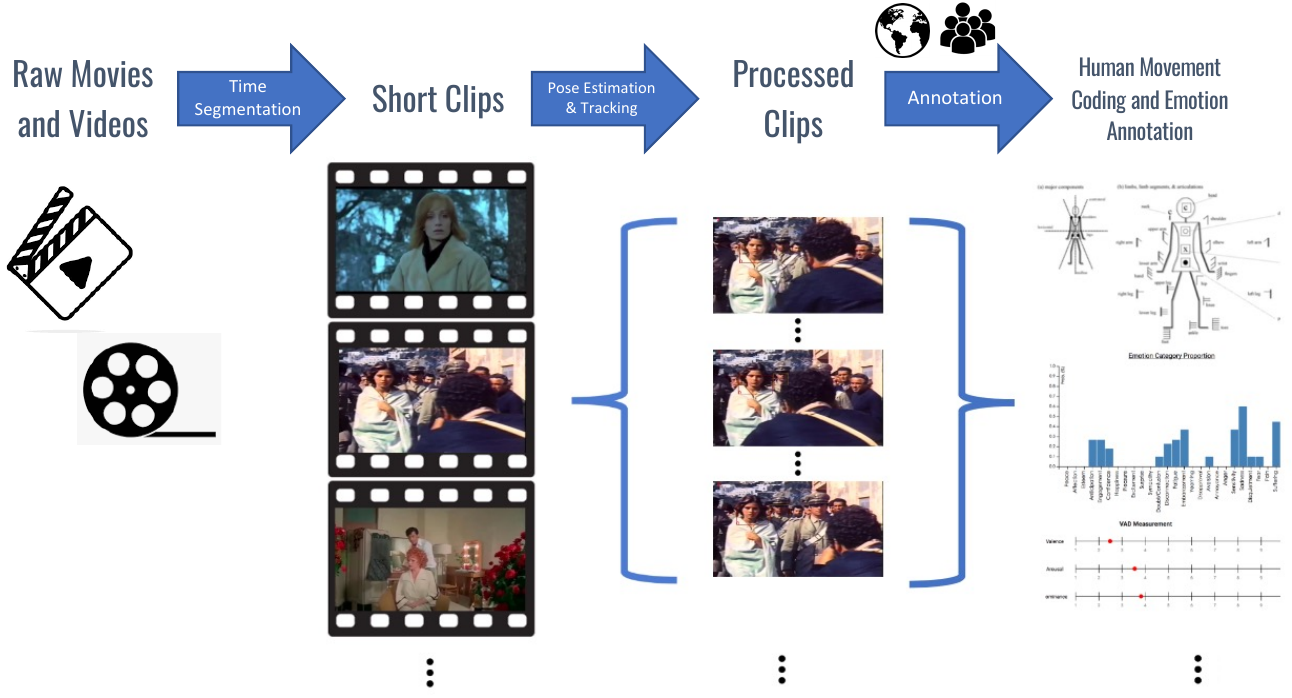}
\caption{A data collection pipeline was developed by \citeauthor{luo2020arbee}~\cite{luo2020arbee}to utilize crowdsourcing to annotate vast amounts of videos available on the Internet. The pipeline involved obtaining raw movies and videos from the Internet, segmenting them into short clips, processing the clips using computer vision techniques to extract posture information and track individual characters, and utilizing crowdsourcing and expert annotation to assign emotion and movement labels to the clips.}\label{fig:bold}
\end{figure*}

\citeauthor{luo2020arbee}~\cite{luo2020arbee} developed the first dataset for bodily expressed emotion understanding (BEEU), named the BoLD (Body Language Dataset), using this approach. The data collection pipeline is illustrated in Fig.~\ref{fig:bold}. The researchers collected hundreds of movies from the Internet and cut them into short clips. 
An identified character with landmark tracking in a single clip is called an {\it instance}. 
They used the AMT platform for crowdsourcing emotion annotations of a total of over 48,000 instances. 
The emotion annotation included the VAD dimensional model~\cite{mehrabian1980basic} and 26 emotion categories~\cite{Kosti_2017_CVPR}. 

\subsection{Data Quality Assurance}\label{sec:reliability}

Quality control is a {\it crucial} aspect for
crowdsourcing, particularly for affect annotations. Different individuals may have varying perceptions of affect, and their understanding can be influenced by factors such as cultural background, current mood, gender, and personal experiences. Even an honest participant may provide uninformative affect annotations, leading to poor-quality data. 
In this case, the variance in acquiring affect
usually comes from two kinds of participants, i.e., dishonest
ones, who give useless annotations for economic motivation, and exotic
ones, who give inconsistent annotations compared with others. The existence of exotic participants is inherent in emotion studies. 
The annotations provided by an exotic participant could be valuable when
aggregating the final ground truth or investigating cultural or gender
effects of affect. 
However, we typically want to reduce the risk of high
variance caused by dishonest and exotic participants in order to collect generalizable annotations.

In the case of the BoLD dataset~\cite{luo2020arbee}, five complementary
mechanisms were used, including three online approaches
(i.e., analyzing while collecting the data) and two offline
(i.e., postcollection analysis), based on a recent technological breakthrough for crowdsourced affective data collection~\cite{ye2017probabilistic}. These mechanisms were participant emotional intelligence (EQ) screening~\cite{wakabayashi2006development}, annotation sanity/consistency check~\cite{luo2020arbee}, gold standard test based on control instances~\cite{luo2020arbee}, and probabilistic multigraph modeling for reliability analysis~\cite{ye2017probabilistic}.

Particularly critical is the probabilistic graphical model \citeauthor{ye2017probabilistic} developed to
jointly model {\it subjective reliability}, which
is independent from supplied questions, and {\it regularity}~\cite{ye2017probabilistic}. 
For brevity of discussion, we focus on using the mode(s) of the posterior as point estimates.
We assumed each subject $i$ had a reliability parameter $\tau_{i} \in [ 0,1 ]$
and regularity parameters $\alpha_{i}$, $\beta_{i} > 0$ characterizing their agreement
behavior with the population, for $i=1, \ldots ,m$. We also used parameter $\gamma$ for the rate of agreement between subjects by pure chance. Let $\Theta = ( \{ \tau_{i} , \alpha_{i} ,
\beta_{i} \}_{i=1}^{m} , \gamma )$ be the set of parameters. Let $\Omega_{k}$
be a random sub-sample from subjects $\{ 1, \ldots ,m \}$ who labeled the
stimulus $k$, where $k=1, \ldots ,n$. We also assumed sets
$\Omega_{k}$'s were created independently from each other. For each image $k$, every subject
paired from $\Omega_{k}^{2}$, i.e., $( i,j )$ with $i \neq j$, had a binary indicator
$I_{i,j}^{( k )} \in \{ 0,1 \}$ coding whether their opinions agreed on the
respective stimulus. We assumed $I_{i,j}^{( k )}$ was generated from a probabilistic process involving two latent variables. The first latent variable $T_j^{(k)}$ indicated whether subject $O_j$ was reliable or not. Given that it was binary, a natural choice of model was the Bernoulli distribution. The second latent variable $J_i^{(k)}$, lying between 0 and 1, measured the extent to which subject $O_i$ agreed with other reliable responses. We used the Beta distribution parameterized by $\alpha_i$ and $\beta_i$ to model $J_i^{(k)}$ because it was a widely used and flexible parametric distribution for quantities on the interval $[0,1]$. 

\begin{figure}[ht!]
\centering
\includegraphics[trim=100 230 450 80,clip,width=0.4\textwidth]{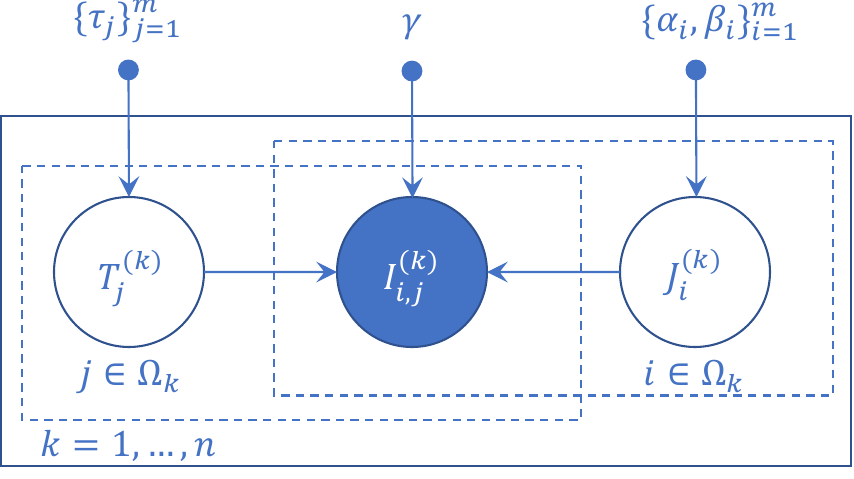}
\caption{Probabilistic graphical model GLBA, was developed by \citeauthor{ye2017probabilistic}~\cite{ye2017probabilistic}, to model the subjective reliability and regularity in a crowdsourced affective data collection. The work enables researchers in affective computing to effectively identify and exclude highly subjective annotation data provided by uninformative human participants, thereby improving the overall quality of the collected data.}\label{fig:pgm}
\end{figure}

In a nutshell, $T_{j}^{( k )}$ is a latent switch (a.k.a. gate) that controls whether $I_{i,j}^{( k )}$ can be used for the posterior inference
of the latent variable $J_{i}^{( k )}$. Hence, the researchers referred to the model as the {\em Gated Latent Beta Allocation} (GLBA). A graphical illustration of the model is shown in Fig.~\ref{fig:pgm}.
If an uninformative annotator was in the subject pool, their reliability parameter $\tau_i$ was zero,
though others could still agree with their answers by chance at a rate of $\gamma$.
On the other hand, if an individual was very reliable yet often provided controversial
answers, their reliability $\tau_i$ could be one, while they typically disagreed
with others, as indicated by their high irregularity
\begin{equation*}
    \mathbb E[J_i^{(k)}] = \frac{\alpha_i}{\alpha_i+\beta_i}\approx 0 \;.
\end{equation*}
We were interested in finding both types of participants. Most participants were between these two extremes. The quantitative characterization of participants by GLBA will assist in selecting subsets of the data collection for quality control or gaining a comprehensive understanding of subjectivity. For more details, please refer to~\cite{ye2017probabilistic,kim2012latent}.

{A recent study~\cite{stappen2021muse} presented a Python-based software program called MuSe-Toolbox, which combines emotion annotations from multiple individuals. 
The software includes several existing annotation fusion methods, such as Estimator Weighted Evaluator (EWE)~\cite{grimm2005evaluation} and Generic-Canonical Time Warping (GCTW)~\cite{zhou2009canonical}. 
In addition, the authors have developed a new fusion method based on EWE, named Rater Aligned Annotation Weighting (RAAW), which is also included in the software. 
Furthermore, MuSe-Toolbox includes the capability to convert continuous emotion annotations into categorical labels.}

\subsection{Existing Datasets}\label{sec:datasets}

Several recent literature surveys have provided an overview of existing datasets for emotions in visual media. In order to avoid duplication of effort, readers are directed to these papers for further information, which include surveys on evoked emotion~\cite{zhao2021affective}, BEEU~\cite{noroozi2021survey}, facial expression recognition (FER)~\cite{li2020deep}, microexpression recognition (MER)~\cite{ben2022video}, and multimodal emotion~\cite{zhao2021emotion,wang2015video,gandhi2022multimodal}. Table~\ref{tab:data} presents a comparison of the properties of some representative datasets. Researchers are advised to thoroughly review the data collection protocol used before utilizing a dataset to ensure that the data has been collected in accordance with appropriate psychological guidelines. In addition, when crowdsourcing is utilized, effective mechanisms are essential for filtering out uninformative annotations.
%\textcolor{red}{will revise this section}
\begin{table*}[t!]
\centering
\caption{Recent representative datasets for emotion recognition}\label{tab:data}
\colorbox{lightyellow}{%
\begin{tabular}{|l|c|c|c|c|c|c|c|} 
\hline
\textbf{Dataset Name}                   & \begin{tabular}[c]{@{}l@{}}\textbf{Labeled}\\\textbf{Samples}\end{tabular} & \begin{tabular}[c]{@{}c@{}}\textbf{Data}\\\textbf{Type}\end{tabular} & \begin{tabular}[c]{@{}c@{}}\textbf{Categorical}\\\textbf{Emotions}\end{tabular} & \begin{tabular}[c]{@{}c@{}}\textbf{Continuous}\\\textbf{Emotions}\end{tabular} & \begin{tabular}[c]{@{}c@{}}\textbf{Lab-}\\\textbf{Controlled}\end{tabular} & \textbf{Year} & \begin{tabular}[c]{@{}c@{}}\textbf{Primary}\\\textbf{Application}\end{tabular} \\
\hline
IAPS~\cite{lang2005international} & 1.2k & I & - & VAD & & 2005 & Evoked \\
\hline
FI~\cite{you2016building} & 23.3k & I & 8\textsuperscript{$\S$} & - & & 2016 & Evoked \\
\hline
VideoEmotion-8~\cite{jiang2014predicting} & 1.2k & V & 8\textsuperscript{†} & - & & 2014 & Evoked \\
\hline
Ekman-6~\cite{xu2018heterogeneous} & 1.6k & V & 6\textsuperscript{†} & - & & 2018 & Evoked \\
\hline
E-Walk~\cite{randhavane2019identifying} & 1k & V & 4\textsuperscript{‡} & - &   & 2019 & BEEU \\ 
\hline
BoLD~\cite{luo2020arbee} & 13k& V & 26\textsuperscript{†} & VAD   &   & 2020 & BEEU  \\ 
\hline
{iMiGUE~\cite{liu2021imigue}} & {0.4k} & {V} & {2} & {-}   &   & {2021} & {BEEU\textsuperscript{$\star$}} \\ 
\hline
CK+~\cite{lucey2010extended} & 0.6k & V  & 7\textsuperscript{‡}  & - & \checkmark   &     2010     & FER \\ 
\hline
Aff-Wild~\cite{10.1007/s11263-019-01158-4}  & 0.3k& V & - & VA    &   &  2017  & FER \\ 
\hline
AffectNet~\cite{mollahosseini2017affectnet} & 450k & I & 7\textsuperscript{‡}  & - &   &  2017   & FER  \\ 
\hline
EMOTIC~\cite{kosti_emotic_2017} & 34k & I  & 26\textsuperscript{†} & VAD     &  & 2017    & FER\textsuperscript{$\ast$} \\
\hline
AFEW 8.0~\cite{6200254}   & 1.8k & V & 7\textsuperscript{‡} & - &  &  2018  & FER  \\ 
\hline
CAER~\cite{lee2019context} & 13k & V  & 7\textsuperscript{‡} &    -  &  & 2019    & FER\textsuperscript{$\ast$}  \\
\hline
DFEW~\cite{jiang2020dfew}  & 16k & V & 7\textsuperscript{‡}  & - &   &  2020   & FER  \\ 
\hline
FERV39k~\cite{wang2022ferv39k}  & 39k & V & 7\textsuperscript{‡} & -  &  &  2022   & FER  \\ 
\hline
SAMM~\cite{davison2016samm}  & 0.2k &V & 7\textsuperscript{‡}  & - & \checkmark  &  2016   & MER  \\ 
\hline
CAS(ME)$^2$~\cite{qu2017cas}  & 0.06k &V & 4\textsuperscript{‡}  & - & \checkmark  &  2017   & MER  \\ 
\hline
ICT-MMMO~\cite{wollmer_youtube_2013}   & 0.4k & V,A,T & - & Sentiment &  & 2013 & Multi-Modal   \\ 
\hline
MOSEI~\cite{bagher_zadeh_multimodal_2018} & 23.5k& V,A,T & 6\textsuperscript{†} & Sentiment  &   & 2018 & Multi-Modal  \\ 
\hline
\end{tabular}}
\begin{tabular}{l} 
\begin{tabular}[c]{@{}l@{}}\textsuperscript{†} A superset of Ekman's basic emotions\quad \textsuperscript{‡} Ekman's basic emotions + neutral\quad \textsuperscript{$\S$} Mikels' emotions  \end{tabular} \\
\begin{tabular}[c]{@{}l@{}}{\textsuperscript{$\star$} Micro-gesture understanding and emotion analysis dataset} \quad  \quad    \quad \textsuperscript{$\ast$} Context-aware emotion dataset \end{tabular} \\
\begin{tabular}[c]{@{}l@{}}Data Type Key: (I)mage, (V)ideo, (A)udio, (T)ext \end{tabular} \\
\end{tabular}
\end{table*}

\subsection{Data Representations}\label{sec:representation1}

After data collection and quality assurance stages, a significant technological challenge is to represent the {\it emotion-relevant information} present in the raw data in a concise form. Whereas current deep neural network (DNN) approaches often utilize raw data, such as matrices of pixels, as input in the modeling process, utilizing a compact data representation can potentially improve the efficiency of the learning process, allowing for larger-scale experiments to be conducted with limited computational resources. In addition, a semantically meaningful data representation can facilitate interpretability, which is crucial for certain applications. There are numerous methods for compactly representing raw visual data, and we discuss several intriguing or widely used data representations for emotion modeling in the following.

\subsubsection{Roundness, Angularity, Simplicity, and Complexity} 

\begin{figure*}[ht!]
\begin{phv}
\centering
%\resizebox{\linewidth}{!}{%
\begin{tabular}{Sc | Sc | Sc | Sc}
    &  {Roundness} & { Angularity} & { Simplicity}\\
    \hline
{\STAB{\rotatebox[origin=c]{90}{High}}} &
    \cincludegraphics[width=1.8in]{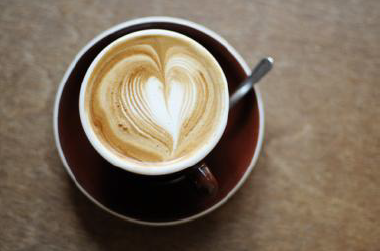} 
    & 
    \cincludegraphics[width=1.8in]{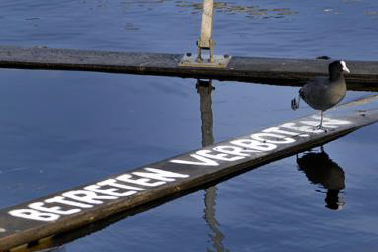} 
    & 
    \cincludegraphics[height=1.6in]{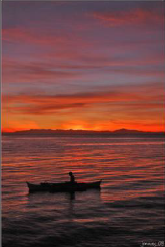} 
\\ \hline
{\STAB{\rotatebox[origin=c]{90}{Medium}}} &     \cincludegraphics[width=1.4in]{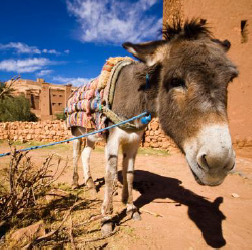} 
    & 
    \cincludegraphics[width=1.8in]{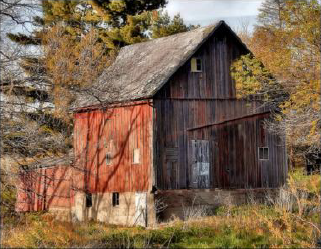} 
    & 
    \cincludegraphics[width=1.8in]{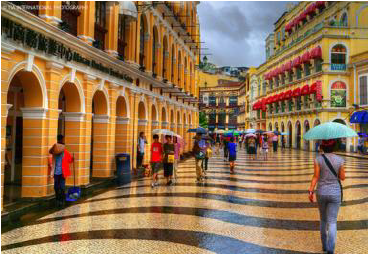} 

\\ \hline
{\STAB{\rotatebox[origin=c]{90}{Low}}} &     \cincludegraphics[width=1.4in]{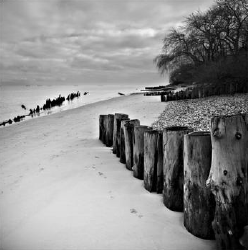} 
    & 
    \cincludegraphics[width=1.8in]{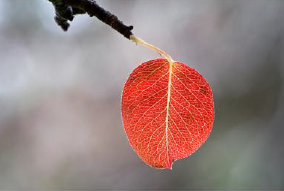} 
    & 
    \cincludegraphics[width=1.5in]{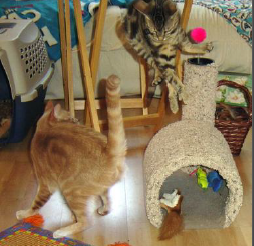} 

\\ \hline
\end{tabular}
\end{phv}
\caption{Researchers developed algorithms to compute the roundness, angularity, and simplicity/complexity of a scene as the representation for estimating the evoked emotion~\cite{lu_adams_li_newman_wang_2017}.}\label{fig:three_constructs}
\end{figure*}

Colors and textures are commonly used in image analysis tasks to represent the content of an image, but research has shown that shape can also be an effective representation when analyzing evoked emotions. In both visual art and psychology, the characteristics of shapes, such as {\it roundness}, {\it angularity}, {\it simplicity}, and {\it complexity}, have been linked to specific emotional responses in humans. For example, round and simple shapes tend to evoke positive emotions, while angular and complex shapes evoke negative emotions. Leveraging this understanding, \citeauthor{lu2012shape} developed a system that predicted evoked emotion based on line segments, curves, and angles extracted from an image~\cite{lu2012shape}. They used ellipse fitting to implicitly estimate roundness and angularity and used features from the visual elements to estimate complexity. Later, they developed algorithms to explicitly estimate these representations~\cite{lu_adams_li_newman_wang_2017}. Fig.~\ref{fig:three_constructs} shows some example images with different levels of roundness, angularity, and simplicity. The researchers found that these three physically interpretable visual constructs achieved comparable classification accuracy to the hundreds of shape, texture, composition, and facial feature characteristics previously examined. This result was thought-provoking because just a few numerical-value representations could effectively predict evoked emotions.

\subsubsection{Facial Action Coding System (FACS) and Facial Landmarks} 
People use particular facial muscles to express certain facial expressions.
For instance, people can express anger by frowning and pursing their lips.
Consequently, each facial expression can be viewed as a combination of some facial muscle movements.
Ekman and Friesen developed the FACS in 1976, which encodes all movements of facial muscles~\cite{ekman1977facial}. 
FACS defines a total of 32 atomic facial muscle actions, called Action Units (AUs), including Lids Tight (AU7), Cheek Raise (AU6), and so on.
By detecting all AUs of a person and linking them to specific expressions, we can identify the individual's facial expressions.

The problem of AU detection can be approached as a multilabel binary classification problem for each AU.
Early work on AU detection used facial landmarks to identify regions of corresponding muscles and then applied neural networks~\cite{tian2001recognizing} or Support Vector Machines (SVMs)~\cite{valstar2006fully} for classification.
More recent work has developed end-to-end AU detection networks~\cite{jacob2021facial}.
Survey papers provide detailed introductions to facial AU detection~\cite{martinez2017automatic,zhi2020comprehensive} and face landmark detection~\cite{wang2018facial,wu2019facial}.
Some researchers also used facial landmarks directly as a representation of facial information in their recognition work.

\subsubsection{Body Pose and Body Mesh} 
People can express emotions through body posture and movement.
By manipulating the positioning of body parts (e.g., the shoulders, arms), people produce various postures and movements.
The coordinates of human joints can serve as a representation of body language, reflecting the individual's bodily expression.
In the field of computer vision, two-dimensional (2D) pose estimation is a well-studied task for detecting the 2-D position of human joints in an image. 
Leveraging large-scale 2-D pose datasets (e.g., COCO~\cite{lin2014microsoft}), researchers have proposed several high-performing pose networks~\cite{sun2019deep,xiao2018simple}. 
Even with challenging scenes, such as  crowded or occluded scenes, these networks are able to provide comprehensive joint detection and linking.

However, 2-D pose estimation does not fully capture the three-dimensional (3D) nature of human posture and movement. 3-D human pose estimation, on the other hand, aims to predict the 3-D coordinates of human joints in space.
Single-person 3-D pose estimation methods determine the 3-D joint coordinates relative to the person's root joint (i.e., the torso)~\cite{sun2018integral,wu2020mebow}. 
In addition, some multiperson 3-D pose estimation approaches comprehensively estimate the absolute distance between the camera and the individuals in the image~\cite{moon2019camera}.

\begin{figure*}[ht!]
    \centering
    \begin{tabular}{ccc}
    \includegraphics[trim=0 0 0 0,clip,width=0.3\linewidth]{fig/pose_raw.jpg} &
        \includegraphics[trim=0 0 0 0,clip,width=0.3\linewidth,cfbox=lightgray 1pt 1pt]{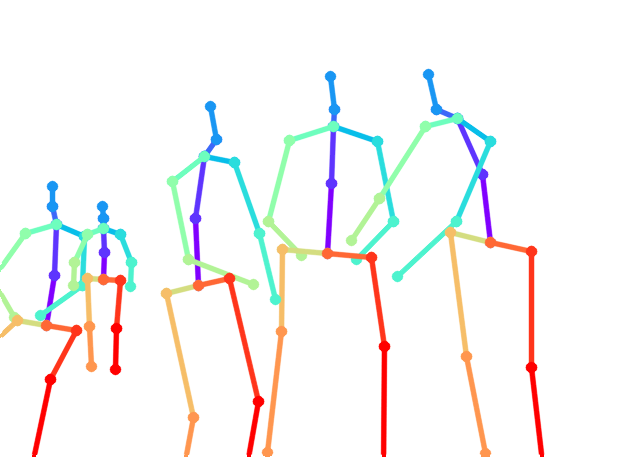} &
                 \includegraphics[trim=0 0 0 0,clip,width=0.3\linewidth]{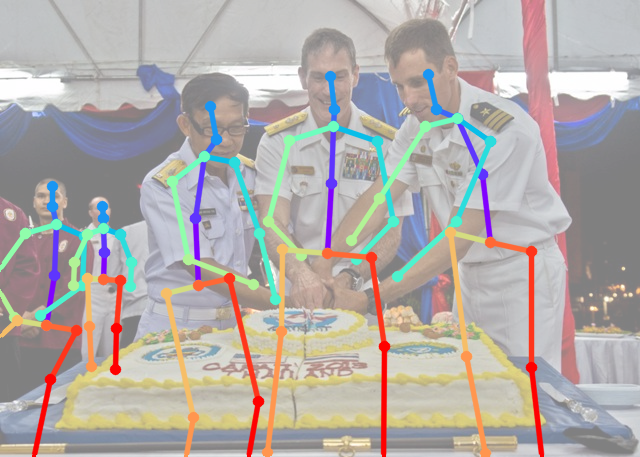} \\
         (a) original photo & (b) 2-D pose & (c) 2-D pose, superimposed \\ \includegraphics[trim=80 95 55 95,clip,width=0.3\linewidth]{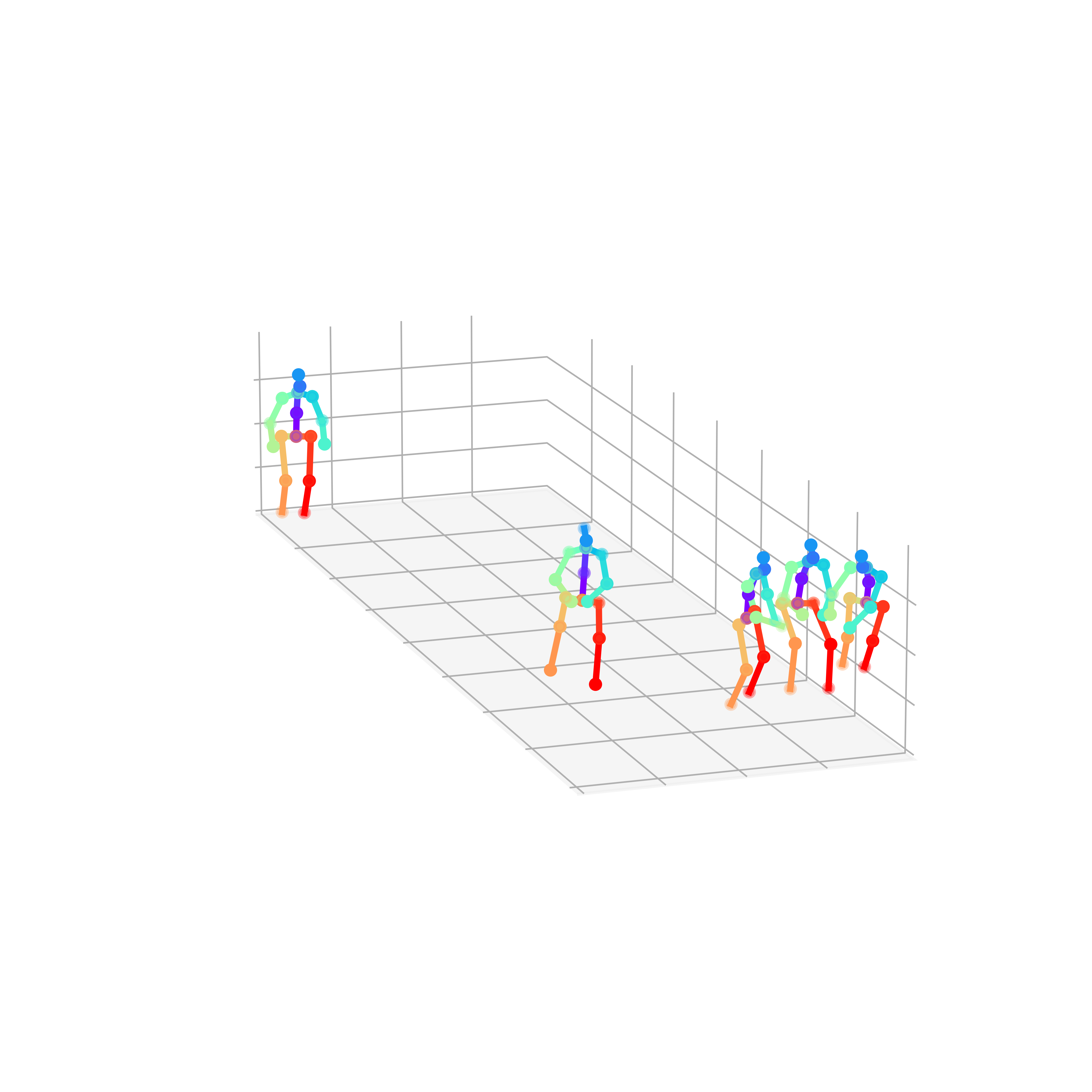}&
               \includegraphics[trim=0 0 0 0,clip,width=0.3\linewidth]{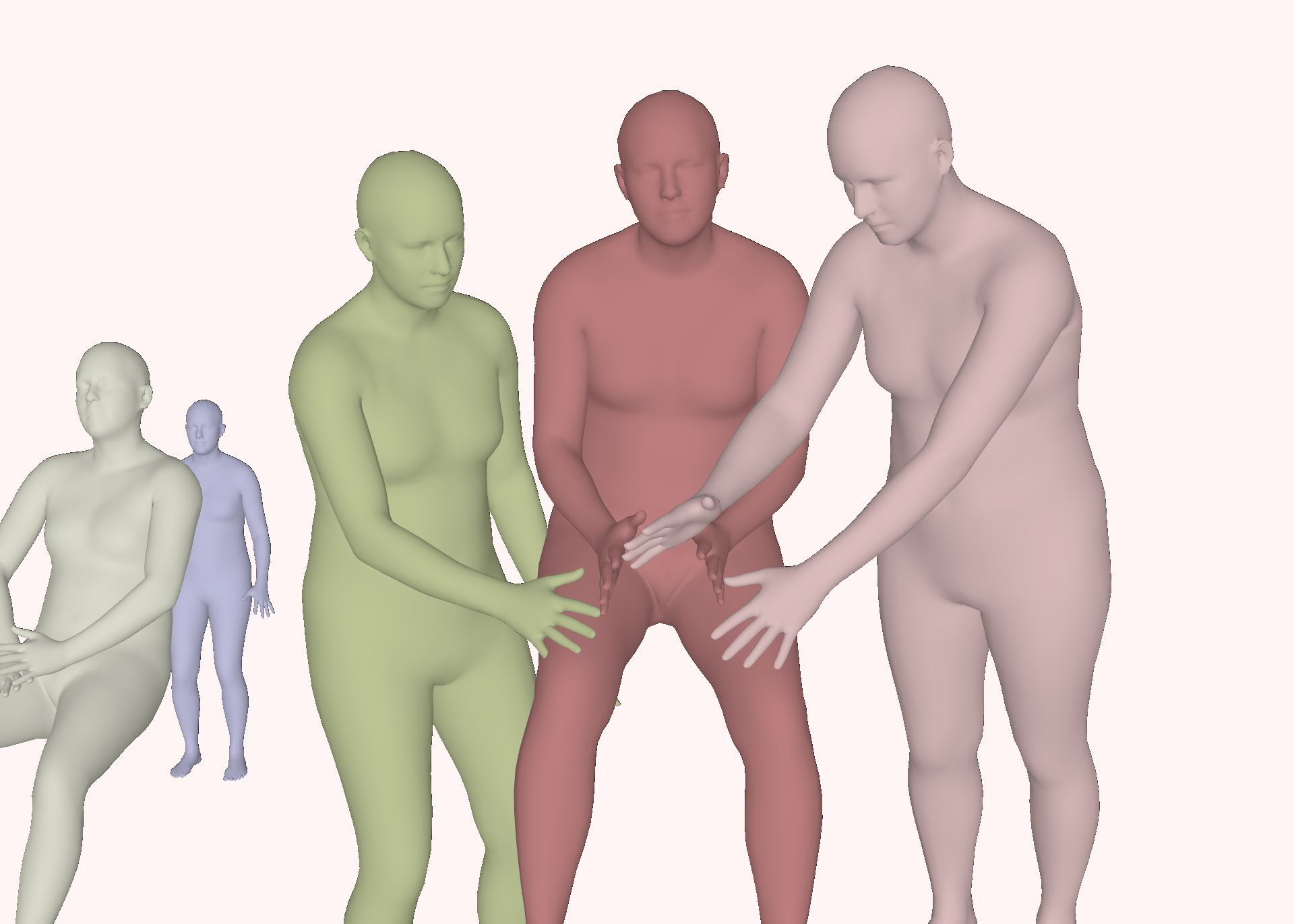}&
    \includegraphics[trim=0 0 0 0,clip,width=0.3\linewidth]{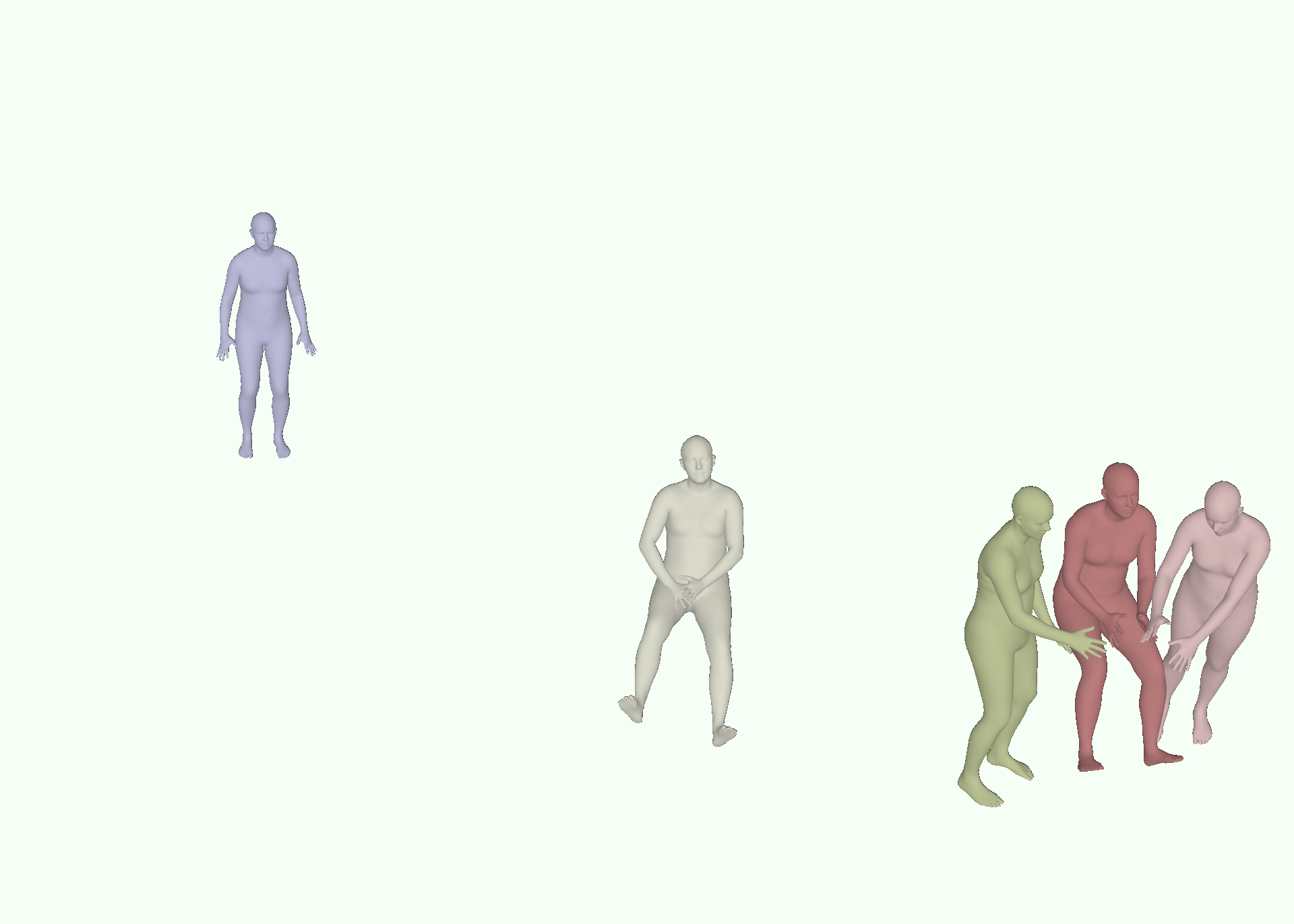}\\
     (d) 3-D pose & (e) 3-D mesh & (f) 3-D mesh, alternative view
    \end{tabular}
    \caption{An example of state-of-the-art deep learning-based human pose and mesh detection with individuals in the scene separated properly~\cite{wu2022mug}. The input to the methods is a standard 2-D RGB photograph. The 3-D mesh method provides an estimation of the relative depths of the people in the scene.}\label{fig:pose}
    \end{figure*}

3D human mesh estimation, which provides the 3-D coordinates of each point on the human mesh, is a further extension of the 3-D pose estimation. Researchers often utilize SMPL~\cite{loper2015smpl,bogo2016keep} or other human graph models to represent the mesh. Fig.~\ref{fig:pose} illustrates an example of 2-D pose, 3-D pose, and 3-D mesh automatically generated from a scene with multiple people~\cite{wu2022mug}.

\begin{figure}[ht!]
\centering
\includegraphics[trim=190 30 310 195,clip,width=0.35\textwidth]{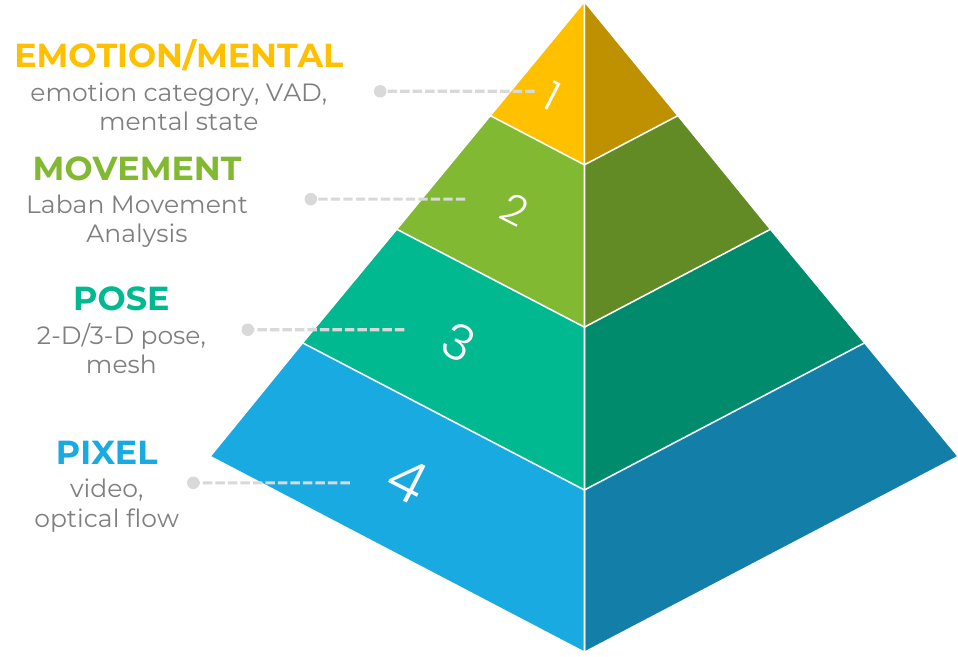}
\caption{For BEEU, Laban Movement Analysis as a movement coding system can serve as an intermediate level of representation to bridge the significant gap between the pose and the emotion/mental levels~\cite{luo2020arbee}.}
\label{fig:layers}
\end{figure}

While human pose and human mesh representations provide a higher level of abstraction compared to low-level representations such as raw video or optical-flow, there is still a significant gap between these intermediate-level representations and high-level emotion representations. To bridge this gap, an intermediate-level representation that effectively describes human movements is proposed. Specifically, Laban Movement Analysis (LMA) is a movement coding system initially developed by the dance community, similar to sheet music for describing music. Fig.~\ref{fig:layers} illustrates the layers of data representation for BEEU, from low pixel-level representations to the ultimate high-level emotion representation. Elevating each layer higher in this information pyramid requires considerable advancements in abstraction technology.

\subsection{Human Movement Coding and  LMA}\label{sec:representation2}

Expressive human movement is a complex phenomenon. The human body has 360 joints, many of which can move various distances at different velocities and accelerations, and in two (and depending on the joint, sometimes more) different directions, resulting in an astronomical number of possible combinations. These variables create an infinite number of movements and postures that can convey different emotions. Beyond this array of body parts moving in space, expressive movement also involves multiple qualitative motor components that appear in unique patterns in different individuals and situations. The complexity of human movement thus  raises the question: {\it How do we determine which of the numerous components present in expressive movement are significant to code for emotional expression in movement?} Thus, when choosing a coding system, the early stages of each research project can benefit from deeply considering which aspects of movements are central to the expression being studied. A multistage methodology, such as first identifying what is potentially relevant and then using preliminary analyses to refine the selection of movements most promising to code, can be helpful before selecting a method to code or quantify the multitude of variables present in unscripted movement (e.g.,~\cite{ahmed2019emotion,tsachor2019shall}).

After deliberating about which movement variables are relevant and meaningful, we must then consider the three main types of coding systems that have been used in various fields such as psychology, computer vision, animation, robotics, and AI, specifically: 
\begin{itemize}
\item Lists of specific motor behaviors that have been found in scientific studies to be typical to the expressions of specific emotions, such as head down and moving slowly as characterizing sadness; moving backward and bringing the arms in front of the body as characterizing fear; jumping, expanding and upward movements as characterizing happiness, and so on. (for review of these studies and lists of these behaviors see~\cite{witkower2019bodily,kleinsmith2013affective}). 
\item Kinematic description of the human body models, such as skeleton-based models, contour-based models, or volume-based models. Most work in the field of emotion recognition is based on skeleton-based models~\cite{ebdali2022overview}. This type of model uses 3-D coordinates of markers that were placed on (using various MoCap systems) or were mapped (using pose estimation techniques) to the main joints of the body, to create a moving ``skeleton,'' which enables researchers to quantitatively capture the movement kinematics (e.g.,~\cite{kleinsmith2011automatic,roether2009critical}).    
\item LMA, a comprehensive movement analysis system that depicts qualitative aspects of movement and, theoretically~\cite{bartenieff1980body,studd2013everybody}, as well as through scientific research~\cite{melzer2019we,shafir2016emotion}, relates the different LMA motor elements (movement qualities) to cognitive and emotional aspects of the moving individual. 
\end{itemize}

The first coding system, based on lists of motor behaviors, was primarily used in earlier studies in the field of psychology, where the encoding and decoding of motor behaviors into emotions and vice versa were done manually by human coders in a labor-intensive process. Other limitations of this coding system include the following: 
\begin{itemize}
\item
It is based on a limited number of behaviors that have been used in prior scientific studies. However, people can physically express emotions in many different ways, so the list of previously observed and studied whole-body expressions may not be exhaustive or inclusive of cultural variations.
\item Likewise, because each study used different lists of behaviors, this method makes it difficult to compare results or to review them additively to arrive at larger verification. Thus, this coding system may miss parts of the range of bodily emotional expression such as those never observed and coded before. This limitation is especially pronounced because many of these previous studies examined emotional expressions performed by actors, whose movement tended to rely upon more stereotypical bodily emotional expressions that were widely recognized by the audience, rather than naturally occurring motor expressions.
\end{itemize}

When using the second coding system, kinematic description, in particular the skeleton-based models, researchers usually employ a set of markers similar to, or smaller than that provided by the Kinect Software Development Kit, and transform the large amount of 3-D data into various low- and/or high-level kinematic and dynamic motion features, which are easier to interpret than the raw data. Researchers using this method have studied specific features such as head and elbow angles~\cite{roether2009critical}, maximum acceleration of hand and elbow relative to spine~\cite{saha2014study}, distance between joints~\cite{zacharatos2021emotion}, among others. For a review of such studies, please refer to~\cite{noroozi2021survey}. In recent years, instead of computing handcrafted features on the sequence of body joints, researchers have employed various deep learning methods to generate representations of the dynamics of the human body embedded within the joint sequence, such as spatiotemporal graph convolutional network (ST-GCN)~\cite{luo2020arbee,shi2020skeleton,ghaleb2021skeleton}. Although the 3-D data from joint markers can provide a relatively detailed, objective description of whole-body movement, this coding system has two main limitations: 
\begin{itemize}
    \item Movement is often captured by a camera from a single view (usually the frontal), which can result in long segments of missing data from markersthat are hidden by other body parts, people, or objects in a video frame. Automatic imputations of such missing data are often impractical as they tend to create incorrect and unrealistic movement trajectories. 
    \item The SDK system has only three markers along the torso, which are  insufficient for capturing subtle movements in the chest area, movements that are usually observed during authentic emotional expressions, as opposed to acted (and often exaggerated) bodily emotional expressions. Another disadvantage is that these systems have not yet been able to successfully and reliably detect many qualitative changes in movement that are significant for perceiving emotional expression.
\end{itemize}

In contrast to the quantitative data from joint markers, which enable the capture of detailed movement of every body part, the third coding system mentioned above, LMA, describes {\it qualitative} aspects of movement and can relate to a general impression from movements of the entire body or to the movement of specific body parts. By identifying the association between LMA motor components and emotions, and characterizing the typical kinematics of these components using high-level features (e.g.,~\cite{ajili2019human,aristidou2015emotion,dewan2018laban,luo2020arbee,senecal2016continuous,wang2020dance}), researchers can overcome the limitations of other coding systems. If people express their emotions with movements that have never been observed in previous studies, we can still decode their emotions based on the quality of their movement. Similarly, if parts of the body that are usually used to express a certain emotion are not visible, it is possible that the emotion could still be decoded by identifying the motor elements composing the visible part of the movement. Moreover, by slightly changing the kinematics of a movement of a robot or animation (i.e., adding to a gesture or a functional movement the kinematics of certain LMA motor elements associated with a specific emotion), we can ``color'' this functional movement with an expression of that emotion, even when the movement is not the typical movement for expressing that emotion (e.g.,~\cite{cui2019laban,inthiam2018development}). Similarly, identifying the quality of a movement can enable decoding the expressed emotion even from functional actions such as walking~\cite{gross2012effort} or reaching and grasping. These advantages and the fact that LMA features have been found to be positively correlated with emotional expressions are why LMA coding is becoming popular in studies that encode or decode bodily emotion expressions (e.g.,~\cite{luo2020arbee,senecal2016continuous}).  In addition, LMA offers the option to link our coding systems to diverse ways in which humans talk about and describe expressive movement--it is a comprehensive movement-theory system that is and can be used across disciplines for application in acting~\cite{adrian2010actor}, therapy~\cite{tsachor2017somatic}, education~\cite{fernandes2014moving}, and animation~\cite{bishko2014animation}, among others. {The last advantage to consider is that LMA is a comprehensive theory of body movement, much like art theory or music theory, including theories of harmony, and thus has been used by artists to attune to aesthetics including movement-perception of visual art (such as that discussed in Section~\ref{sec:visualart}) and visual, auditory and movement elements of film and theatre (discussed in Section~\ref{sec:acted}). Like music theory, LMA is capable of attending to rhythm and phrasing as elements shift and unfold over time, aspects that may be crucial to communicating and interpreting emotion expression.} 

\begin{figure}[ht!]
\centering
\includegraphics[width=0.47\textwidth]{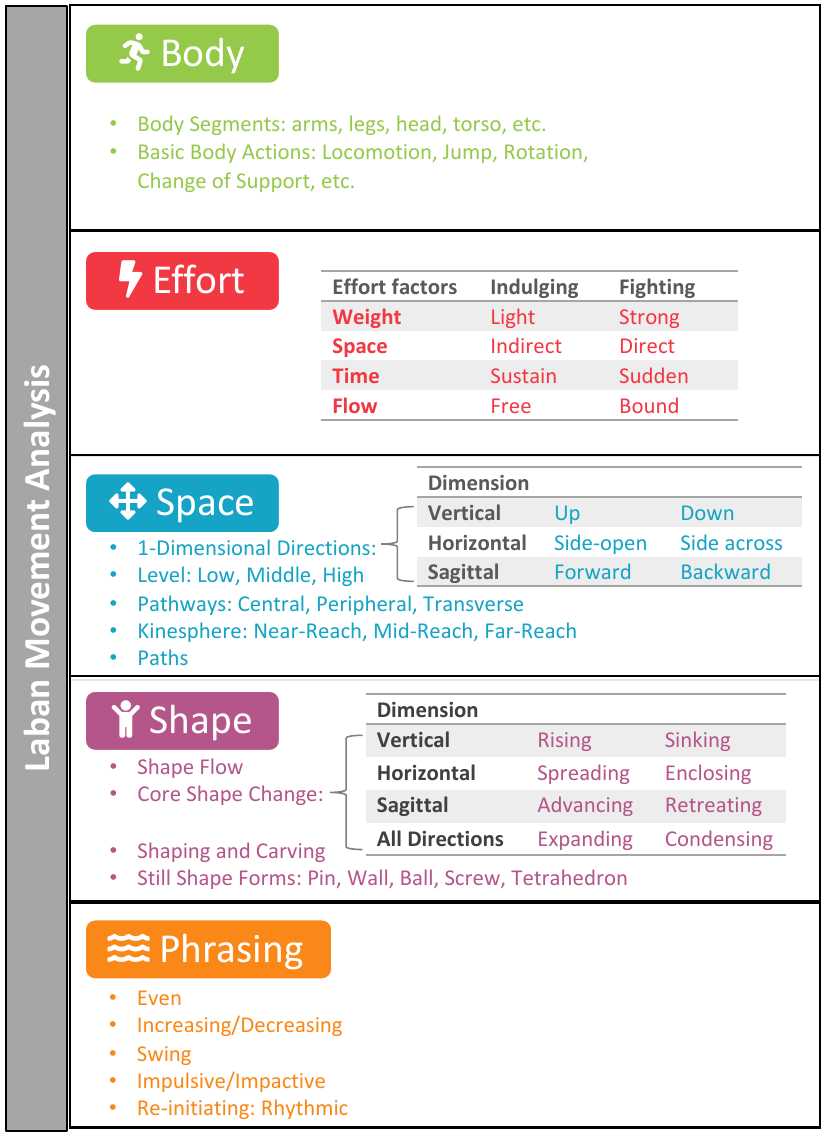}
\caption{Some basic components of LMA that are often used in coding.}\label{fig:lma}
\end{figure}

LMA identifies four major categories of movement: \textbf{Body}, \textbf{Effort}, \textbf{Shape}, and \textbf{Space}. Each category encompasses several subsets of motor components (LMA terms are spelled with capital letters to differentiate them from regular usage of these words). Fig.~\ref{fig:lma} illustrates some basic components of LMA that are often used in coding.

The \textbf{Body} category describes what is moving and it is composed of the elements of {\it Body Segments} (e.g., arms, legs, head), their coordination, and basic {\it Body Actions} like Locomotion, Jump, Rotation, Change of Support, and so on.

The \textbf{Effort} category describes qualitative aspect of movement, or how we move. It expresses a person's inner attitude toward movement, and it has four main factors, each describing the continuum between two extremes: indulging in the motor quality of that factor and fighting against that quality. Effort has four factors:
\begin{itemize}
\item {\it Weight Effort}, meaning the amount of force or pressure exerted by the body. Activated Weight Effort can be {Strong} or {Light}. Alternatively, there may be a lack of weight activation when we give in to the pull of gravity, which we describe as {Passive} or {Heavy Weight}. 
\item {\it Space Effort} describes  attention to space, denoting the focus or attitude toward a chosen pathway, i.e., is the movement {Direct} or flexibly {Indirect}.
\item {\it Time Effort}, describing the mover's degree of urgency or acceleration/deceleration involved in a movement, i.e., is the movement {Sudden} or {Sustained}.
\item {\it Flow Effort}, reflecting the element of control or the degree to which a movement is {Bound}, i.e., restrained or controlled by muscle contraction (usually cocontraction of agonist and antagonist muscles), versus {Free}, i.e., being released and liberated.
\end{itemize}

The \textbf{Shape} category reflects why we move: Shape describes how the body adapts its shape as we respond to our needs or the environment: Do I want to connect with or avoid something, dominate or cower under? The way the body sculpts itself in space reflects a relationship to self, others, or to the environment. This component includes {\it Shape Flow} which describes how the body changes to relate to oneself; it includes {\it Shape Change} which describes changes in the form or shape of the body; and includes the motor components of {Expanding} the body or {Condensing} it in all directions and {Rising} or {Sinking} in the vertical dimension, {Spreading} and {Enclosing} in the horizontal dimension and {Advancing} or {Retreating} in the sagittal dimension. Another Shape component is {\it Shaping and Carving} which describes how a person shapes their body to shape or affect the environment or other people. For example, when we hug somebody, we might shape and carve the shape of our body, adjusting it to the shape of the other person's body, or we might shape our ideas by carving or manipulating them through posture and gesture. 

The \textbf{Space} category describes where the movement goes in the environment. It describes many spatial factors such as the {\it Direction} where the movement goes in space, such as: {Up} and {Down} in the vertical dimension, {Side open} and {Side across} in the horizontal dimension, and {Forward} and {Backward} in the sagittal dimension; the {\it Level} of the movement in space relative to the entire body or parts of the body, such as {Low level} (movement toward the ground), {Middle level}, (movement maintaining level, without lowering or elevating) or {High level} (moving upward in space); {\it Paths} or how we travel through space by locomoting; and Pathways through the {\it Kinesphere} (the personal bubble of reach-space around each mover that can be moved in and through without locomoting or traveling.) Movement in the Kinesphere might take {Central} pathways, crossing the space close to the mover's body center, {Peripheral} pathways along the periphery of the mover's reach space, or {Transverse} pathways cutting across the reach space.

In addition, another important aspect of LMA which is particularly helpful and meaningful to expression is the \textbf{Phrasing} of movements. Phrasing describes changes over time, such as changes in the intensity of the movement over time, similar to musical phrases, which can be {Increasing}, {Decreasing}, or {Rhythmic}, among others. It can also depict how movement components shift during the same action or a series of actions occurring over time, for example beginning emphatically with strength and then ending by making a light direct point conclusion. 

Previous research has highlighted the lack of a notation system that directly encodes the correspondence between bodily expression and body movements in a way similar to FACS for face~\cite{luo2020arbee,ahmed2019emotion}. LMA, by its nature, has the potential to serve as such a coding system for emotional expressions through body movement. \citeauthor{shafir2016emotion}~\cite{shafir2016emotion} identified the LMA motor components (qualities), whose existence in a movement could evoke each of the four basic emotions: anger, fear, sadness, and happiness. \citeauthor{melzer2019we}~\cite{melzer2019we} identified the LMA components whose existence in a movement caused that movement to be identified as expressing one of those four basic emotions. In an additional experiment by the Shafir group, Gilor, for her Master's thesis, has been studying the motor elements that are used for expressing sadness and happiness. In this series of studies, the LMA motor components found to evoke these emotions through movement were, for the most part, the same as those used to identify or recognize each emotion from movement, or to express each emotion through movement. For example, anger was associated with Strong Weight Effort, Sudden Time Effort, Direct Space Effort, and Advancing Shape during both emotion elicitation and emotion recognition. Fear was associated with Retreating Shape, moving backward in Space for both emotion elicitation and recognition. In addition, Enclosing and Condensing Shape, and Bound Flow Effort, were also found for emotion elicitation through movement. Sadness was associated with Passive Weight Effort, Sinking Shape, and Head drop for emotion elicitation, emotional expression, and emotion recognition, and arms touching the upper body was also a significant indicator for emotion elicitation and expression. Sadness expression was also associated with using Near-Reach Space and Stillness. In contrast, Happiness was associated with jumping, rhythmic movement, Free and Light Efforts, Spreading and Rising Shape, and moving upward in Space for emotion elicitation, emotion expression, and emotion recognition. Happiness expression was also associated with Sudden Time Effort and Rotation. These findings for Happiness were also validated by \citeauthor{van2021move}~\cite{van2021move}. Whereas these studies represent a promising start, further research is needed to create a comprehensive LMA coding system for bodily emotional expressions. 

\subsection{Context and Functional Purpose}\label{sec:context}
While human emotions are shown through the face and body, they are closely connected to context and purpose. Thus, the image context surrounding the human can also be used to further identify human emotions.
The context information includes the semantic information of the background and what the person is holding, which can assist in identifying activities of the human, thereby allowing for more accurate prediction of the human's emotion. 
For instance, people are more likely to feel happy than sad during a birthday party.
Context information also includes interactions among humans, which can help infer emotion. 
For example, people are more likely to be angry when they are engaged in a heated argument with others.

Since the early 2000s, researchers in the field of image retrieval have been developing context recognition systems using machine learning and statistical modeling~\cite{wang2002learning,chen2006machine,datta2008image}. With the advent of deep learning and the widespread use of modern graphics processing units (GPUs) or AI processors, accurate object annotation from images has become more feasible. Several deep learning-based approaches have been proposed to leverage context information to enhance basic emotion recognition networks~\cite{kosti2019context,lee2019context,mittal2020emoticon}. {Sections~\ref{sssec:EEContextualFeature} and~\ref{sec:integrated} provide more details.} 

{In addition to contextual information, the functional purpose of a person's movement can also provide valuable insights when inferring emotions. Movement is a combination of both functional and emotional expression, and thus, emotion recognition systems must be able to differentiate between movement that serves a functional purpose and movement that expresses emotions. Action recognition~\cite{carreira2017quo}, an actively studied topic in computer vision, has the potential to provide information on the function of a person's movement and assist in disentangling functional and emotional components of movement.}

\subsection{Acted Portrayals of Emotion}\label{sec:acted}

{Research on emotions has often turned to acted portrayals of emotion to overcome some of the challenges inherent in accessing adequate datasets, particularly because evoking intense, authentic emotions in a laboratory can be problematic both ethically~\cite{banziger2007using} and practically. This is because, as noted in Section~\ref{sec:psych}, emotional responses vary. Datasets relying upon actors have been useful in overcoming challenges of obtaining ample, adequately labeled emotion-expression data because compared to unscripted behavior (``in the wild''), where emotion expression often appears in blended or complex ways, actors are typically serving a narrative, in which emotion is part of the story. In addition, sampling emotional expression in the wild encounters cultural distinctions for the expressions themselves, as well as social norms for emotion expression or its repression~\cite{mesquita2016cultural}, which may also be culturally scripted for gender. Thus, researchers interested in emotional expressivity and nonverbal communication of emotion often turn to trained actors~\cite{banziger2007using} both to generate new datasets of emotionally expressive movement (e.g.,~\cite{atkinson2004emotion}) and for the sampling of emotion expression (e.g.,~\cite{luo2020arbee}). 
Such datasets are useful because actors coordinate all three channels of emotions expression, namely vocal, facial, and body, to produce natural or authentic-seeming expressions of emotions. 
Some researchers have validated the perception of actor-based and expert movement-based datasets in the lab by showing them to lay observers~\cite{atkinson2004emotion,melzer2019we}. This approach also entails problems, in that whereas it may capture norms of emotion expression and its clear communication, it may miss distinctions related to demographics such as gender~\cite{hutson2002gender}, ethnolinguistic heritage, individual, or generational norms~\cite{chaplin2015gender,cordaro2018universals}. According to cultural dialect theory, we all have non-verbal emotion accents~\cite{cordaro2018universals}, meaning 
emotion is expressed differently by different people in different regions, from different cultures. Only some of those cultural dialects appear when sampling films. Such films have often been edited so that viewers beyond that cultural dialect can ``read'' the emotional expression central to the narrative. Nuance in nonverbal dialects may be excluded in favor of general appeal.
}

{
Yet, an advantage of generating datasets from actors or other trained expressive movers is that ground truth can be better established. The intention of the emotion expressed, the emotion felt, and later the emotion perceived from viewing can all be assessed when generating the dataset. Likewise, because actors coordinate image, voice, and movement in the service of storytelling, the context and purpose are clarified, and thus many of the multiple expressive modes can be organized into individual emotion expression~\cite{jurgens2015effect,keltner2019emotional}. Moreover, because performing arts productions, such as movies and films made of theatre, music, and dance performances, integrate multiple communication modes, the creative team collaborates frequently about the emotional tone or intent of each work, coordinating lighting, scenery/background, objects, camerawork, and sound, with the performers (actors, dancers, musicians). While the team articulates their intentions during the creative process, the resulting produced art often resonates with wider variation to different audiences, according to their perceptions and tastes. 
}

{
For researchers relying upon acted emotions, it may be helpful to understand some ways in which actor training considers the role of emotions in theatre and film~\cite{hetzler2007actors}. The role of emotion in narrative arts may reflect some of the theories about the role of emotion itself~\cite{semmer2016disentangling}--to approximately inform the character (organism) about their needs, in order to drive them to take action to meet their needs, or to provide feedback on how recent actions meet or do not meet their needs. 
As actors prepare, they identify a character's needs (called their objective) and are moved by emotion to drive the character's action to overcome obstacles as they pursue the character's objective. Thus, when collecting data from acted examples, emotion expression can often be found preceding action in the narrative, or in response to it. Actors are also trained to listen with their whole being to their scene partner, to ``hold space'' during dialogue for emotion expression to fully unfold and complete its purpose of moving either the speaker or the listener to response or action. This expectation is particularly true in opera or other musical theatre genres, which often extend the time for emotion expression. 
}

{
In terms of 3-D modeling, actors trained for theatre not only highly develop the specific emotion-related action tendencies of the body but also consider viewer perception from 3D, for example, when performing on a thrust stage or theatre-in-the-round. Thus, their emotion expression may be more easily picked up by systems marking parts of the body during movement. Because bodily emotion expression is so crucial to a narrative, an important application of this field might be to automate audio description of emotion expression in film for the visually impaired or audio description of movement features salient to emotion expression.
}

{
Whereas we often do not recognize subtle emotion expression in strangers as well as we can in those we know, when actors portray characters within the circumstances of a play, the audience, gradually over the arc of the play, comes to perceive the emotional expression of each character as revealed over time. Understanding how this works in art can help us develop systems that take the time to learn and better understand emotion expression in diverse contexts and individuals, similar to how current voice recognition learns individual accents.
}

\subsection{Cultural and Gender Dialects}\label{sec:culture}
In a meta-analysis examining cultural dialects in the nonverbal expression of emotion, measurable variations were found in response to facial expression across cultures~\cite{elfenbein2002universality}. Moreover, it has long since been acknowledged that learned display rules~\cite{ekman_what_1997} and decoding rules~\cite{matsumoto1990cultural} vary across different cultures. For example, in Japanese culture, overt displays of negative emotional expression are discouraged. Further, smiling behavior in this context is often considered an attempt to conceal negative emotions. In contrast, in American culture, negative expressions are considered more appropriate to display~\cite{matsumoto1990cultural}.

%Reg
{Previous research has similarly demonstrated a powerful influence of gender-related emotion stereotypes in driving differential expectations for the type of emotions men and women are likely to experience and express. Men are expected to experience and express more power-related emotions like anger and contempt, whereas women are expected to experience and express more powerless emotions like sadness and fear~\cite{adams2015intersection}. These findings match self-reported emotional experiences and expressions with strong cultural consistency across 37 different countries worldwide~\cite{fischer2004gender}. Furthermore, there are even differences in the extent to which neutral facial appearance resembles different emotions, with male faces physically resembling anger expressions more (low, hooded brows, thin lips, angular features) and female faces resembling fear expressions more~\cite{adams2022angry}. 
}

While cultural norms affect how and whether emotions are displayed, such norms also influence how displays of emotion are perceived. For example, when a culture's norm is to express emotion, less intensity is read into such displays, whereas, in cultures where the norm is not to express emotion intensely, the same displays are read as more intense~\cite{matsumoto1999american}. In this way, visual percepts derived from objective stimulus characteristics can generate different subjective experiences based on culture and other individual factors. For instance, there is notable cultural variation in the extent to which basic visual information, such as background versus foreground, is integrated into observers' perceptual experiences~\cite{park2010perceiving}.

Culture and gender add complexity to human emotional expression, yet little research to date has examined individual variation in responses to visual scenes, either in terms of basic aesthetics or the emotional responses people have. Future work assessing simple demographic details (e.g., gender, age) will begin to explore this important source of variation. 

\subsection{Structure}\label{sec:structure}
Some basic visual properties have been found that characterize positive versus negative experiences and preferences. Most notably, round features--whether represented in faces or objects--elicit feelings of positivity and warmth and tend to be preferred over angular visual features. This preference has been used to explain the roundness of smiling faces, and the angularity of anger displays~\cite{aronoff1992stimuli}. Such visual curvature has also been found to influence attitudes toward, and preference for even meaningless patterns represented in abstract visual designs~\cite{bar2006very}. The connection between affective responses and these basic visual forms has helped computer vision predict emotions evoked from pictorial scenes, as mentioned earlier~\cite{lu2012shape}. 

Importantly, the dimensional approach to assessing visual properties underlying emotional experience can be used to examine both visual scenes and faces found within those scenes. Indeed, the dimensional approach adequately captures both ``pure'' and mixed emotional facial expressions~\cite{arya2009perceptually}, as well as affective responses to visual scenes, as demonstrated by the IAPS. Critically, even neutral displays have been found to elicit strong spontaneous impressions~\cite{van1999spontaneous}, ones that are effortless, nonreflective, and highly consensual. Recent research utilizing computer-based models suggests that these inferences are largely driven by the same physical properties found in emotional expressions. For instance, Said and colleagues~\cite{said2009structural} employed a Bayesian network trained to detect expressions in faces, and then applied this analysis to images of neutral faces that had been rated on a number of common personality traits. The results showed that the trait ratings of faces were meaningfully associated with the perceptual resemblances that these ``neutral'' faces had with emotional expressions. Thus, these results speak to a mechanism of perceptual overlap whereby expression and identity cues can both trigger similar emotional responses. 

A reverse engineering model of emotional inferences has suggested that perceptions of stable personality traits can be derived from emotional expressions as well~\cite{hareli2010emotional}. This work implicates appraisal theory as a primary mechanism by which observing facial expressions can inform stable personality inferences made of others. This account suggests that people use appraisals that are associated with specific emotions to reconstruct inferences of others' underlying motives, intents, and personal dispositions, which they then use to derive stable impressions. It has likewise been shown that emotion-resembling features in an otherwise neutral face can drive person perception~\cite{adams2015intersection}. Finally, research has also suggested that facial expressions actually take on the form that they do to resemble static facial appearance cues associated with certain functional affordance, such as facial maturity~\cite{marsh2005fear}, and gender-related appearance~\cite{adams2015intersection}. 

\subsection{Personality}\label{sec:personality}

{Personality describes an individual's relatively stable patterns of thinking, feeling, and behaving. These are characterized by certain personality traits which represent a person's disposition towards the world. Emotions, on the other hand, are the consequence of the individual's response to external or internal stimuli. They may change when the stimuli change, and therefore are considered as states (as opposed to traits). Just as a full-blown emotion represents an integration of feeling, action, appraisal, and wants at a particular time and location, personality represents integration of these components over time and space. Researchers have tried to characterize the relationships between personality and emotions. Several studies found correlations between certain personality traits and specific emotions. For instance, the trait neuroticism was found to be correlated with negative emotions, while extroverted people were found to experience higher levels of positive emotions than introverted people~\cite{donovan2020quantifying}. These correlations were explained by relating different personality traits to specific emotion regulation strategies e.g.,~\cite{hughes2020personality} or by demonstrating that evaluation mediates between certain personality traits and negative or positive affect~\cite{fossum2000distinguishing}. Whatever the reason for these correlations is, the fact that they are correlated might help to create state-to-trait inferences. Therefore, if emotional states can be mapped to personality, the ability to automatically recognize emotions could provide tools for automatic detection of personality.}

\begin{figure}[ht!]
\centering
\includegraphics[width=0.35\textwidth]{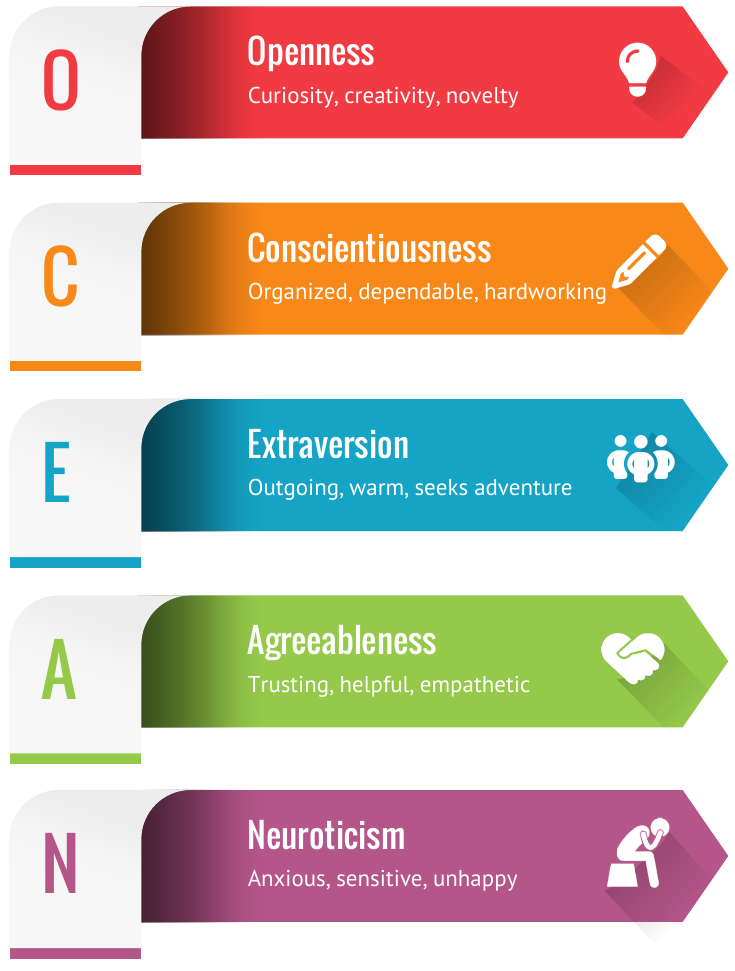}
\caption{The Big Five personality traits, or OCEAN.}
\label{fig:OCEAN}
\end{figure}

{Advances in computationally, data-driven methods offer promising strides toward personality traits being predicted based on a dataset of individuals' behavior, self-reported personality traits, or physiological measures. It is also possible to use factor analyses to identify underlying dimensions of personality, such as the Big Five personality traits, or OCEAN (openness, conscientiousness, extraversion, agreeableness, and neuroticism)~\cite{digman1990personality,goldberg1993structure} (Fig.~\ref{fig:OCEAN}). Personality can likewise be inferred from facial expressions of emotion~\cite{hareli2010emotional} and even from emotion-resembling facial appearance~\cite{adams2012emotion}. Machine learning algorithms have recently been successfully employed to predict human perceptions of personality traits based on such facial emotion cues~\cite{albohn2021expressive}. Further, network analysis, such as social network analysis, can also be incorporated to identify patterns of connectivity between different personality traits or behavioral measures. Finally, interpreting a person's emotional expression might also depend on the perception of their personality, when additional data is available for inferring personality~\cite{friedman1980personality}. For further information, readers can refer to recent research in this area, e.g.,~\cite{reisenzein2020personality,azucar2018predicting}.}

\subsection{Affective Style}\label{sec:style}
Affective styles driven by the tendencies to approach reward versus avoid punishments  found their way into early conflict theories~\cite{lewin1938conceptual} and remain a mainstay of contemporary psychology in theories such as Carver and Scheier's Control Theory~\cite{carver1982control} and Higgins's Self-discrepancy Theory~\cite{higgins1989self}.

Evidence for the specific association between emotion and approach-avoidance motivation has largely involved the examination of differential hemispheric asymmetries in cortical activation.  Greater right frontal activation has been associated with avoidance motivation as well as with flattened positive affect, and increased negative affect. Greater left frontal activation has been associated with approach motivation and positive affect~\cite{davidson1998affective}. Supporting the meaningfulness of these findings, Davidson argued that projections from the mesolimbic reward system, including basal ganglia and ventral striatum, which are associated with dopamine release, give rise to greater left frontal activation. Projections from the amygdala associated with the release of the primary vigilance-related transmitter norepinephrine give rise to greater right frontal activation. 

Further evidence supporting the emotion/behavior orientation link stems from evidence accumulated in studies using measures of behavioral motivation based on Gray's~\cite{gray1987neuropsychology} proposed emotion systems. The most widely studied of these are the Behavioral Activation System (BAS) and the Behavioral Inhibition System (BIS). The BAS is argued to be highly related to appetitive or approach-oriented behavior in response to reward, whereas the BIS is argued to be related to inhibited or avoidance-oriented behavior in response to punishment. Carver and White~\cite{carver1994behavioral} developed a BIS/BAS self-report rating measure that is thought to tap into these fundamental behavioral dispositions. They found that extreme scores on BIS/BAS scales were linked to behavioral sensitivity toward reward versus punishment contingencies, respectively~\cite{carver1994behavioral}. BIS/BAS measures have been shown to be related to emotional predisposition, with positive emotionality being related to the dominance of BAS over BIS and depressiveness and fearful anxiety being related to the dominance of BIS over BAS~\cite{davidson2003asymmetrical}. 
 
Notably, for many years there existed a valence (positive/negative) versus motivational (approach/avoidance) confound in all work conducted in the emotion/behavior domain. Negative emotions were associated with avoidance-oriented and positive emotions with approach-oriented behavior, a contention supported by much of the work reviewed above. The valence/motivation confound led researcher Harmon-Jones and colleagues to test for hemispheric asymmetries in activation associated with anger, a negative emotion with an approach orientation (aggression)~\cite{harmon1998anger}. They argued that if left hemispheric lateralization was associated with anger this would indicate that the hemispheric lateralization in activation previously found was in fact due to behavioral motivation. However, right hemispheric lateralization would indicate that they were due to valence. In these studies, they found that dispositional anger~\cite{harmon1998anger} was associated with left lateralized EEG activation, consistent with the first interpretation and with that previously reported only for positive emotion. They supported this conclusion by showing that the dominance of BAS over BIS was associated with anger~\cite{harmon2003clarifying}.

\section{Emotion Recognition: Key Ideas and Systems}\label{sec:survey}

The field of computer-based emotion recognition from visual media is nascent, but has seen encouraging developments in the past two decades. Interest in this area has also grown sharply (Section~\ref{sec:growth}). We will highlight some existing research ideas and systems related to modeling evoked emotion (Section~\ref{sec:evoked}), facial expression (Section~\ref{sec:facial}), and bodily expression (Section~\ref{sec:body}). In addition, we will discuss integrated approaches to emotion recognition (Sections~\ref{sec:integrated} and~\ref{sec:multimodal}). Because of the breadth of the field, it is not possible to cover all exciting developments, so we will focus our review on the evolution of the field and some of the most current, cutting-edge results. 

\subsection{Exponential Growth of the Field}\label{sec:growth}

\begin{figure}[ht!]
\centering
\includegraphics[width=0.98\linewidth]{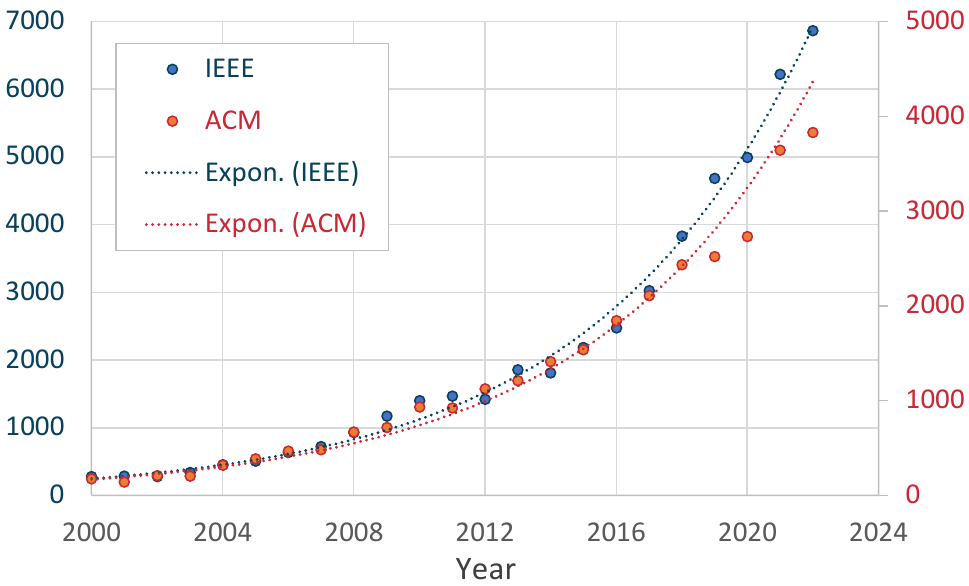}
\caption{An exponential growth in the number of IEEE and ACM publications related to emotion in images and videos in the recent two decades was observed. Statistics are based on querying the IEEE and ACM publication databases. The lower rate of expansion observed in the year 2020 can be attributed to the disruptions caused by the COVID-19 pandemic and the limitations it imposed on human subject studies.}
\label{fig:trend}
\end{figure}

To gain insight into the growing interest of the IEEE and computing communities in emotion recognition research, we conducted a survey of publications in the IEEE Xplore and ACM Digital Library (DL) databases. Results revealed an exponential increase in the number of publications related to emotion or sentiment in images and videos over the last two decades (Fig.~\ref{fig:trend}). 

As of February 2023, a total of 48,154 publications were found in IEEE Xplore, with the majority (35,247 or 73.2\%) being in conference proceedings, followed by journals (8,214 or 17.1\%), books (2,400 or 5.0\%), magazines (1,413 or 2.9\%), and early access journal articles (831 or 1.7\%).
The field has experienced substantial growth, with a 25-fold increase during the period from the early 2000s to 2022, rising from an average of 275 publications per year to about 7,000 per year in 2022.

In the ACM DL, a total of 30,185 publications were found,  with conference proceedings making up the majority (24,817 or 82.2\%), followed by journals (3,851 or 12.8\%), magazines (780 or 2.6\%), newsletters (505 or 1.7\%), and books (264 or 0.9\%).
In the ACM community, the field has seen a 22-fold growth during the same period, with an average of 170 publications per year in the early 2000s rising to about 3,800 per year in 2022.
A baseline search indicated that the field related to images and videos had a roughly linear, 6-fold growth during the same period. Emotion-related research accounted for 14.2\% of image- and video-related publications. The growth in emotion-related research outpaced the baseline growth by a significant margin, suggesting that it has a higher future potential.
The annual growth rate for the field of emotion in images and videos is 15-16\%. If this growth continues, the annual number of publications in this field is expected to double every five years.

\subsection{Modeling Evoked Emotion}\label{sec:evoked}

\begin{figure}[!ht]
\begin{center}
\includegraphics[trim=50 40 600 0,clip,keepaspectratio,width=\linewidth]{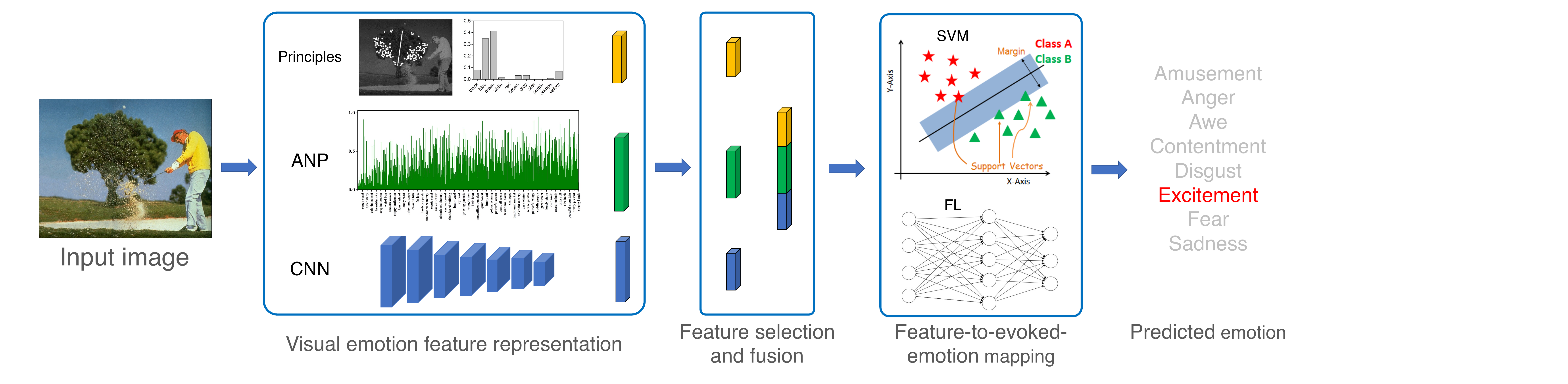}
\caption{A generalized framework for evoked emotion prediction from images.}
\label{fig:EvokedFramework1}
\end{center}
\end{figure}

\begin{figure}[!ht]
\begin{center}
\includegraphics[trim=80 70 700 0,clip,keepaspectratio,width=\linewidth]{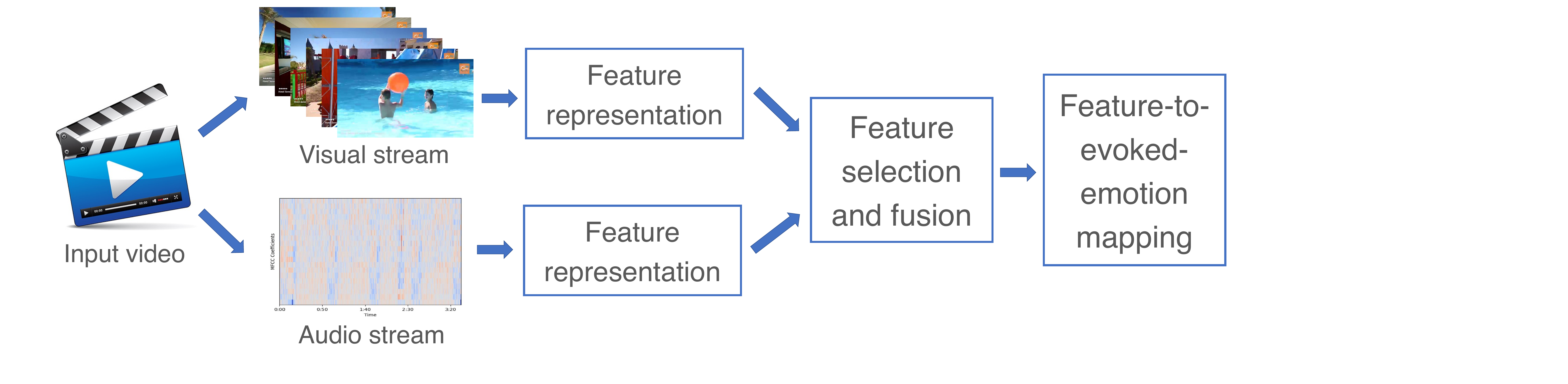}
\caption{A generalized framework for evoked emotion prediction from videos.}
\label{fig:EvokedFramework2}
\end{center}
\end{figure}

Evoked emotions can be inferred when viewers see an image or a video, either based on changes in their physical body or the stimuli being viewed, i.e. explicit affective cues or implicit affective stimuli~\cite{melzer2019we,zhao2021emotion,wang2015video}. In this section, we will focus on implicit affective stimuli, particularly images and videos. Similar to other machine learning and computer vision tasks~\cite{joshi2011aesthetics,zhao2021affective}, an evoked emotion recognition system from images or videos typically consists of three components: emotion feature extraction, feature selection and fusion, and feature-to-evoked-emotion mapping. The generalized frameworks for this process are illustrated in Figs.~\ref{fig:EvokedFramework1} and~\ref{fig:EvokedFramework2}. The first step is to extract emotion features from the original images and videos, typically after pre-processing, and converting them to numerical representations that are easier to process. Feature selection and fusion aim to select discriminative features that are more relevant to emotions, reduce the dimensionality of features, and combine different types of features into a unified representation. Finally, each unified representation is classified. Feature-to-evoked-emotion mapping learns a classifier to project the feature representation to specific evoked emotions.

\subsubsection{Visual Emotion Feature Representation}
\label{sssec:EEVisualFeature}

Prior to the emergence of deep learning, visual emotion features were primarily developed manually, drawing inspiration from various interrelated disciplines including computer vision, art theory, psychology, and aesthetics.

One straightforward approach employs existing {\it low-level vision features} as representations of emotion. \citeauthor{yanulevskaya2008emotional}~\cite{yanulevskaya2008emotional} extracted holistic Wiccest and Gabor features. Other researchers have employed scale-invariant feature transform (SIFT), histogram of oriented gradients (HOG), self-similarities, GIST, and local binary patterns (LBP) to extract low-level vision representations of emotion for video keyframes~\cite{jiang2014predicting,pang2015deep}. \citeauthor{yang2018retrieving}~\cite{yang2018retrieving} compared their method with multiple low-level vision features, including SIFT, HOG, and Gabor. 
\citeauthor{rao2016multi}~\cite{rao2016multi} first employed bag-of-visual-words of SIFT features for each image block from multiple scales and then represented each block by extracting latent topics based on probabilistic latent semantic analysis. While such low-level vision representations are still used, such as LBP and optical flows~\cite{muszynski2021recognizing}, they do not effectively bridge the semantic gap and cannot bridge the more challenging affective gap~\cite{zhao2021affective}.

More robust approaches to exploring visual emotion features employ {\it theories on art and aesthetics} to design and develop features.
Art is often created with the intention of evoking emotional reactions from the audience. 
As Pablo Picasso claims, ``Art washes away from the soul the dust of everyday life,'' and as Paul C\'ezanne asserts, ``A work of art which did not begin in emotion is not art,''~\cite{machajdik2010affective,zhao2014exploring,achlioptas2021artemis}. Generally, art theory includes elements of art and principles of art. The elements of art, such as color, line, texture, shape, and form, serve as building blocks in creating artwork. Meanwhile, the principles of art correspond to the rules, tools, or guidelines for arranging and orchestrating the elements of art, such as balance, emphasis, harmony, variety, and gradation. Among these elements, color is the most commonly used artistic feature~\cite{machajdik2010affective,peng2015mixed,baveye2015liris,zhao2017continuous,zhao2018predicting,zhao2020discrete,muszynski2021recognizing,song2022have}, followed by texture. \citeauthor{lu2012shape}~\cite{lu2012shape} investigated the relationship between shape and emotions through an in-depth statistical analysis. \citeauthor{zhao2014exploring}~\cite{zhao2014exploring} systematically formulated and implemented six artistic principles except for rhythm and proportion and combined them into a unified representation. 

Aesthetics is widely acknowledged to have a strong correlation with emotion~\cite{joshi2011aesthetics,wang2013interpretable}. Artwork that is aesthetically designed tends to attract viewers' attention and create a sense of immersion. As early as 2006, \citeauthor{datta2006studying} developed computer algorithms to predict the aesthetic quality of images~\cite{datta2006studying}. Among the various  aesthetic features, composition, such as the rule of thirds, has been the most popular~\cite{datta2006studying,machajdik2010affective,wang2013interpretable,peng2015mixed,baveye2015liris,liu2019affective}. The figure-ground relationship, which refers to cognitive feasibility in distinguishing the foreground and the background, and several other aesthetic features were designed~\cite{wang2013interpretable}. \citeauthor{sartori2015s}~\cite{sartori2015s} to analyze the influence of different color combinations on evoking binary sentiment from abstract paintings. Artistic principles and aesthetics were used to organize artistic elements from different but correlated perspectives, which were sometimes not clearly differentiated and extracted together~\cite{wang2013interpretable,liu2019affective}. By considering  relationships among different elements, artistic principles and aesthetics have been demonstrated to be more interpretable, robust, and accurate than artistic elements and low-level vision representations for recognizing evoked emotions~\cite{wang2013interpretable,zhao2014exploring}.

To bridge the gap between low-level visual features and high-level semantics, an {\it intermediate level of attributes and characteristics} was designed. These intermediate attributes and characteristics were then applied to the prediction of evoked emotions~\cite{yuan2013sentribute,zhao2017continuous,zhao2018predicting,zhao2020discrete}. For example, in addition to generic scene attributes, eigenface-based facial expressions were also considered~\cite{yuan2013sentribute}. These attributes performed better than low-level vision representations, but the interpretability was still limited.

{\it High-level content and concept} play an essential role in evoking emotions for images with obvious semantics, such as natural images. The number and size of faces and skins contained in an image have been used as an early and simple content representation~\cite{machajdik2010affective}. Facial expressions of  images containing faces are a direct cue for viewers to produce emotional reactions and are therefore often employed as content representation~\cite{zhao2017continuous,zhao2018predicting}. \citeauthor{jiang2014predicting}~\cite{jiang2014predicting} developed a method that involved detecting objects and scenes from keyframes. Based on the observation that general nouns like ``baby'' were detectable but had a weak link to emotion, whereas adjectives like ``crying'' could provide strong emotional information but were difficult to detect, \citeauthor{borth2013large}~\cite{borth2013large} introduced adjective noun pairs (ANPs) by adding an adjective before a noun, such as ``crying baby.'' The combination enabled strong ability to map concepts to emotions while remaining detectable. A large visual sentiment ontology named SentiBank was proposed to detect the probability of 1,200 ANPs. Thus, as a milestone, SentiBank cannot be ignored as a baseline in performance evaluation, even in the current deep learning era. A multilingual visual sentiment ontology (MVSO)~\cite{jou2015visual} was later extended to deal with different cultures and languages. About 16K sentiment-biased visual concepts across 12 languages are constructed and detected. ANP representations are widely used as high-level semantic representations~\cite{jiang2014predicting,zhao2017continuous,acar2017comprehensive,zhao2018predicting,zhao2020discrete}. These content and concept representations achieve the best performance for the images containing such semantics but fail for abstract paintings.

Fortunately, the rise of deep and very deep Convolutional Neural Networks (CNNs) in image classification has led to deep learning becoming the primary learning strategy in various fields, including computer vision and natural language processing. This is also the case for evoked emotion modeling~\cite{zhao2021affective}. Given sufficient annotated data, deep learning models can be trained to achieve superior performance on various types of images, including natural and social photographs, as well as abstract and artistic paintings. In many cases, features are automatically learned without the need for manual crafting.

Recently, {\it global representation at the image level} has been demonstrated to hold promise for evoked emotion analysis.
One approach to extract deep features is to directly apply pretrained CNNs, such as AlexNet~\cite{krizhevsky2012imagenet}, VGGNet~\cite{simonyan2015very}, and ResNet~\cite{he2016deep}, to the given images and obtain responses of the last (few) fully connected (FC) layers~\cite{Xu2014Visual,peng2015mixed,you2016building,acar2017comprehensive,zhang2018recognition,liu2019affective,zhao2020discrete,zhao2020end,jin2021visual,wei2021user,shukla2022recognition}. Other methods have begun to use the output of a transformer's encoder as visual reorientation~\cite{deng2022simple,pan2022representation}. \citeauthor{Xu2014Visual}~\cite{Xu2014Visual} demonstrated that the deep features from the last but one FC layer outperformed those from the last FC layer. The extracted CNN features were transformed to another kernelized space via discrete Fourier transform and sparse representation to suppress the impact of noise in videos~\cite{zhang2018recognition}. 
{\citeauthor{chen2016emotion}~\cite{chen2016emotion} first obtained event, object, and scene scores with state-of-the-art pretrained detectors based on deep neural networks. They then integrated these into a context fusion network to generate a unified representation.} More recently, \citeauthor{song2022have}~\cite{song2022have} used the pretrained MS Azure Cognitive Service API to extract objects contained in an image and corresponding confidence scores as visual hints, which were then transformed to TF-IDF representation.
This pretrained deep feature can be essentially viewed as another handcrafted feature since the classifier used for the final emotion prediction needs to be separately trained. To enable end-to-end training, fine-tuning is widely employed to adjust the deep features to be more correlated with evoked emotions~\cite{you2015robust,peng2015mixed,you2016building,campos2017pixels}. Other inspiring improvements include multitask learning~\cite{yang2017joint,yang2018retrieving} and emotion correlation mining~\cite{yang2021circular,yang2021stimuli,liang2022chain} to better learn emotion-specific deep features. The image-level global representation extracts deep features from the global aspect taking the original image as input. To deal with the temporal correlations among successive frames in videos, 3-D CNN (C3D)~\cite{tran2015learning} is adopted taking a series of frames as input~\cite{ou2021multimodal}. Although contextual information is considered, the global representation treats local regions equally without considering their importance.

Strategies such as attention~\cite{you2017visual,song2018boosting,fan2018emotional,zhao2019pdanet} and sentiment maps~\cite{yang2018weakly,li2022weakly} have been widely used to extract {\it region-level representations} in order to emphasize the importance of local regions in evoked emotion prediction. Spatial attention is used to determine the correspondence between local image regions and detected textual visual attributes~\cite{you2017visual}. Besides operating on the global representation to obtain attended local representation, the spatial attention map is enforced to be consistent with prior knowledge contained in the detected saliency map~\cite{song2018boosting}. Both spatial and channel-wise attentions are considered to reflect the importance of the spatial local regions and the interdependency
between different channels~\cite{zhao2019pdanet}. In contrast, \citeauthor{fan2018emotional}~\cite{fan2018emotional} investigated how image emotion could be used to predict human attention and found that emotional content could strongly and briefly attract human visual attention. \citeauthor{yang2018weakly}~\cite{yang2018weakly} designed a weakly supervised coupled CNN to explore local information by detecting a sentiment-specific map using a cross-spatial pooling strategy, which only required image-level labels. The holistic and local representations were combined by coupling the sentiment map. To address the issue of overemphasis on local information and neglect of some discriminative sentiment regions, a discriminate enhancement map was recently constructed by spatial weighting and channel weighting~\cite{li2022weakly}. Both the discriminate enhancement map and sentiment map were coupled with the global representation.

In recent years, a growing body of research has focused on developing effective {\it multilevel representations}~\cite{zhu2017dependency,rao2019multi,rao2020learning,yang2021stimuli,yang2021solver,zhang2022multiscale,xu2022mdan,zhang2022image,xu2019video,sun2019gla,cheng2021context}. One strategy has been to view different CNN layers as different levels~\cite{zhu2017dependency,rao2020learning,xu2022mdan}. Multiple levels of features were extracted at different branches (four~\cite{rao2020learning} and four plus one main~\cite{zhu2017dependency} branches). The features from different levels were then integrated by a Bidirectional Gated Recurrent Unit~\cite{zhu2017dependency} or a fusion layer with basic operations like the mean. 
These two methods both claim that features from global to local levels can be extracted at different layers, but the correspondence between them is still unclear. This issue is partially addressed in the multilevel dependent attention network (MDAN)~\cite{xu2022mdan}. Based on the assumption that different semantic levels and affective levels are correlated, affective semantic mapping disentangles the affective gap by one-to-one mapping between semantic and affective levels. Besides the global-level learning at the highest affective level with emotion hierarchy preserved, local learning is incorporated at each semantic level to differentiate
among emotions at the corresponding affective level~\cite{xu2022mdan}.
Further, multihead cross channel attention and level-dependent class activation maps are designed to model level-wise channel dependencies and  spatial attention within each semantic level, respectively. Further research is needed to explore more effective ways to map semantic and affective levels, particularly when the number of levels varies significantly.

Another strategy for extracting multilevel representations involves utilizing object, saliency, and affective region detection~\cite{yang2021stimuli,yang2021solver,xu2019video,sun2019gla,cheng2021context,rao2019multi,zhang2022image}. Inspired by the Stimuli-Organism-Response (S-O-R) emotion model in psychology, \citeauthor{yang2021stimuli}~\cite{yang2021stimuli} selected specific emotional stimuli, including image-level color and region-level objects and faces, using off-the-shelf detection models. Corresponding to the selected stimuli, three specific networks were designed to extract the features, including CNN-based color and other global representations, Long-Short Term Memory (LSTM)-based semantic correlations between different objects, and CNN-based facial expressions. Besides the correlations between different objects based on graph
convolutional network (GCN), \citeauthor{yang2021solver}~\cite{yang2021solver} also mined the correlations between scenes and objects using a scene-based attention mechanism, motivated by the assumption that scenes guide objects to evoke distinct emotions. A similar approach was applied by \citeauthor{cheng2021context}, but with the input to the object detector being the extracted temporal-spatial features by C3D~\cite{cheng2021context}.
These methods treat objects, typically detected by Faster R-CNN, as regions, and reflect the importance of different regions through attention~\cite{yang2021stimuli} or graph reasoning~\cite{yang2021solver,cheng2021context}. Objects and faces were also respectively detected in~\cite{xu2019video} and~\cite{sun2019gla}, and corresponding CNN features were extracted. \citeauthor{rao2019multi}~\cite{rao2019multi} employed an emotional region proposal method to select emotional regions and remove non-emotional ones. The region's emotion score was a combination of the probability of the region containing an object and that of the region evoking emotions. A similar approach for selecting emotionally charged regions was used by \citeauthor{zhang2022image}, but the emotion score was a weighted combination of two probabilities~\cite{zhang2022image}.

Efforts have been made to {\it combine handcrafted and deep representations} in order to take advantage of their complementary information. For example, \citeauthor{liu2019affective}~\cite{liu2019affective} combined them after feature dimension reduction, whereas \citeauthor{chen2014deepsentibank}~\cite{chen2014deepsentibank} used AlexNet-style deep CNNs~\cite{krizhevsky2012imagenet} to train ANP detectors and achieved improved ANP classification accuracy. The DeepSentiBank has also been used to extract keyframe features in videos~\cite{gu2020sentiment,wei2021user}. More recently, the correlations of content features based on pretrained CNN and color features from color moments have been taken into consideration~\cite{ruan2021color}. Specifically, the proposed cross-correlation model consists of two modules: an attention module to enhance color representation using content vectors and a convolution module to  enrich content representation using pooled color embedding matrices. Further research is needed to investigate the potential of this approach to interact between handcrafted and deep representations.

\subsubsection{Audio Emotion Feature Representation}
\label{sssec:EEAudioFeature}

The audio features used in video-based evoked emotion analysis are often sourced from the fields of acoustics, speech, and music signal processing~\cite{wang2015video}. Similar to visual features, audio features for emotion analysis can be divided into two categories: handcrafted and deep learning-based. 

Among {\it handcrafted feature representations}, the Mel-frequency cepstral coefficients (MFCC) are commonly adopted features~\cite{jiang2014predicting,baveye2015liris,pang2015deep}. Energy entropy, signal energy, zero crossing rate, spectral rolloff, spectral centroid, and spectral flux are also employed~\cite{jiang2014predicting,pang2015deep}. The openSMILE toolkit is a popular option for extracting handcrafted audio features~\cite{eyben2010opensmile}. \citeauthor{el2011survey}~\cite{el2011survey} classified speech features into four groups: continuous, qualitative, spectral, and Teager
energy operator-based. \citeauthor{panda2020audio}~\cite{panda2020audio} summarized music features into eight dimensions, including melody, harmony, rhythm, dynamics, tone color, expressivity, texture, and form. For more information on handcrafted audio feature representations, please refer to these papers.

{\it Learning-based deep audio feature representations}
mirror the learning-based strategy used in the visual stream in videos, and some studies have explored the use of CNN-based deep features for audio streams. One approach is to send the raw audio signal directly to a 1-dimensional CNN. However, a more popular method involves transforming the raw audio signal into 
MFCC, which can be viewed as an image, before feeding it into a 2-D CNN~\cite{acar2017comprehensive,sun2019gla,zhao2020end,cheng2021context,ou2021multimodal,shukla2022recognition,pan2022representation}. When simultaneously considering the MFCC from multiple video segments, 3-D CNN or LSTM can be utilized. For further information on modeling the temporal information in videos, see Section~\ref{sssec:EEFeatureFusion}.

\subsubsection{Contextual Feature Representation}
\label{sssec:EEContextualFeature}

{The context information that is important in evoked emotion prediction can be divided into two categories: within visual media and across modalities. For context within visual media, one common approach is to extract features from different levels that correspond to semantics at different scales, such as the global and local levels. For more details, see Section~\ref {sssec:EEVisualFeature}. When considering context across modalities, the social context of users, including their common interest groups, contact lists, and similarity of comments to the same images, is taken into account in personalized image emotion recognition~\cite{zhao2018predicting}. In addition to the visual and audio representations, motion representation is also considered~\cite{pan2022representation}.}

\subsubsection{Feature Selection and Fusion}
\label{sssec:EEFeatureFusion}
As previously mentioned, features can be extracted within a single modality (e.g., image) or across multiple modalities. On the one hand, different types of features can have varying discriminative abilities. High-dimensional features may suppress the power of low-dimensional ones when fusing them together. High-dimensional features may cause a ``curse of dimensionality'' and corresponding overfitting. On the other hand, fusing different features together can improve the performance of emotion analysis by jointly exploiting their representation ability and complementary information. Therefore, feature selection and fusion techniques are often used, particularly for datasets with a small number of training samples.

Before being processed by the feature extractor, an image or video often undergoes preprocessing. Resizing an image to a fixed spatial resolution is straightforward, but videos pose a greater challenge due to differences in both spatial and temporal structures. It is important to determine how to combine frame-level or segment-level features into a unified video-level representation. Below, we first explore commonly used temporal information modeling in videos, followed by a summary of feature selection and fusion techniques.

In the area of {\it temporal information modeling}, there is a need to divide the input video into segments and extract keyframes to be used for feature extraction. A straightforward method is to use all frames of the entire video~\cite{zhang2018recognition,xu2019video}, which ensures the least information loss but also results in high computational complexity. Common methods for segment sampling include fixed time interval~\cite{jiang2014predicting,pang2015deep,acar2017comprehensive,sun2019gla,shukla2022recognition}, fixed number of segments~\cite{zhao2020end,ou2021multimodal}, and video shot detection~\cite{cheng2021context,wei2021user}. Sampling keyframes within a segment can be done through fixed frame sampling (uniform sampling)~\cite{jiang2014predicting,pang2015deep,sun2019gla,shukla2022recognition}, random sampling~\cite{zhao2020end}, middle frame selection~\cite{acar2017comprehensive}, and mean histogram sampling (i.e., the frame with the closest histogram to the mean histogram)~\cite{baveye2015liris}.
Recently, researchers have developed dedicated keyframe selection algorithms~\cite{gu2020sentiment,wei2021user}. One such method uses low-rank and sparse representation with Laplacian matrix regularization in an unsupervised manner, considering both the global and local structures~\cite{gu2020sentiment}. Another method trains an image emotion recognition model on an additional image dataset to estimate the affective saliency of video frames~\cite{wei2021user}. Keyframes are selected based on the largest inter-frame difference of color histograms, after first sorting the segments (shots) based on affective saliency. Despite progress, effective and efficient selection of emotion-aware keyframes to enhance accuracy and speed remains an open challenge.

Another challenge is combining frame-level or segment-level features into a unified video-level representation. Some straightforward methods include average pooling~\cite{jiang2014predicting,pang2015deep,zhang2018recognition,xu2019video}, average pooling with temporal attention~\cite{zhao2020end}, max pooling~\cite{wei2021user}, bag-of-words quantization~\cite{jiang2014predicting,pang2015deep}, LSTM~\cite{cheng2021context,ou2021multimodal}, and gated recurrent units (GRUs)~\cite{sun2019gla}. Another approach is to perform emotion prediction at the segment level and then combine the results using methods such as majority voting~\cite{acar2017comprehensive,shukla2022recognition}.

{\it Feature selection}, especially for handcrafted features, is often employed when extracted features have high dimensions that contain redundant, noisy, or irrelevant information, the dimensions of different kinds of features to be fused vary significantly, or the size of the training samples is small. Some commonly used feature selection methods in evoked emotion modeling include cross-validation on the training set~\cite{yanulevskaya2008emotional}, wrapper-based selection~\cite{machajdik2010affective}, forward selection~\cite{lu2012shape}, and principal component analysis (PCA)-based selection~\cite{machajdik2010affective,lu2012shape,zhao2014exploring,you2016building}. 
Feature selection can be combined with feature fusion, particularly in the case of early fusion~\cite{zhao2014exploring,liu2019affective}. If dimensions of the features being fused are similar, feature selection is carried out after fusion~\cite{zhao2014exploring}. However, if dimensions are significantly different, feature selection is usually performed first~\cite{liu2019affective} to prevent an overemphasis on high-dimensional features.

The integration of multiple features within a single modality or different kinds of representations across different modalities
through {\it feature fusion} plays a crucial role in emotion recognition~\cite{zhao2021emotion}. Model-free fusion, which operates independently of learning algorithms, is widely used and encompasses early fusion~\cite{machajdik2010affective,wang2013interpretable,zhao2014exploring,sartori2015s,peng2015mixed,zhu2017dependency,fan2018emotional,liu2019affective,rao2019multi,yang2021stimuli,yang2021solver,zhang2022multiscale,song2022have,sun2019gla,zhao2020end}, late fusion~\cite{yuan2013sentribute,zhao2020discrete,ruan2021color,acar2017comprehensive,shukla2022recognition}, and hybrid fusion~\cite{wei2021user}. Hierarchical fusion that incorporates various feature sets at different levels of the hierarchy is also employed~\cite{muszynski2021recognizing}. 

Model-based fusion, however, is  performed explicitly during the learning process. Kernel-based fusion~\cite{jiang2014predicting,xu2019video} and graph-based fusion are often used for shallow learning models, whereas attention-based~\cite{zhang2022image,ou2021multimodal,pan2022representation}, neural network-based~\cite{pang2015deep,cheng2021context}, and tensor-based fusion strategies have been recently employed for deep learning models. 

A Transformer encoder with multihead self-attention layers is used in 
sentiment region correlation analysis to exploit dependencies between regions~\cite{zhang2022image}. The self-attention transforms original regional features into a higher-order representation between implied regions. \citeauthor{ou2021multimodal} used a local-global attention mechanism to explore the intrinsic relationship among different modalities and for different segments~\cite{ou2021multimodal}. The local attention evaluates the importance of different modalities in each segment, whereas the global attention captures the weight distribution of different segments. In contrast, \citeauthor{pang2015deep}~\cite{pang2015deep} employed the deep Boltzmann machine (DBM) to infer non-linear relationships among different features within and across
modalities by learning a jointly shared embedding space. \citeauthor{cheng2021context}~\cite{cheng2021context} proposed an adaptive gated multimodal fusion model, which first mapped features from different modalities to the same dimension and then employed a gated multimodal unit (GMU) to find an intermediate representation. For a comprehensive survey on feature fusion strategies, please refer to~\cite{zhao2021emotion}.

\begin{figure*}[!t]
\begin{center}
\begin{tabular}{ccc}
      \includegraphics[width=0.3\linewidth]{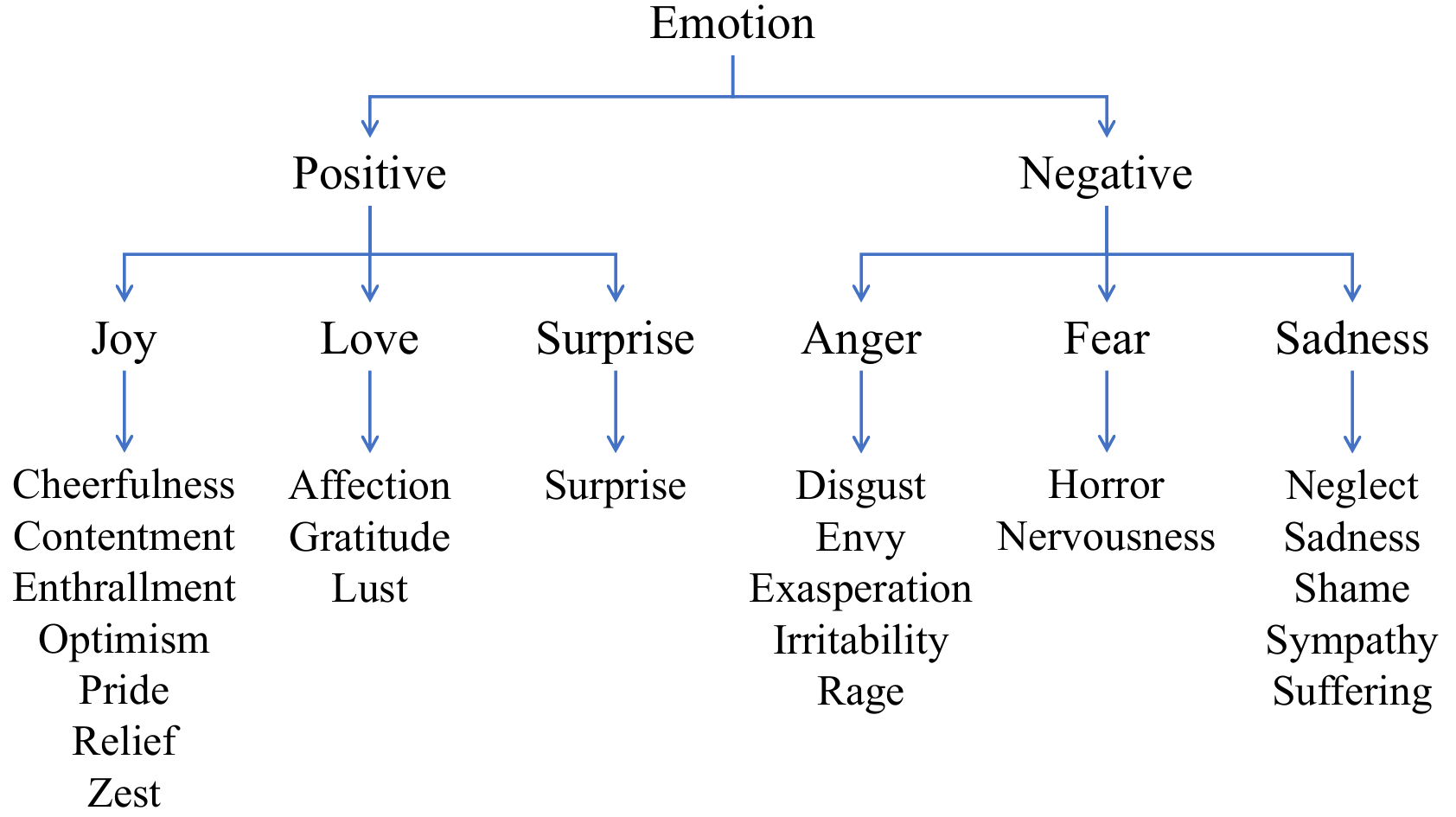} & \includegraphics[width=0.3\linewidth]{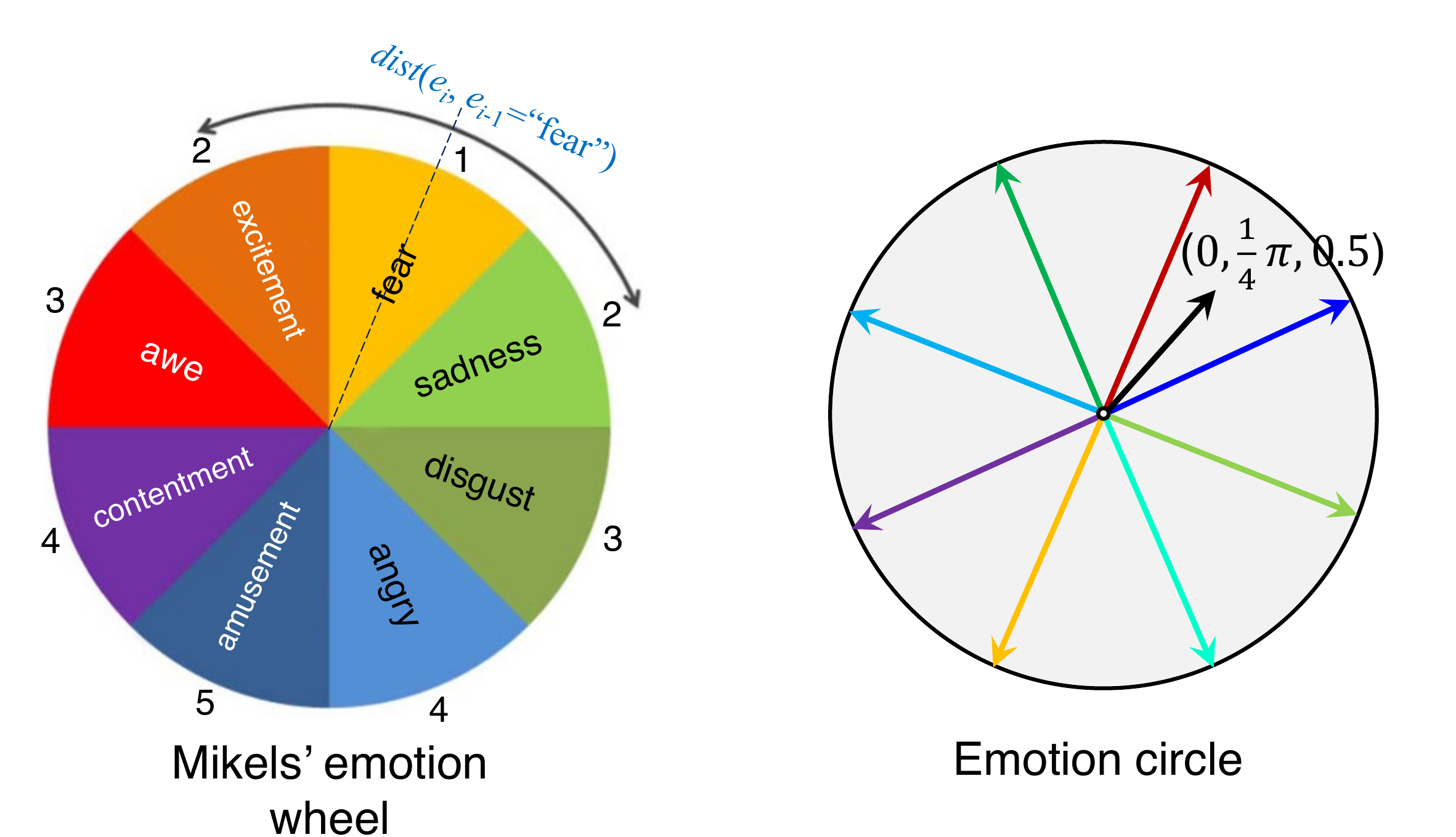} & \includegraphics[width=0.3\linewidth]{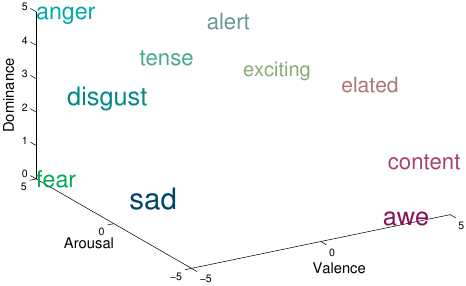}\\
(a) Emotion hierarchy & (b) Emotion similarity & (c) Emotion boundedness
\end{tabular}
\caption{Illustration of different emotion correlations. (a) Emotion hierarchy: a three-level hierarchy from sentiment to primary emotion and secondary emotion~\cite{xu2022mdan}. (b) Emotion similarity: Mikels' emotion wheel on the left~\cite{zhao2018predicting} and Emotion circle on the right~\cite{yang2021circular}. (c) Emotion boundedness: rough transformation between discrete emotion categories and continuous emotion values in the VAD space~\cite{lu2012shape}.}\label{fig:EmotionCorrleations}
\end{center}
\end{figure*}

\subsubsection{Feature-to-Evoked-Emotion Mapping}
\label{sssec:EEMapping}

Based on the emotion representation models discussed in Section~\ref{sec:define}, various evoked emotion analysis tasks can be undertaken, including classification, regression, retrieval, detection, and label distribution learning~\cite{zhao2021emotion}. The former three can also be classified into visual media-centric dominant emotion analyses and viewer-centric personalized emotion analyses, while the latter two are typically visual media-centric. After obtaining a unified representation, shallow or deep learning-based methods can be employed to map the features to evoked emotions.

A variety of machine learning algorithms have been utilized to {\it learn the mapping} between the unified representation after feature fusion and evoked emotions. These algorithms include na\"ive Bayes~\cite{machajdik2010affective,borth2013large,wang2013interpretable}, logistic regression~\cite{borth2013large,yuan2013sentribute,song2022have,pang2015deep}, SVM and support vector regression with linear and radial basis function (RBF) kernels~\cite{yanulevskaya2008emotional,lu2012shape,borth2013large,yuan2013sentribute,zhao2014exploring,liu2019affective,song2022have,jiang2014predicting,baveye2015liris,zhang2018recognition,xu2019video,gu2020sentiment,wei2021user,muszynski2021recognizing,shukla2022recognition}, linear discriminant analysis (LDA)~\cite{shukla2022recognition}, multiple instance learning~\cite{rao2016multi}, random forest~\cite{song2022have,wei2021user}, decision tree~\cite{song2022have}, ensemble learning~\cite{acar2017comprehensive}, a mixture of experts~\cite{sun2019gla}, sparse learning~\cite{zhao2017continuous,zhao2020discrete}, and graph/hypergraph learning~\cite{zhao2018predicting}. There is still room for innovation in designing emotion-sensitive mapping algorithms.

The most common {\it deep learning-based mapping} is a multilayer perceptron based on one or more fully connected layers. The difference mainly lies in objective loss functions, such as cross-entropy loss for classification~\cite{you2015robust,you2016building,yang2017joint,zhu2017dependency,campos2017pixels,yang2018retrieving,yang2018weakly,rao2019multi,rao2020learning,yang2021solver,ruan2021color,zhang2022multiscale,li2022weakly,xu2022mdan,deng2022simple,zhang2022image,gu2020sentiment,cheng2021context,wei2021user,ou2021multimodal}, Euclidean and mean squared error loss for regression~\cite{peng2015mixed,song2022have}, contrastive loss and triplet loss for retrieval, and Kullback-Leibler (KL) divergence loss for label distribution learning~\cite{yang2017joint,yang2021circular,achlioptas2021artemis}. Unlike shallow learning methods, these deep learning techniques can typically be trained in an end-to-end manner.
Some improvements have also been made to better explore the interactions between different tasks and the characteristics and relationships of different emotion categories, such as through the use of polarity-consistent cross-entropy loss and regression loss~\cite{zhao2019pdanet,zhao2020end} and hierarchical cross-entropy loss~\cite{yang2021stimuli}. More details are given in the following.

The aforementioned evoked emotion analysis tasks are interconnected. For instance, the detection of emotional regions can inform the emotion classification task, and the emotion category with the largest probability in label distribution learning often aligns with the emotion classification results. However, a notable difference between evoked emotion modeling and traditional computer vision or machine learning lies in the existence of specific correlations among emotions. We summarize recent advancements in multitask learning and the exploration of emotion correlations in evoked emotion modeling.

{\it Multi-task learning} has been shown to significantly improve performance compared to single-task learning by leveraging appropriate shared information and imposing reasonable constraints across multiple related tasks~\cite{zhang2021survey}. It has become popular in evoked emotion modeling, especially when training data is limited for each task. Based on the relations explored, multitask learning can be categorized into task relation learning-oriented~\cite{yang2017joint,yang2018retrieving,yang2018weakly,rao2019multi,zhang2022multiscale,li2022weakly}, testing data relation learning-oriented~\cite{zhao2017continuous}, feature relation learning-oriented~\cite{zhao2020discrete}, and viewer relation learning-oriented approaches~\cite{zhao2018predicting}. Different evoked emotion analysis tasks are often jointly performed, such as classification and distribution learning~\cite{yang2017joint,rao2019multi}, classification and retrieval~\cite{yang2018retrieving}, and classification and detection~\cite{yang2018weakly,li2022weakly,zhang2022multiscale}. Joint optimization of different objective losses allows models to extract more discriminative feature representations. For example, multitask shared sparse regression was proposed to predict continuous emotion distributions of multiple testing images with sparsity constraints, which takes advantage of  feature group structures~\cite{zhao2017continuous}. Different constraints across features are considered~\cite{zhao2020discrete} to reflect their importance by a weighting strategy, which can be viewed as a special late fusion category. Rolling multitask hypergraph learning is proposed to simultaneously predict
personalized emotion perceptions of different viewers where social connections among viewers are considered~\cite{zhao2018predicting}.

{\it Exploring correlations} among different emotion categories or continuous emotion values can improve evoked emotion analysis. Commonly considered emotion correlations include:

\begin{itemize}
\item \textit{Emotion hierarchy.} As emotion categories researchers attempt to model become more diverse and nuanced, level of granularity increases~\cite{xu2022mdan}. As shown in Fig.~\ref{fig:EmotionCorrleations}(a), emotions can be organized into a hierarchy, which has been exploited in evoked emotion analysis. By considering the polarity-emotion hierarchy, i.e. whether two emotion categories belong to the same polarity, polarity-consistent cross-entropy loss~\cite{zhao2020end} and regression loss~\cite{zhao2019pdanet} are designed to increase the penalty of predictions that have the opposite polarity to the ground truth. Hierarchical cross-entropy loss has been proposed to jointly consider both emotion and polarity loss~\cite{yang2021stimuli}. For each level in the emotion hierarchy, one specific semantic level is mapped with local learning to acquire corresponding discrimination~\cite{xu2022mdan}.

\item \textit{Emotion similarity.} Similarities or distances between emotions can vary, with some being closer than others. For example, sadness is closer to disgust than it is to contentment. To account for these similarities, the Mikels' emotion wheel was introduced~\cite{zhao2018predicting} (Fig.~\ref{fig:EmotionCorrleations}(b) left). Pairwise emotion similarity is defined as the reciprocal of ``1 plus the number of steps required to discriminate one emotion from another''. Using the Mikels' wheel, the emotion distribution is transformed from a single emotion category~\cite{yang2017joint,rao2019multi}. Chain center loss is derived from the triplet loss, with anchor-related-negative triplets selected based on emotion similarity~\cite{liang2022chain}. A more accurate method of measuring emotion similarity, based on a well-grounded circular-structured representation called the Emotion Circle, has also been designed~\cite{yang2021circular} (Fig.~\ref{fig:EmotionCorrleations}(b) right). Each emotion category can be represented as an emotion vector with three attributes
(i.e., polarity, type, and intensity) and two properties (i.e., similarity and additivity), allowing for vector addition operations.

\item \textit{Emotion boundedness.} Not every combination of valence, arousal, and dominance values make sense in an emotion space. The transformation between discrete emotion states and their rough continuous values is often possible~\cite{lu2012shape,zhao2021affective} (Fig.~\ref{fig:EmotionCorrleations}(c)). For example, positive valence is linked to happiness, whereas negative valence is linked to sadness or anger, even though exact boundaries may not be clear. When performing multitask learning involving both classification and regression, it is helpful to consider the constraints on these values. For example, it would not be valid to predict happiness with a negative valence. The BoLD dataset (Section~\ref{sec:bold}) leverages this concept to check the validity of crowdsourced annotations~\cite{luo2020arbee}.
\end{itemize}

\begin{table}[!t]
    \centering
    \caption{Performance comparison of representative methods for 8-class evoked emotion classification from images on the FI dataset, measured by average accuracy (\%). *The results are mainly obtained from MDAN~\cite{xu2022mdan} and SimEmotion~\cite{deng2022simple}.}\label{tab:EEImageBenchmark}
    \colorbox{lightyellow}{%
    \begin{tabular}{c|c|c}
    \hline
        Method & Venue & Accuracy (\%)\\ \hline
        SentiBank~\cite{borth2013large} & ACM MM 2013 & 44.5  \\
        Artistic principles~\cite{zhao2014exploring} & ACM MM 2014 & 46.1  \\ 
        DeepSentiBank~\cite{chen2014deepsentibank} & arXiv 2014 & 53.2  \\ 
        Fine-tuned AlexNet~\cite{krizhevsky2012imagenet} & NeurIPS 2012 & 58.3  \\ 
        Fine-tuned VGG-16~\cite{simonyan2015very} & ICLR 2015 & 65.5  \\ 
        Fine-tuned ResNet-101~\cite{he2016deep} & CVPR 2016 & 66.2  \\ 
        Porgressive CNN~\cite{you2015robust} & AAAI 2015 & 56.2  \\ 
        MldrNet~\cite{rao2020learning} & NPL 2020 & 65.2  \\ 
        WSCNet~\cite{yang2018weakly} & CVPR 2018 & 70.1  \\ 
        PDANet~\cite{zhao2019pdanet} & ACM MM 2019 & 72.1  \\ 
        MlrCNN~\cite{rao2019multi} & NEUCOM 2019 & 75.5  \\ 
        SAVEAN~\cite{yang2021stimuli} & TIP 2021 & 72.4  \\ 
        SOLVER~\cite{yang2021solver} & TIP 2021 & 72.3  \\ 
        MDAN~\cite{xu2022mdan} & CVPR 2022 & 76.4  \\ 
        SimEmotion~\cite{deng2022simple} & TAFFC 2022 & 80.3 \\ \hline
    \end{tabular}
    }
\end{table}

\begin{table}[!t]
    \centering
    \caption{Performance comparison of representative methods for 8-class evoked emotion classification from videos on the VideoEmotion-8 dataset, measured by average accuracy (\%). *The results are mainly obtained from VAANet~\cite{zhao2020end} and TAM~\cite{pan2022representation}.}\label{tab:EEVideoBenchmark}
    \colorbox{lightyellow}{%
    \begin{tabular}{c|c|c}
    \hline
        Method & Venue & Accuracy (\%) \\ \hline
        SentiBank~\cite{borth2013large} & ACM MM 2013 & 35.5   \\ 
        E-MDBM~\cite{pang2015deep}  & TMM 2015 & 40.4   \\ 
        ITE~\cite{xu2018heterogeneous}  & TAFFC 2018 & 44.7   \\ 
        V.+Au.+At.~\cite{jiang2014predicting} & AAAI 2014 & 46.1   \\
        CFN~\cite{chen2016emotion}  & ACM MM 2016 & 50.4   \\ 
        V.+Au.+At.+E-MDBM~\cite{pang2015deep}  & TMM 2015 & 51.1   \\ 
        Kernelized~\cite{zhang2018recognition} & TMM 2018 & 49.7   \\
        Kernelized+SentiBank~\cite{zhang2018recognition} & TMM 2018 & 52.5   \\
        VAANet~\cite{zhao2020end} & AAAI 2020 & 54.5   \\ 
        KeyFrame~\cite{wei2021user} & MTAP 2021 & 52.9 \\
        FAEIL~\cite{zhang2018recognition} & TMM 2021 & 57.6   \\ 
        TAM~\cite{pan2022representation} & ACM MM 2022 & 57.5  \\ \hline
    \end{tabular}
    }
\end{table}

\subsubsection{Preliminary Benchmark Analysis}
\label{sssec:EEBenchmark}

{We provide a summary of image- and video-based evoked emotion classification accuracy of various representative methods, reported on the FI~\cite{you2016building} and VideoEmotion-8~\cite{jiang2014predicting} datesets, respectively. The compared methods for image emotion prediction include SentiBank~\cite{borth2013large}, artistic principles~\cite{zhao2014exploring}, DeepSentiBank~\cite{chen2014deepsentibank}, 
fine-tuned AlexNet~\cite{krizhevsky2012imagenet}, fine-tuned VGG-16~\cite{simonyan2015very}, fine-tuned ResNet-101~\cite{he2016deep}, progressive CNN~\cite{you2015robust}, multilevel deep representation network (MldrNet)~\cite{rao2020learning}, weakly supervised coupled network (WSCNet)~\cite{yang2018weakly}, polarity-consistent deep attention network (PDANet)~\cite{zhao2019pdanet}, Multi-level region-based CNN (MlrCNN)~\cite{rao2019multi}, stimuli-aware visual emotion analysis network (SAVEAN)~\cite{yang2021stimuli}, scene-object interrelated visual emotion reasoning network (SOLVER)~\cite{yang2021solver}, multilevel dependent attention network (MDAN)~\cite{xu2022mdan}, and SimEmotion~\cite{deng2022simple}. Results in Table~\ref{tab:EEImageBenchmark} indicate that: (a) fine-tuning deep neural networks, particularly those with more layers, outperforms handcrafted features (e.g., fine-tuned ResNet-101~\cite{he2016deep} versus SentiBank~\cite{borth2013large}); (b) exploring local feature representations and combining  representations from different levels can increase accuracy to around 75\% (e.g., MDAN~\cite{xu2022mdan}); and (c) SimEmotion, which uses large-scale language-supervised pretraining, achieves an overall accuracy of around 80\%, which remains lower than traditional computer vision tasks. The results highlight the challenges in evoked emotion prediction due to large intra-class variance and the need for further progress toward human-level emotion understanding.}

{The compared methods for video emotion prediction include SentiBank~\cite{borth2013large}, enhanced multimodal deep
Bolzmann machine (E-MDBM)~\cite{pang2015deep}, image transfer encoding (ITE)~\cite{xu2018heterogeneous}, visual+audio+attribute (V.+Au.+At.)~\cite{jiang2014predicting},  CFN~\cite{chen2016emotion}, V.+Au.+At.+E-MDBM~\cite{pang2015deep}, Kernelized features and Kernelized+SentiBank~\cite{zhang2018recognition}, visual-audio attention network (VAANet)~\cite{zhao2020end}, KeyFrame~\cite{wei2021user}, frame-level adaptation and emotion intensity learning (FAEIL)~\cite{zhang2018recognition},  and temporal-aware multimodal (TAM) methods~\cite{pan2022representation}. Results in Table~\ref{tab:EEVideoBenchmark} indicate that: (a) fusing information from multiple modalities is more effective than using a single modality (e.g., VAANet~\cite{zhao2020end} versus SentiBank~\cite{borth2013large}); (b) the current best accuracy is less than 60\%; and (c) effectively fusing information from different modalities and selecting key segments or frames are two challenges. Evoked emotion prediction from videos is even more challenging than image-based prediction and requires further research efforts.}

\subsection{Facial Expression and Microexpression Recognition}\label{sec:facial}
Facial expressions play a crucial role in natural human communication and emotion perception.
Facial expression recognition (FER) involves automatic identification of a person's emotional state through analysis of images or video clips and has been a long-standing research topic in computer vision and affective computing.

\subsubsection{Earlier Approaches}
Before 2012, traditional handcrafted features and pipelines were commonly used. The process generally included the following steps:  
detecting facial regions, extracting handcrafted features (e.g., LBP~\cite{shan2009facial}, non-negative matrix factorization (NMF)~\cite{zhi2010graph}, HOG~\cite{dalal2005histograms}) from facial regions,
and employ a statistical classifier (e.g., SVM~\cite{michel2003real}) to recognize emotions. 

Readers interested in traditional methodologies are advised to refer to survey articles~\cite{zeng2008survey,sariyanidi2014automatic,pantic2000automatic,fasel2003automatic}.

Since the creation of the ImageNet dataset and deep CNN AlexNet in 2012, DNNs have demonstrated remarkable image representation capabilities.
FER researchers have established large-scale datasets (e.g., EmotoNet~\cite{fabian2016emotionet}, AffectNet~\cite{mollahosseini2017affectnet}) that provide ample training data for DNNs.
Consequently, deep learning approaches have become the dominant approach in FER.
A survey by \citeauthor{li2020deep}~\cite{li2020deep} provides a comprehensive summary of deep learning-based FER methods from 2012 to 2019. 
During this period, researchers proposed several DNN techniques to improve FER performance, which the authors categorize as follows: 
\begin{itemize}
\item Adding auxiliary blocks to the typical backbone network (e.g., ResNet~\cite{he2016deep} and VGG~\cite{simonyan2015very}). The Scoring ensemble (SSE)~\cite{hu2017learning} proposed three auxiliary blocks to extract features from the shallow-, intermediate-, and deep-layers, respectively.
\item Ensembling different models to achieve outstanding performance. For example,~\cite{bargal2016emotion} concatenated  features from three different networks--VGG13~\cite{simonyan2015very} and VGG16~\cite{simonyan2015very}, and ResNet~\cite{he2016deep}.
\item Designing specialized loss functions (such as Center loss~\cite{wen2016discriminative}, Island loss~\cite{cai2018island}, and (N+M)-tuple cluster loss~\cite{liu2017adaptive}) to learn  facial features.
\item Leveraging multitask learning to learn various features from facial images. For instance,~\cite{zhang2017facial} proposed MSCNN to jointly learn face verification and facial expression recognition tasks.
\end{itemize}
Moreover, some studies concentrated on FER from video clips using spatiotemporal networks to capture temporal correlations among video frames. 
Some researchers~\cite{yan2018multi} also used recurrent neural networks (RNNs) (e.g., LSTM), wheras others~\cite{ouyang2017audio,abbasnejad2017using} used 3-D convolutional networks such as C3D~\cite{tran2015learning}.

\begin{figure}[!ht]
    \setlength{\tabcolsep}{3pt}
    \begin{center}
    \begin{tabular}{ccc}
          \includegraphics[width=0.31\linewidth]{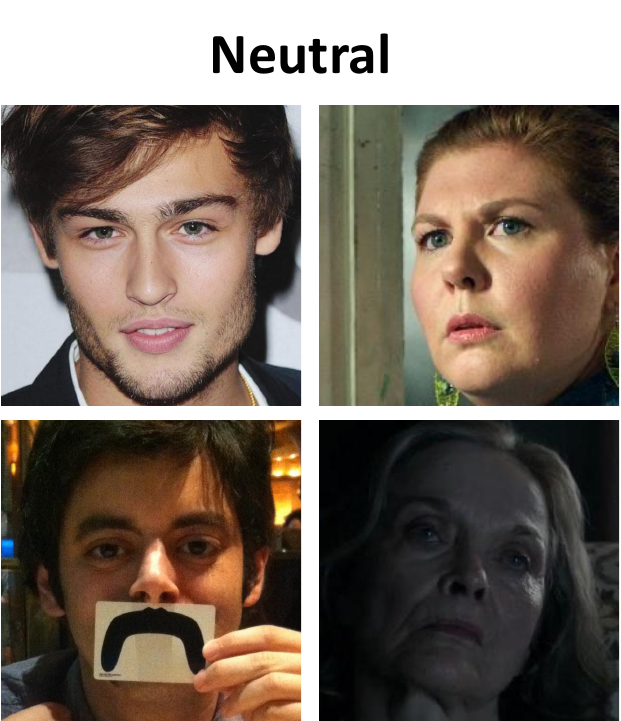} & \includegraphics[width=0.31\linewidth]{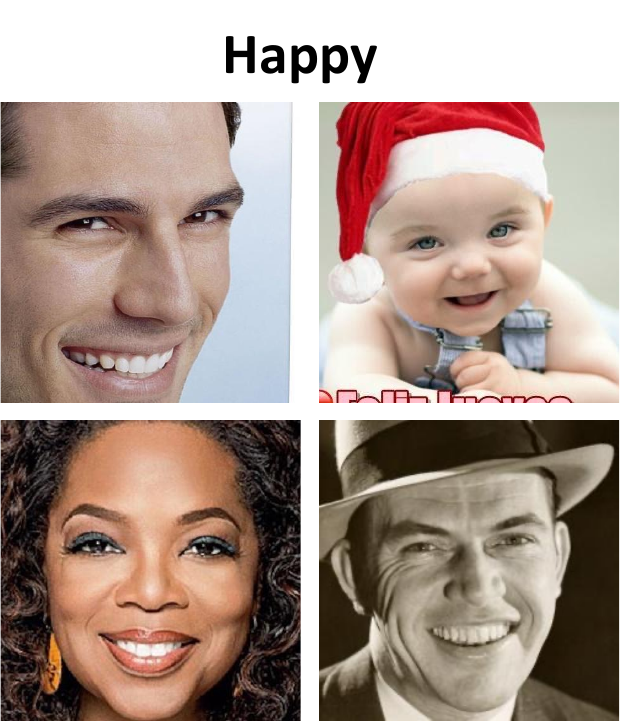} & \includegraphics[width=0.31\linewidth]{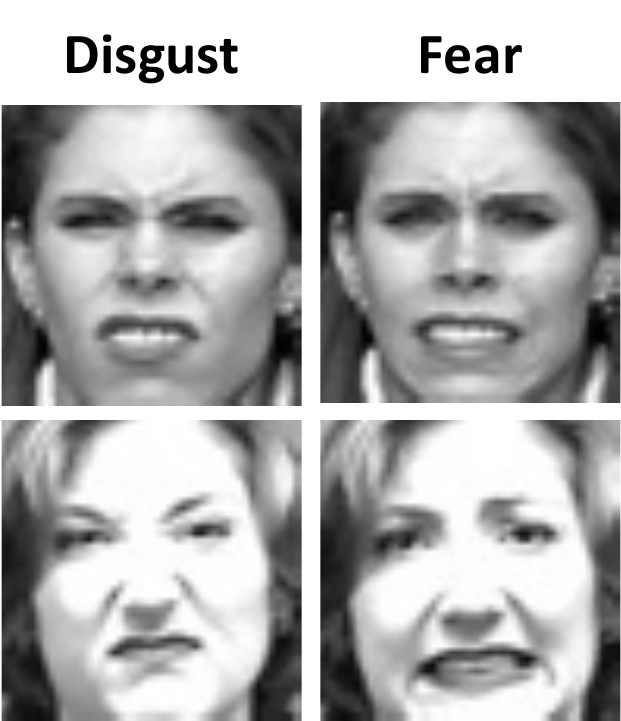} \\
    (a) Annotation & (b) Robustness & (c) Subtlety \\
    \end{tabular}
    \caption{Three challenges in facial expression recognition. 
    Images in (a-b) are obtained from the AffectNet dataset~\cite{mollahosseini2017affectnet} and images in (c) are obtained from the CK+ dataset~\cite{lucey2010extended}.
    }\label{fig:FER}
    \end{center}
    \end{figure}

\subsubsection{Recent Approaches}
Although deep learning-based methods have achieved remarkable success, FER remains a difficult task due to these three challenges. 
\begin{itemize}
\item {\it Annotation.} Annotations of FER datasets contain much ambiguity.
Each annotator subjectively evaluates facial expressions in images, leading to different annotations to the same image.
Some images also have inherent ambiguity, making it difficult to assign a clear emotion label.
Low image quality (such as occlusion or poor lighting) and ambiguous facial expressions exacerbate this problem.
For instance, the images in Fig.~\ref{fig:FER} (a) are labeled as neutral, 
but this label is uncertain due to annotator subjectivity and/or image quality.
\item {\it Robustness.} FER contains several sources of disturbance. Datasets are made up of  individuals of varying ages, genders, and cultures.
In addition, visual variations (such as human facial pose, illumination, and occlusions) commonly exist in facial images, which cause distinct appearance differences. 
These identity and visual variations can make it difficult for FER models to extract useful features.
As illustrated in Fig.~\ref{fig:FER} (b), image appearances can vary greatly due to differences in age, gender, race, facial pose, and lighting.
Moreover, individuals with different identities and cultures might display their facial expressions differently, adding another challenge for FER models.
\item {\it Subtlety.} Some facial expressions are delicate, and 
some emotions can also be conveyed through subtle facial actions.
Such fine distinctions can make it challenging to distinguish between emotions. 
For instance, the difference between ``fear'' and ``disgust'' images in Fig.~\ref{fig:FER} (c) is nuanced.
Thus, FER models must efficiently extract discriminative features to differentiate emotions.
\end{itemize}

Most deep learning methods developed after 2019 have attempted to address these problems.
Regarding the annotation problem, \citeauthor{zeng2018facial} pointed out the inconsistency of emotion annotation across different FER datasets due to annotators' subjective evaluations of emotion~\cite{zeng2018facial}.
To address this issue, they proposed a framework called Inconsistent Pseudo Annotations to Latent Truth (IPA2LT), which trains multiple independent models on different datasets separately. 
These models may assign inconsistent pseudo labels to the same image because each model reflects the subjectivity of the corresponding dataset annotators.
By comparing image's inconsistent labels, IPA2LT  estimates the latent true label.
Another factor that can contribute to annotation ambiguity is uncertain facial images, such as blurry images or those with ambiguous emotions.
To mitigate the effect of uncertain images, researchers have proposed various approaches.
\citeauthor{wang2020suppressing} developed a Self-Cure Network (SCN) to identify and then suppress uncertain images~\cite{wang2020suppressing}.
SCN uses a self-attention mechanism to estimate the uncertainty of facial images and a relabeling mechanism to adjust labels of those images.
\citeauthor{chen2020label} argued that annotating uncertain images with multilabel and intensity, rather than the one-hot label commonly used in current FER datasets, is more suitable~\cite{chen2020label}.
Label Distribution Learning (LDL) allows models to learn label distribution (i.e., multilabel with intensity).
\citeauthor{chen2020label} introduced an approach called Label Distribution Learning on Auxiliary Label Space Graphs (LDL-ALSG)~\cite{chen2020label}. 
Given one image, LDL-ALSG first leverages models of related tasks (such as AU detection) to find its neighbor images. 
Then LDL-ALSG employs a task guide loss to let images learn a similar label distribution (i.e.,multilabel with intensity) with the neighbors.
\citeauthor{she2021dive} combined LDL and uncertain image estimation. The proposed DMUE network mines the label distribution and estimates the uncertainty of images together~\cite{she2021dive}.

To address robustness challenges in FER, researchers have explored mitigating the impact of identity variations on recognition~\cite{chen2021understanding,zeng2022face2exp}.
\citeauthor{chen2021understanding} investigated the influence of gender and revealed that models tend to recognize women's faces as happy more often than men's, even when smile intensities are the same~\cite{chen2021understanding}.
To overcome this issue, the authors proposed a method that first detected facial AUs and then applied a triplet loss to ensure that people with similar AUs exhibited similar expressions, regardless of gender~\cite{chen2021understanding}.
\citeauthor{zeng2022face2exp} demonstrated that emotion categories can introduce bias into a dataset~\cite{zeng2022face2exp}.
In some datasets, emotions (such as disgust) occur less frequently than more prevalent emotions such as happiness and sadness, leading to poor performance of networks trained on these datasets on minority emotion classes.
The authors utilized a million-image-level facial recognition dataset (much larger than the FER dataset) and a meta-learning framework to address the issue~\cite{zeng2022face2exp}.
{\citeauthor{li2021adaptively} recognized the category imbalance challenge and addressed it by proposing AdaReg loss to dynamically adjust the importance of each category~\cite{li2021adaptively}.}

Other researchers have focused on visual disturbance variations, which lead to the robustness issue.
\citeauthor{wang2020region} designed a region attention network to capture important facial regions, thus obtaining occlusion-robust and pose-invariant image features~\cite{wang2020region}.
\citeauthor{zhang2021learning} used a Deviation Learning Network (DLN) to learn identity-invariant features~\cite{zhang2021learning}.
\citeauthor{wang2019identity} considered both identity and pose variations together~\cite{wang2019identity}, where an encoder was followed by two discriminators that classified pose and identity independently, while the encoder extracted features that were invariant to both. A classifier for expressions was then used to produce expression predictions.

To address the subtlety problem in FER, \citeauthor{ruan2021feature} decomposed facial expression features into shared features that represented expression similarities, and unique features that represented expression-specific variations using a Feature Decomposition Network (FDN) and a Feature Reconstruction Network (FRN), respectively~\cite{ruan2021feature}.
The authors further addressed robustness and subtlety problems by proposing a disturbance feature extraction model (DFEM) to identify disturbance features (such as pose and identify) and an adaptive disturbance-disentangled model (ADDM) to remove disturbance features extracted by DFEM and extract the discriminative features across different facial expressions.
\citeauthor{farzaneh2021facial} enhanced Center Loss~\cite{wen2016discriminative,farzaneh2021facial}. Although Center Loss can learn discriminative features, it can also include some irrelevant features.
The proposed Deep Attentive Center Loss adopts an attention mechanism to adaptively select important discriminative features.
\citeauthor{xue2021transfer} leveraged Transformers to detect discriminative features and showed that the original vision transformer (ViT)~\cite{dosovitskiy2020image} can only capture most discriminative features but neglects other features~\cite{xue2021transfer}.
Thus this work proposes the Multi-Attention Dropping (MAD) technique to randomly drop some attention maps, such that the network can characterize all comprehensive features except the most discriminative ones.
{Furthermore, \citeauthor{savchenko2022classifying}~\cite{savchenko2022classifying} achieved remarkable performance on FER using the well-performing network EffientNet~\cite{tan2019efficientnet} to extract discriminative features.}

\begin{table}[ht!]
    \centering
    \caption{Performance comparison of representative methods for 7-class Facial Expression Recognition on the AffectNet dataset, measured by average accuracy (\%).}\label{tab:FERBenchmark}
    \colorbox{lightyellow}{%
    \begin{tabular}{c|c|c}
    \hline
        Method & Venue & Accuracy (\%) \\ \hline
        IPA2LT~\cite{zeng2018facial} & ECCV 2018 & 57.31 \\
        IPFR~\cite{wang2019identity}   & ACM MM 2019 &57.40\\
        RAN~\cite{wang2020region} & TIP 2020 & 52.97 \\
        SCN~\cite{wang2020suppressing} & CVPR 2020 & 60.23 \\
        DACL~\cite{farzaneh2021facial}   & WACV 2021 &65.20\\
        DMUE~\cite{she2021dive} & CVPR 2021 & 63.11 \\
        KTN~\cite{li2021adaptively} & TIP 2021 & 63.97 \\
        TranFER~\cite{xue2021transfer} & CVPR 2021 & 66.23 \\ 
        Face2Exp~\cite{zeng2022face2exp} & CVPR 2022 & 64.23 \\
        ADDL~\cite{ruan2022adaptive}          & IJCV 2022   & 66.20 \\ 
        EfficientNet-B2~\cite{savchenko2022classifying} & TAFFC 2022  & 66.34 \\ \hline
    \end{tabular}
    }
\end{table}

\subsubsection{Preliminary Benchmark Analysis}
{Table~\ref{tab:FERBenchmark} provides a concise overview of the recent FER method performance on the AffectNet dataset, one of the most comprehensive FER benchmarks. The table shows the emotion classification accuracy of selected representative methods, which were previously discussed in the text. As seen in the table, TranFER, ADDL, and EfficientNet-B2 are among the top-performing methods on the AffectNet benchmark. The utilization of advanced network structures from general image recognition tasks in TranFER and EfficientNet-B2 highlights the importance of drawing knowledge and expertise from the image recognition field in FER. However, it is important to note that the highest reported accuracy remains below 70\%. Given the over 90\% top-1 accuracy achieved by state-of-the-art image recognition methods on the challenging ImageNet benchmark, this deficit suggests significant potential for improvement in FER.}

\subsubsection{Microexpression Recognition (MER)}

The methods described above are for recognizing facial expressions.
However, individuals may consciously exhibit certain facial expressions to conceal their authentic emotions. 
In contrast to conventional facial expressions, which can be deliberately controlled, microexpressions are fleeting and spontaneous and can uncover an individual's genuine emotions. 
A microexpression is brief in duration and can be imperceptible with the naked eye. MER often requires high frame-rate videos as input and the development of spotting algorithms to isolate temporally the microexpression within videos.
A recent survey gives a more in-depth introduction to developments in MER~\cite{ben2022video}.

\subsection{Bodily Expressed Emotion Understanding (BEEU)}\label{sec:body}

In everyday life, people express their emotions through various means, including via their facial expressions and body movements. 
Recognizing emotions from body movements has some distinct advantages over recognizing emotions from facial images for many computer and robotic applications: 
\begin{itemize}
\item In crowded environments where facial images may be obscured or lack sufficient resolution, body movements and postures can still be reasonably estimated. This context is particularly important in robotic applications where the robot may not be close to all individuals in its environment.
\item Due to privacy concerns, facial information may be inaccessible. For example, in some medical applications, sharing of facial images or videos is restricted to protect sensitive patient identity information.
\item Incorporating body expressions as an additional modality can result in more accurate emotion recognition compared to using facial images alone. For example, when the person is not facing the camera, the camera cannot obtain a frontal view.
\end{itemize}

Psychologists have conducted extensive studies to examine the relationship between body movements and emotions.
Research suggests that body movements and postures are crucial for understanding emotion, encoding rich information about an individual's status, including awareness, intention, and emotional state~\cite{wallbott1998bodily,meeren2005rapid,de2006towards,aviezer2012body,melzer2019we}.
Several studies, including one published in {\it Science}, found that 
the human body may be more diagnostic than the face for emotion recognition~\cite{aviezer2012body,nelson2017adults,karaaslan2020does}.

However, the field of BEEU in visual media has progressed relatively slowly.
Unlike FER, which has seen significant progress with deep learning methods since 2013, most BEEU studies relied on traditional, handcrafted features until 2018. 
The bottleneck for BEEU is the scarcity of large-scale, high-quality datasets.
As mentioned in Section~\ref{sec:bold}, collecting and annotating a dataset of bodily expressions with high-quality labels is extremely challenging and costly.
Understanding and perception of emotions from concrete observations is heavily influenced by context, interpretation, ethnicity, and culture. 
There is often no gold-standard label for emotions, especially bodily expressions.
Prior to 2018, research on bodily expression was limited to small, acted, and constrained lab-setting video data~\cite{gunes2007bi,kleinsmith2006cross,schindler2008recognizing,dael2012emotion}.
These datasets were insufficient for deep learning-based models that required a large amount of data.
The recent BoLD dataset by \citeauthor{luo2020arbee}~\cite{luo2020arbee} introduced in Section~\ref{sec:bold} is the largest BEEU dataset to date. It contains over 10,000 video clips of body movements with high-quality emotion labels.
In addition, \citeauthor{randhavane2019identifying}~\cite{randhavane2019identifying} proposed the E-Walk data, including 1,136 3-D pose sequences extracted from raw video clips.
As a result, computer vision research is increasingly focused on emotion recognition through body movement recordings.

\begin{figure}[ht!]
\begin{tabular}{c}
 \includegraphics[width=0.47\textwidth]{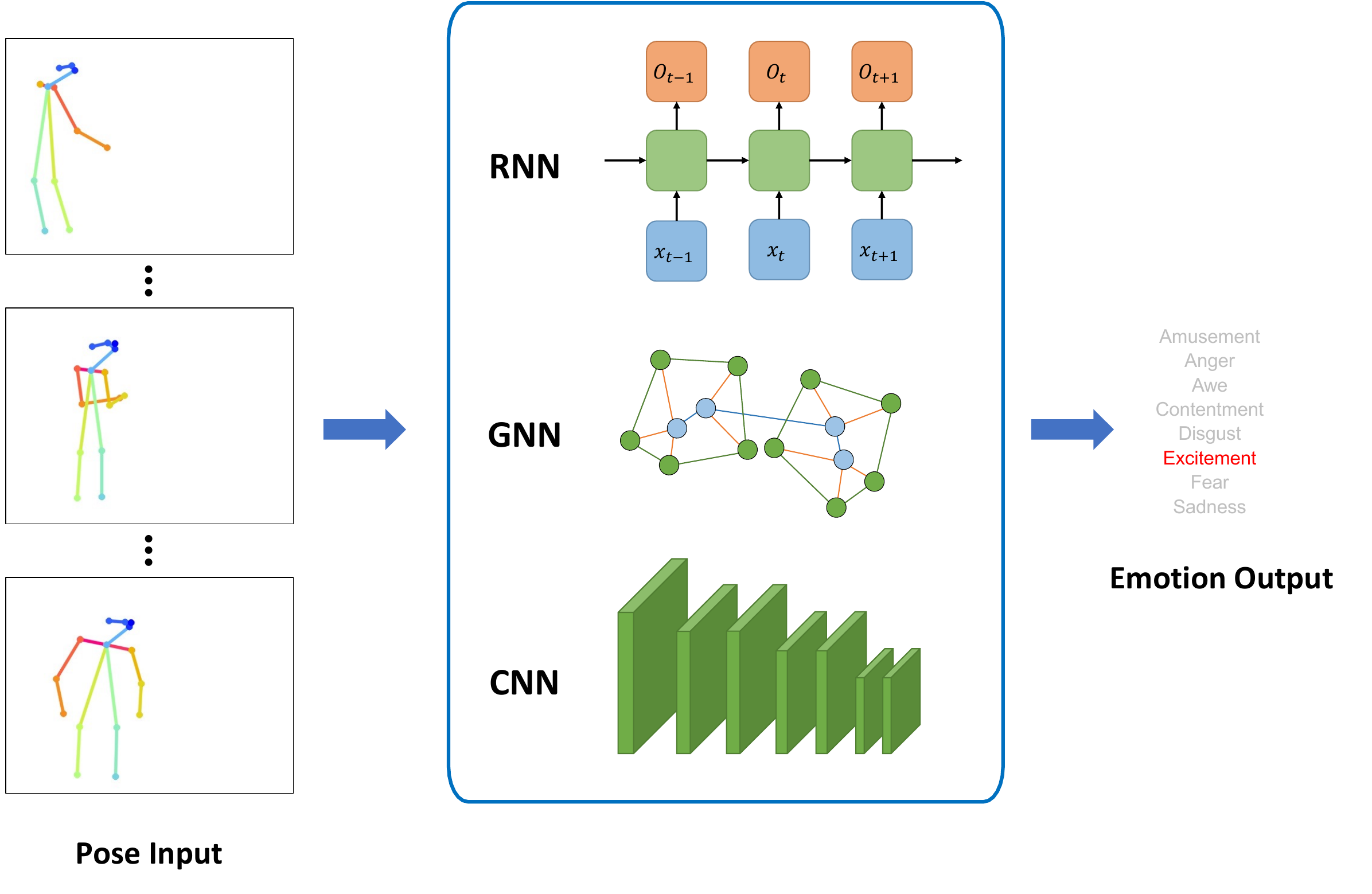}
\end{tabular}
\caption{Skeleton-based BEEU takes the 2D/3D pose sequence of the target human instance as input.}
\label{fig:frameworkbeeu1}
\end{figure}

\begin{figure}[ht!]
\begin{tabular}{c}
    \includegraphics[width=0.47\textwidth]{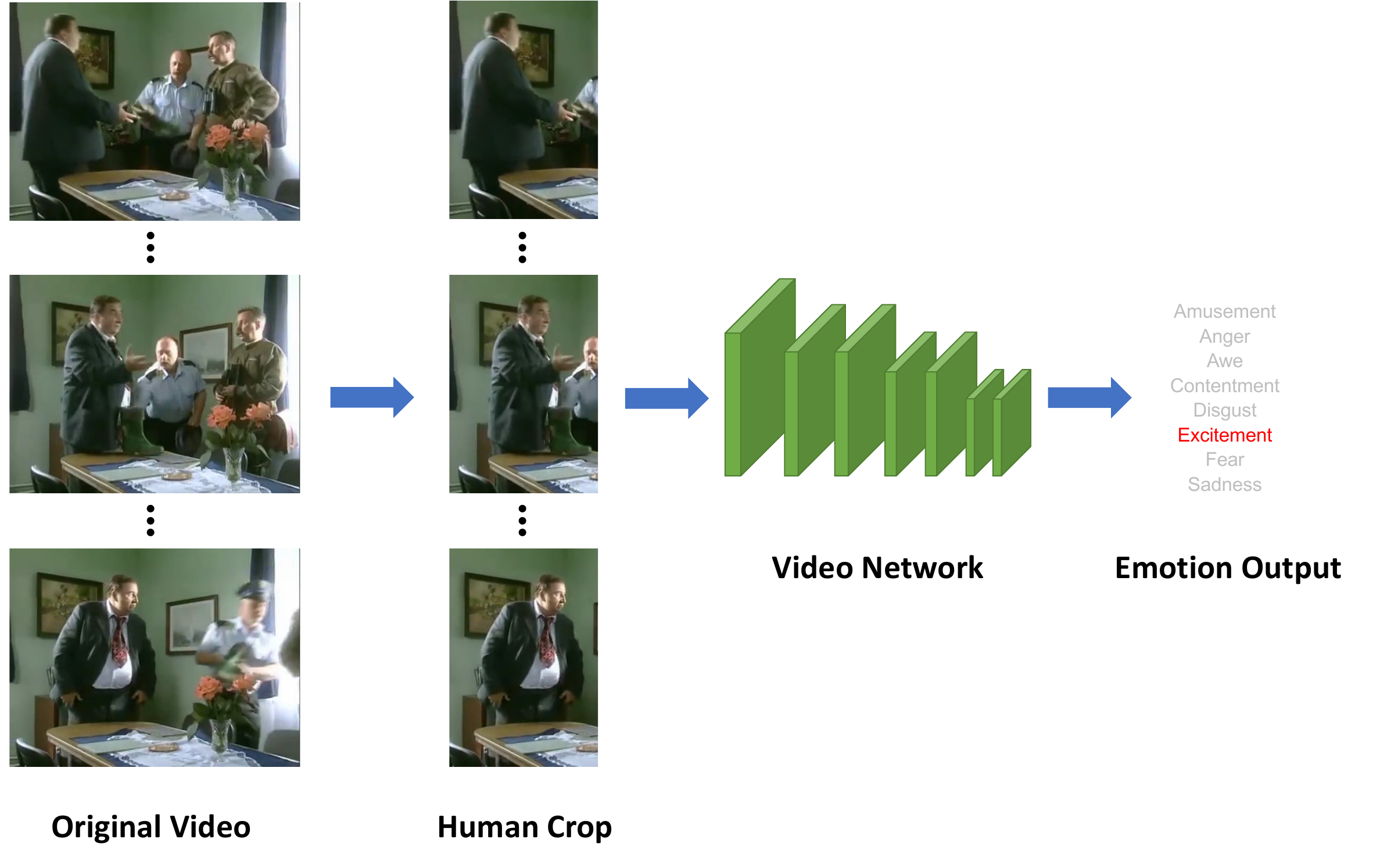} 
\end{tabular}
\caption{Pixel-based BEEU uses raw videos as input. The target human is cropped and the resulting frames are input into a video analysis neural network.}
\label{fig:frameworkbeeu2}
\end{figure}

Because static body images alone can hardly inform a person's emotions, BEEU often considers human video clips.
According to input modality of the models, BEEU methods can be classified as pixel-based and skeleton-based. Pixel-based methods use entire video clips, whereas skeleton-based methods first extract 2D/3D pose information and then feed it into the models.
Some BEEU works, which focus on movement when walking, are known as gait-based. Herein, we refer to them as skeleton-based as well because gait is also a 2D/3D pose. Figs.~\ref{fig:frameworkbeeu1} and~\ref{fig:frameworkbeeu2} illustrate the two kinds of methods.

\subsubsection{Skeleton-based Methods} 

Our review of past publications shows a greater adoption of skeleton-based approaches compared to pixel-based. This is due to two reasons. First, skeleton data, consisting of sequences of 2D/3D joint coordinates, requires less engineering effort to process than video clips. 
A straightforward method is to feed coordinates into a machine learning classifier (e.g. SVM) for direct emotion prediction results. Second, improved MoCap systems enable researchers to easily collect accurate 3-D poses from individuals walking in laboratory settings.

In early stages of skeleton-based BEEU research, a conventional approach was followed, which entailed the extraction of low-level features from 2D/3D pose sequences and subsequent utilization of a machine learning classifier to predict emotion. These features were categorized as follows:
\begin{itemize}
\item Frequent domain features, which transform temporal information into a frequency domain, are obtained through Fourier transformation.
For instance, \citeauthor{li2016identifying} used Fourier transformation to convert 3-D pose sequences into frequency domain features, which were then classified using linear discriminant analysis, na\"ive Bayes, decision tree, and SVM algorithms~\cite{li2016identifying}. 
\item Motion features, characterize the movement of body joints, include joint velocity and acceleration~\cite{luo2020arbee}.
\item Geometry features, which describe self-transformation of the body, encompass angles of specific skeletons and distance between certain joints, among others.
In particular, \citeauthor{crenn2016body} combined motion features, geometry features, and Frequent domain features into an SVM classifier~\cite{crenn2016body}.
\end{itemize}

\citeauthor{luo2020arbee} compared traditional and deep learning-based methods using the BoLD dataset and developed the Automated Recognition of Bodily Expression of Emotion (ARBEE) system~\cite{luo2020arbee}.
A traditional machine learning approach was designed, where motion and geometry features were extracted from 2-D pose sequences, and a random forest classifier was employed to categorize  emotions.
In addition, a ST-GCN was utilized to train and evaluate the BoLD dataset. 
The result indicated that carefully designed traditional machine learning methods outperformed the ST-GCN model from scratch.

Since the development of ARBEE, various deep learning-based methods have emerged. These methods can be broadly categorized into three groups based on the type of neural network used: RNN, Graph Neural Network (GNN), and CNN. 

\citeauthor{randhavane2019identifying} used a LSTM network, which is a type of RNN, to extract temporal features from a 3-D pose sequence, which were then concatenated with handcrafted features for classification~\cite{randhavane2019identifying}.
\citeauthor{bhattacharya2020take} leveraged a semi-supervised technique to improve the performance of an RNN model~\cite{bhattacharya2020take}. 
The work consisted of a GRU (a kind of RNN model) for feature extraction from a 3-D pose sequence, followed by an autoencoder with both encoder and decoder components. 
During training, when the input data was labeled with emotions, the classifier after the encoder produced the emotion prediction, and the decoder reconstructed the 3-D pose. 
If the input lacked emotion labels, only the decoder was used for reconstruction.

\citeauthor{bhattacharya2020step}~\cite{bhattacharya2020step} adopted ST-GCN~\cite{yan2018spatial} to classify emotion categories from 3-D pose sequences. To increase the size of the training set, the authors used a Conditional Variational Autoencoder (CVAE) to generate some synthetic data.
\citeauthor{banerjee2022learning} combined GCN and natural language processing (NLP) techniques to achieve zero-shot emotion recognition, which entailed recognizing novel emotion categories not seen during training~\cite{banerjee2022learning}.
The authors used ST-GCN to extract visual features from the 3-D pose sequences and used the word2vec method to obtain word embeddings from emotion labels.
An adversarial autoencoder was used to align  visual features with word embeddings.
During inference, the system searched for the emotion label that best matched the output visual feature.

Inspired by the success of image recognition, some studies have attempted to convert skeleton sequences into images.
\citeauthor{narayanan2020proxemo} embedded 3-D pose sequences into images, then utilized a CNN for classification~\cite{narayanan2020proxemo}.
\citeauthor{hu2022tntc} employed a two-stream CNN, where one stream directly embedded the 3-D pose into an image, and the other stream converted  handcrafted features from the 3-D pose into another image~\cite{hu2022tntc}. The two CNNs were integrated using Transformer layers.

\subsubsection{Pixel-based Methods}

Pixel-based networks for BEEU require video clips as input.
Due to an increased redundancy in video clips compared to 2D/3D pose sequences, these networks necessitate larger training datasets to extract distinctive features.
With the availability of the large-scale BEEU dataset, researchers have focused on pixel-based methods. Because the well-established field of action recognition, which also uses videos to analyze human behavior, has many similarities to BEEU, current BEEU research often incorporates networks from action recognition.

The ARBEE study benchmarked the performance of various action recognition networks on the BoLD dataset~\cite{luo2020arbee}.
They cropped human body regions from raw videos as the input and then fed them into different methods, including a traditional handcrafted feature method and three deep learning methods (Two-Stream (TS)~\cite{simonyan2014two}, I3D~\cite{carreira2017quo}, and TSN~\cite{wang2016temporal}). 
Both TS and TSN consisted of a two-stream network, where one stream processed the RGB image sequence and the other processed the optical flow sequence.
TS and TSN used 2-D convolutional networks, whereasI3D employed a 3-D convolutional network.
Results indicated that all deep learning methods significantly outperformed the traditional handcrafted feature method.
Among the three deep learning methods, I3D performed worse than TS and TSN, potentially due to its requirement for more training data to reach a comparable level of performance. 
The BoLD dataset is not as extensive as action recognition datasets like Kinetics-400~\cite{kay2017kinetics}, and thus, 
I3D trained on BoLD could not fully exhibit its capability as it could on Kinetics-400.
At the same time, TS and TSN produced similar results.

The ARBEE study also evaluated the impact of a person's face on BEEU model performance~\cite{luo2020arbee}.
They designed an ablation study with three different video inputs: the whole body (crop the whole human body from the raw video clips), just the face (only crop the face part), and the body without the face (crop the whole human body but mask the facial region)~\cite{luo2020arbee}. Results showed that using either the face or body alone was comparable to using the whole body.
This demonstrated that both the face and body contributed significantly to the final prediction.
Although the whole body setting of the TSN model outperformed the separate models, it did so by combining facial and body emotions.

Most BEEU research has followed ARBEE to continue pixel-based approaches.
Because ARBEE cropped human body regions as input, a direct idea to improve upon ARBEE is to explicitly utilize facial images and background images as input as well.
Recent studies~\cite{huang2021emotion,filntisis2020emotion} used an extra network to extract context information from whole images, then fused the context features with features extracted from body images.
Some research adopts an additional network with facial images as input~\cite{pikoulis2021leveraging}.
{Moreover, inspired by the cutting-edge vision-language research of CLIP~\cite{radford2021learning}, \citeauthor{zhang2023emotionclip} developed EmotionCLIP, a contrastive vision-language pretraining paradigm to extract comprehensive visual emotion representations from whole images, encompassing both context information and human body information simultaneously~\cite{zhang2023emotionclip}. Because EmotionCLIP uses only uncurated data, it addresses the challenge of data scarcity in emotion understanding. EmotionCLIP outperforms state-of-the-art supervised visual emotion recognition methods and competes with many multimodal approaches across various benchmarks, demonstrating its effectiveness and transferability.}

{Certain studies delve deeply into the analysis of body gestures and movement in the context of BEEU. ARBEE currently provides only a generalized emotion label for entire video clips, lacking specific descriptions for human body gestures or movements. In contrast, \citeauthor{liu2021imigue} and \citeauthor{chen2023smg} introduced datasets for {\it micro-gesture} understanding and emotion analysis, featuring detailed body gesture labels, such as crossed fingers, for each human movement clip~\cite{liu2021imigue,chen2023smg}. These enriched datasets have the potential to improve machines' understanding of emotions conveyed through gestures. As mentioned in Section~\ref{sec:representation2}, LMA is a comprehensive method for describing human movement. \citeauthor{wu2023bodily} further advanced BEEU by presenting an LMA dataset that provides accurate LMA labels for human movements~\cite{wu2023bodily}. \citeauthor{wu2023bodily} incorporated a novel dual-task model structure that simultaneously predicts emotions and LMA labels, achieving remarkable performance on the BoLD dataset.}

\begin{table}[ht!]
    \centering
    \caption{Performance comparison of different methods for Bodily Expressive Emotion Understanding on the BoLD dataset.}\label{tab:BEEUBenchmark}
    \colorbox{lightyellow}{%
    \begin{tabular}{c|c|c}
    \hline
     Method & Venue & mAP (\%) \\ \hline
     \multicolumn{3}{l}{\textit{Skeleton-based:}}\\ \hline
     ST-GCN~\cite{zeng2018facial} & AAAI 2018 & 12.63 \\
      Random Forest~\cite{luo2020arbee}   & IJCV 2020 & 13.59\\ \hline
      \multicolumn{3}{l}{\textit{Pixel-based:}}\\ \hline
      TS~\cite{simonyan2014two} & NeurIPS 2014 & 17.04 \\
      TSN~\cite{wang2016temporal} & ECCV 2016 & 17.02 \\
      I3D~\cite{carreira2017quo}   & CVPR 2017 & 15.34 \\
      \citeauthor{filntisis2020emotion}~\cite{filntisis2020emotion} & ECCVW 2020 & 17.96 \\
      \citeauthor{pikoulis2021leveraging}~\cite{pikoulis2021leveraging} & FG 2021 & 21.87 \\ 
      {EmotionCLIP~\cite{zhang2023emotionclip}} & {CVPR 2023} & {22.51} \\ 
      {\citeauthor{wu2023bodily}~\cite{wu2023bodily}} & {arXiv 2023} & {23.09} \\ \hline
    \end{tabular}
    }
\end{table}

\subsubsection{Preliminary Benchmark Analysis and Current Directions}

{Table~\ref{tab:BEEUBenchmark} presents the results of various BEEU methods using the BoLD dataset, with performance measured by mean average precision (mAP) across 26 emotional categories. Results indicated that pixel-based methods outperformed skeleton-based ones, which was not surprising given RGB images contained more information. Despite significant progress in BEEU, performance remains relatively low, with mAP scores below 25\%.}

{As BEEU is a relatively new area in computer vision, we explore its potential future directions inspired by the trajectory of related areas like action recognition and FER.}
First, existing bodily expression datasets are not sufficiently large. BoLD contains only thousands of instances, which is much less than contained in action recognition datasets such as Kinetics-400.
The BoLD team is developing larger and more comprehensive datasets to satisfy data requirements of deep learning methods, but those expansions are not yet complete.
Second, existing approaches are largely based on action recognition methods without leveraging deep affective features. Previous work only applied low-level features, neglecting characteristics of concurrent body movement. 
Third, annotation ambiguity, similar to FER, is also challenging. Fourth, any method that segments body regions must account for changing relationships among body sections (i.e., Shape Change in LMA) which is crucial for bodily expressed emotion, and may be particularly relevant to the dimensional emotion model because Shape Change reveals approach and retreat. Section~\ref{sec:challenge} discusses technological barriers more broadly and in greater depth.

\subsection{Integrating Multiple Visual Input to Model Expression}\label{sec:integrated}

We discussed earlier how facial and body images have been used as separate inputs to identify human emotions. 
As was already established, the context of people in a scene also contributes to inferring their emotion.
Humans synthesize all visual information to produce emotion determinations.
Naturally, computers are also capable of making an identification by combining all visual inputs including facial images, body images, and context information.

\begin{figure}[ht!]
    \centering
    \includegraphics[trim=0 0 0 0,clip,width=0.47\textwidth]
    {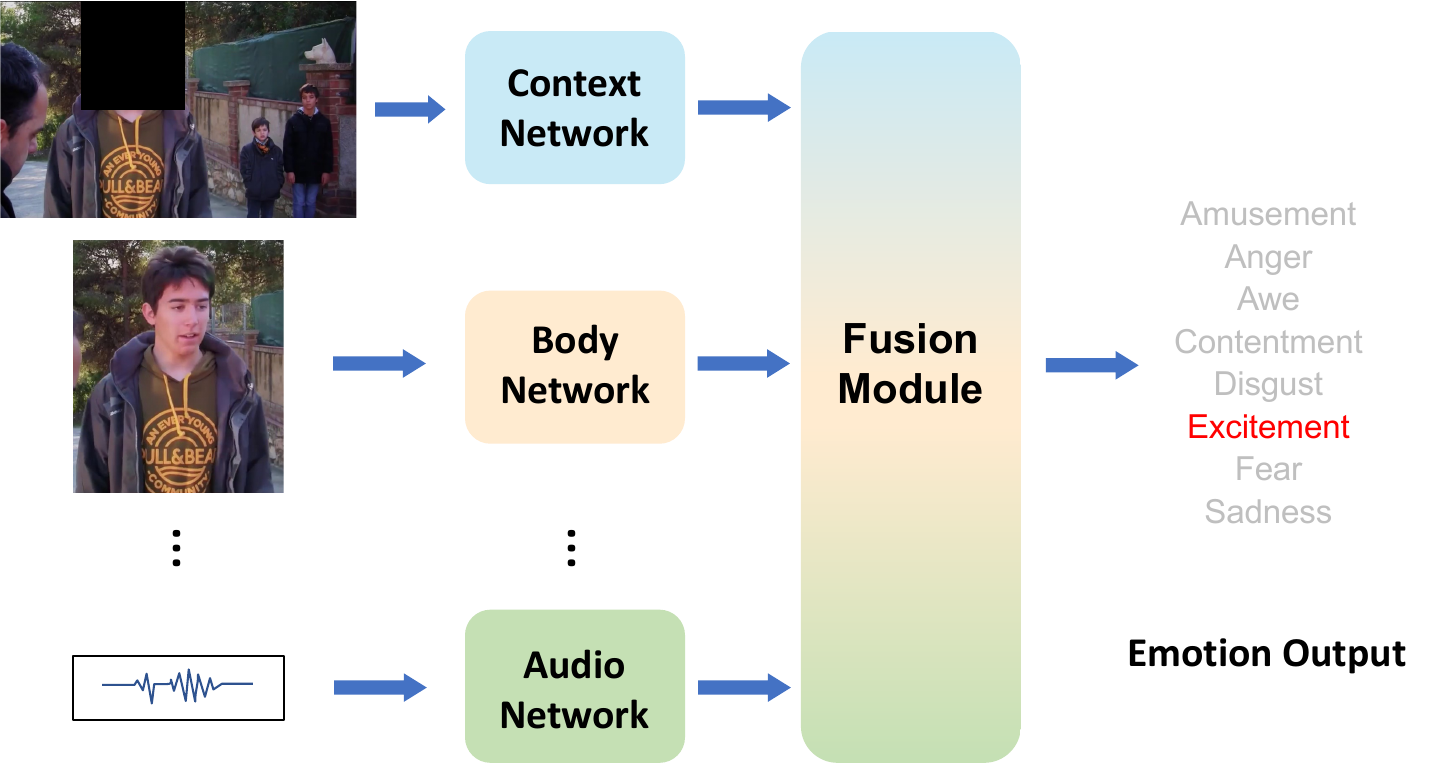}
    \caption{The multistream network extracts features from different inputs, which can be various visual inputs (such as entire and body images) or inputs from other modalities (such as audio). The fusion module combines those features.}
    \label{fig:multimodel}
    \end{figure}
    
As illustrated in Fig.~\ref{fig:multimodel}, recent work used multistream networks to fuse different visual inputs.
\citeauthor{kosti2019context}~\cite{kosti2019context} and \citeauthor{lee2019context}~\cite{lee2019context} developed initial approaches by adopting a two-stream network.
Specifically, \citeauthor{kosti2019context} used body images and entire images as input to extract context information and body features separately.
\citeauthor{lee2019context} adopted faceless images (i.e., cropped the human face out of the entire image) and facial images as input, and a feature fusing network that dynamically fused context and facial features based on their significance.

Subsequent research has taken two paths.
Some researchers have explored more kinds of visual inputs.
For example, \citeauthor{mittal2020emoticon} demonstrated that a depth map can indicate how people interact with each other~\cite{mittal2020emoticon}.
The estimated depth map from the raw image served as one input.
In addition, this work used three other inputs--the facial image, the 2-D pose, and the bodiless (i.e., cropping the human body out of the entire image) images--to extract facial information, body posture information, and context information separately.
This was a four-stream network overall.
Studies~\cite{huang2021emotion,filntisis2020emotion,pikoulis2021leveraging} have also used a three-stream network with the entire, facial, and body images as input.

Other researchers focused on more effective fusion of visual features.
Existing multistream approaches in typical emotion recognition adopt a simple fusion strategy, i.e., ensemble the predictions from each stream~\cite{huang2021emotion,filntisis2020emotion,pikoulis2021leveraging}.
To improve upon this, \citeauthor{le2021global} proposed the Global-Local Attention (GLA) module to enhance interaction between facial and context features~\cite{le2021global}.

\subsection{Multi-modal Modeling of Expressions}\label{sec:multimodal}
{
Emotions can be conveyed and perceived not just through visual signals, but also through text and audio. To effectively process multimodal signals, however, is nontrivial. 
With advancements in deep learning, new technical approaches to this problem have emerged, particularly the vision-and-language model~\cite{kim2021vilt,radford2021learning}.
Similar to methods discussed in Section~\ref{sec:integrated}, multimodal approaches also follow a multistream pipeline, as illustrated in Fig.~\ref{fig:multimodel}. These techniques use independent networks to extract features from different inputs of multiple modalities and fuse these features with a fusion network. 
}

{
Multi-modal approaches typically utilize backbone networks, such as BERT~\cite{bert} and ResNet, to extract text and audio features.
The visual feature extraction process, however, differs and is typically performed using one of the following three methods:
\begin{itemize}
\item {\it Region features.} A detection network is employed to identify regions of interest (ROIs),  from which features are extracted. 
\item {\it Grid features.} A backbone network, such as ResNet 101, is used to extract features for the entire image.
\item {\it Patch projection.} The image is split into patches, and a linear layer is used to generate a liner embedding, as described in ViT~\cite{dosovitskiy2020image}.
\end{itemize}
For the fusion process, a simple approach is to directly ensemble the final prediction of the different networks. 
Another method involves concatenating feature maps from different modality networks and using a single network to fuse the features. 
Yet another approach is to use multiple networks to process different features, with interactions between the different networks.
}

{
Some research in emotion recognition focuses on integrating audio and visual inputs.
For example, \citeauthor{tzirakis2017end} attempted to fuse audio signals with facial images~\cite{tzirakis2017end}.
They incorporated an audio stream network to extract audio features, followed by an LSTM network to fuse facial and audio features.
\citeauthor{antoniadis2021audiovisual} combined audio signals with multiple visual modalities, including facial image, body image, and context image~\cite{antoniadis2021audiovisual}.
In contrast to using CNN, \citeauthor{shirian2021dynamic} utilized a GCN structure to process audio, facial, and body inputs, and then used a pooling layer to process the fused features~\cite{shirian2021dynamic}.
}

{
With the rapid advancement of NLP, there has been a surge in research on multimodal sentiment analysis, which incorporates text, audio, and visual input.
Survey papers by \citeauthor{gandhi2022multimodal}~\cite{gandhi2022multimodal} and \citeauthor{zhao2021emotion}~\cite{zhao2021emotion} provided comprehensive overviews of advancements in multimodal sentiment analysis techniques.
Noteworthy methods from recent years include the development of a multitask network by
\citeauthor{akhtar2019multi} that used text, acoustic, and visual frames of videos as input with inter-modal attention modules to adjust the contribution of each modality~\cite{akhtar2019multi}.
Most existing methods adopt multimodal emotion labels as the supervision, ignoring unimodal labels.
\citeauthor{yu2021learning} proposed a label generation module that generated unimodal labels for each modality in a self-supervised manner~\cite{yu2021learning}, enabling
a multitask network to train all unimodal and  multimodal tasks simultaneously.
\citeauthor{jiang2021multitask} adopted several baseline models for each modality input and used PCA to find the optimal feature for the modality~\cite{jiang2021multitask}. Further, an early fusion strategy combined all features.
\citeauthor{yang2022disentangled} and considered the consistency and differences among various modalities in a unified manner, using common and private encoders to learn modality-invariant and modality-unique features, respectively, across all modalities~\cite{yang2022disentangled}.
With a similar motivation, \citeauthor{zhang2022multimodal} proposed a cascade and specific scoring model to represent the inter- and intra-relationship across different modalities~\cite{zhang2022multimodal}.
\citeauthor{zhang2021real} used reinforcement learning and domain knowledge to process fused features from multiple modalities in conversational videos~\cite{zhang2021real}. Through reinforcement learning, Dueling DQN predicted the sentiment of the target sentence based on features of previous sentences. Moreover, information in the first several sentences was used as domain knowledge for subsequent predictions.
\citeauthor{mittal2021multimodal} incorporated multiple visual inputs, resulting in the inclusion of  five distinct types of input: face, body, context, audio, and text~\cite{mittal2021multimodal}.
}

{To show recent advancements in multimodal sentiment analysis, Table~\ref{tab:MMBenchmark} presents the performance of selected representative methods on the widely used benchmark, the MOSEI dataset.  Performance is evaluated through mean absolute error (MAE) and binary accuracy (Acc-2) metrics. The top-performing methods were Self-MM~\cite{yu2021learning} and FDMER~\cite{yang2022disentangled}. Self-MM utilizes a multitask network to train all modalities simultaneously, whereas FDMER employs modules to capture shared and individual features across modalities. The essence of multimodal sentiment analysis remains the optimization of relationships between different modalities.}

{In addition, some research centers on fusing multimodal features. For example, \citeauthor{cambria2013sentic} presented Sentic Blending to fuse scalable multimodal input~\cite{cambria2013sentic}.
In a multidimensional space, it constructed a continuous stream for each modality, which depicted the semantic and cognitive development of humans.
The streams of different modalities were then fused over time.}

\begin{table}[!t]
    \centering
    \caption{Performance comparison of representative methods for multimodal sentiment analysis on the MOSEI dataset, *The results are mainly obtained from FDMER~\cite{yang2022disentangled}.}\label{tab:MMBenchmark}
    \colorbox{lightyellow}{%
    \begin{tabular}{c|c|cc}
    \hline
        Method & Venue & MAE & Acc-2 (\%) \\ \hline
        TFN~\cite{zadeh2017tensor} & EMNLP 2017  & 0.593 & 82.5 \\
        LMF~\cite{liu2018efficient} & ACL 2018  & 0.623 & 82.0 \\
        MFM~\cite{tsai2018learning} & ICLR 2019  & 0.568 & 84.4 \\
        ICCN~\cite{sun2020learning} & AAAI 2020  & 0.565 & 84.2 \\
        MISA~\cite{hazarika2020misa} & ACM MM 2020  & 0.555 & 85.5 \\
        Self-MM~\cite{yu2021learning} & AAAI 2021  & 0.530 & 85.2 \\ 
        FDMER~\cite{yang2022disentangled} & ACM MM 2022  & 0.536 & 86.1 \\ \hline
    \end{tabular}
    }
\end{table}

\section{Modeling Emotion: Significant Technological Barriers}\label{sec:challenge}

Despite advancements propelled by deep learning and big data, solving the problem of emotion modeling is nowhere in sight. In this section, we share insights on some of the most significant technological barriers in computer vision (Section~\ref{sec:vision}), statistical modeling and machine learning (Section~\ref{sec:stat}), AI (Section~\ref{sec:ai}), and emotion modeling (Sections~\ref{sec:demographics},~\ref{sec:disentangle},~\ref{sec:emotionspace}, and~\ref{sec:benchmark}) that hinder progress in the field. At present, there are no straightforward solutions to these issues. Some fundamental technologies in multiple related fields must be further developed before substantial progress can be made in emotion understanding. Emotional intelligence plays a significant role in our cognitive functions such as decision-making, information retrievability, and attention allocation. Artificial emotional intelligence (AEI) will likely become an integral part of future-generation AI and robotics. To put it succinctly, AEI based on visual information is a ``Holy Grail'' research problem in computing.

\subsection{Fundamental Computer Vision Methods}\label{sec:vision}

\subsubsection{Pre-training Techniques}
The task of annotating emotion is a time-consuming process.
Acquiring a large-scale dataset for FER or BEEU on the level of ImageNet is difficult.
To completely overcome the challenge of emotion recognition, DNNs must acquire sufficient representation capability from extensive datasets.
This gap can be bridged by using pretraining techniques, in which a model is initially trained on a massive dataset for an upstream task and then fine-tuned for a downstream task.
During pretraining, the model is expected to learn emotion recognition capabilities from the upstream task.
The choice of upstream task is critical because it determines the amount of emotion-related capabilities the model can acquire.

The widely adopted pretraining strategy is to train the model on ImageNet for image classification tasks.
However, the ARBEE team has demonstrated that this approach does not significantly enhance the performance of the BEEU task~\cite{luo2020arbee}.

Our hypothesis is that image classification and emotion recognition require different types of discriminative features to be extracted. This observation reflects the fact that whereas most individuals can recognize common objects, a portion of the population is unable to discern subtle emotions.
In light of the current state of computer vision, two upstream tasks have been promising for pretraining: self-supervised learning (SSL)~\cite{he2020momentum,he2022masked} and image-text retrieval~\cite{radford2021learning}. 
Regarding SSL, our concern is that despite pretraining with state-of-the-art SSL techniques, the model still required large downstream datasets in the fine-tuning phase. For instance, even after pretraining with SSL, action recognition tasks required fine-tuning on large-scale Kinetics-400 for several epochs to achieve remarkable performance~\cite{feichtenhofer2022masked}. As for image-text retrieval, it is crucial to carefully design the text prompts to effectively bridge an emotion recognition model with a pretrained image-text model. Further research is needed to determine how pretraining techniques can substantially enhance emotion recognition.

\subsubsection{Comprehensive and Robust Body Landmarks}

The effectiveness of current 2-D pose estimation methods has led to their widespread usage in downstream tasks such as action recognition~\cite{sun2019deep, yan2018spatial}.
However, for emotion recognition, these methods face significant limitations.
First, body landmarks used in pose estimation are insufficient for the analysis of subtle emotions, as they do not account for all relevant movements. 
For instance, many methods only provide one landmark on the chest, and this limitation prevents skeleton-based emotion recognition algorithms from leveraging information regarding chest expansion or contraction, which is a common indicator of a person's confidence. 
Similarly, a person can express emotions through finger movements, which are not captured by most pose-estimation algorithms due to being too fine-grained. Developing an emotion-specific pose estimation method would require constructing a new, large-scale annotated dataset with additional landmarks, which can be a time-consuming and costly process.
Second, 2-D pose estimation results can be noisy due to the jitter errors~\cite{ruggero2017benchmarking}.
Although these errors may have a minimal effect onmetrics of pose estimation benchmarks, they can significantly impact understanding of subtle bodily expressions, which demands substantially higher precision of landmark locations~\cite{luo2020arbee}.
Given that pose estimation serves as the starting point for skeleton-based BEEU methods, any errors in human pose can have a ripple effect on a final emotion prediction.

\subsubsection{Accurate 3-D Pose or Mesh In-the-Wild}
The integration of 3-D pose information has the potential to substantially enhance BEEU algorithms because the extra dimension allows for a better understanding of movements in 3-D space. In other words, BEEU algorithms could depend on precise 3-D pose or mesh as input which should enhance its overall accuracy.
Whereas accurate 3-D poses can be obtained through MoCap systems for laboratory-collected data, obtaining accurate 3-D poses for in-the-wild data is challenging.
It is impossible to set up MoCap to capture the 3-D pose of a person because the area of movement is often too large for the placement of MoCap cameras.
Furthermore, existing 3-D pose or mesh estimation approaches have poor performance when applied to in-the-wild images.
The difficulty of collecting 3-D annotations for in-the-wild images has resulted in a lack of large-scale, high-quality 3-D pose datasets.
Current 3-D pose models are heavily reliant on lab-collected 3-D datasets for training, which 
are subject to a distinct domain shift from in-the-wild images.
In a laboratory setting, the lighting and environment are fixed, and the appearance and posture of individuals are monotonous. 
Conversely, in-the-wild images exhibit significant variations in these factors, making it difficult for current models to generalize well to in-the-wild images.

\subsection{Fundamental Statistical Modeling and Learning Methods}\label{sec:stat}

The field of emotion modeling has seen a shift from conventional computer vision and machine learning techniques (e.g.,~\cite{lu2012shape}) to deep learning-based methods (e.g.,~\cite{lu2015deep,luo2020arbee}). For many other computer vision problems, deep learning has shown its power in substantially advancing the state of the art compared to other machine learning methods. However, some intrinsic limitations of deep learning continue to limit progress in the field of emotion modeling.  In this context, we aim to highlight some fundamental data-driven AI capabilities that, if developed, could drive the field forward.

\subsubsection{Modeling a Complex Space with Scarce Data}

A person's emotions and behaviors are influenced by various factors such as personality, gender, age, race/ethnicity, cultural background, and situational context. Modeling this multidimensional, complex space requires a massive amount of data to sample it sufficiently and properly. Despite the potential to collect data from the Internet, manually annotating such a large dataset for emotion can be prohibitively expensive. This conundrum challenges machine learning researchers to find ways to learn meaningful information with limited data collection.

A relatively more specific challenge facing researchers developing datasets for emotion research is determining the appropriate amount of data needed to obtain meaningful results through machine learning. Emotion data, when collected from public sources, tends to be naturally imbalanced, with some emotion categories (e.g., happiness, sadness) having a much higher number of samples than others (e.g., yearning, pain)~\cite{luo2020arbee}. Unlike in typical object recognition where metadata (e.g., keywords, image file names) can aid in crawling a reasonably balanced dataset, we usually cannot determine or estimate the emotion label for a piece of visual data from just the available metadata. Making a balanced, representative dataset depends on crawling or collecting a very large dataset, in the hope of obtaining a sufficient number of samples for the less prevalent categories. Such a laborious process greatly increases the cost of data collection. A potential solution to this challenge is to use AI to guide a more efficient data collection process, so that limited annotation resources can be used to achieve maximum benefit for AI training.

\subsubsection{Explainability, Interpretability, and Trust}

The ability for AI systems to provide clear and interpretable explanations of their reasoning processes is particularly crucial in critical applications, such as healthcare, because end-users, who may not have a data science background, need to have confidence in the AI system's quantitative findings. There is ongoing research in this area within the broader field of AI, with some progress being made~\cite{shin2021effects,arya2020ai,gilpin2018explaining}. However, it is widely considered an open challenge. 

The task of emotion understanding presents unique challenges in terms of explainability, interpretability, and trust. First, emotions are highly subjective, and AI models typically learn from a general population's responses based on data collected from numerous human subjects. The AI's ability to explain or interpret results, thus, is limited to a general perspective, which may naturally be considered unreasonable to a certain extent by virtually any particular individual, especially if their opinions on emotions frequently deviate from the norm.

Second, causal relationships are often more significant in emotion-related applications. For example, in determining why a sunset scene evokes positive emotions in many viewers, it is important to understand whether it is the sunset scenario, the orange hues, the gradient of colors, the horizontal composition, or some other properties that cause viewers to feel positive emotions. Without this knowledge, algorithms developed may behave unpredictably in many situations, acting as black boxes. Currently, methods for causal discovery are often not scalable to high-dimensional or complex data and are sensitive to sparse data. Therefore, there is a need to develop appropriate frameworks and methods for high-dimensional and complex causal discovery specifically tailored to the understanding of emotions.

Third, overlapping semantics among emotion labels add complexity to interpreting results. The BoLD dataset has shown, using video clip annotations, that several pairs of emotion labels are highly correlated~\cite{luo2020arbee}. Examples are pleasure and happiness ($\text{correlation}=0.57$), happiness and excitement ($0.40$), sadness and suffering ($0.39$), annoyance
and disapproval ($0.37$), sensitivity and sadness ($0.37$),
and affection and happiness ($0.35$).
Even in the dimensional VAD model, researchers
have detected correlations between valence and dominance ($0.359$)
and between arousal and dominance ($0.356$). Current data-driven approaches can often provide a probability score for each emotion label in the classification system. Although it remains common practice to sort these scores to determine the most likely emotion, it is not always clear what a mixture of scores represents. For example, what does it mean to have 80\% happiness, 60\% excitement, and 50\% sadness? Is it reasonable to just assume the data point should be classified as happiness? Or, should it be considered a mixture of happiness, excitement, and sadness all at the same time, perhaps as representative of humans often complex emotional states? Or, could it be some partial combination of these categories? More fundamental statistical learning methods will likely be needed to address this issue in a principled way.
 
One potential strategy in automated emotion recognition is to uncover useful patterns from a large amount of data, with the aim of gaining a deeper understanding of emotions beyond simple classification. Such findings may either support or challenge existing psychological theories or inspire new hypotheses. However, black-box machine learning models can provide little insight into why a decision is made, which restricts our ability to gain knowledge through automated learning. As a result, there has been a recent and growing focus on developing interpretable machine learning. 

There are two primary approaches to interpretable machine learning. The first is known as model-agnostic approaches~\cite{ribeiro2016model, ribeiro2016should} and involves using relatively simple models such as decision trees and linear models to approximate locally the output of a black-box model like DNN. A fundamental issue with this approach is that the explanation is local, usually a neighborhood around every input point. This outcome raises the question of whether such limited explanations are truly useful. Because the power of explaining a phenomenon implies the capability to reveal underlying mechanisms coherently for a wide range of cases, a severe lack of generality in interpretation undermines this goal. 

In the second approach, the emphasis is on developing models that are {\it inherently interpretable}. Classic statistical models, due to their simple structure, are often inherently interpretable. However,  their accuracy is often significantly lower compared to top-performing black-box models. This drawback of classic models has motivated researchers to develop models with enhanced accuracy without losing interpretability. For example, \citeauthor{seo2022mixture}~\cite{seo2022mixture} proposed the concept of cosupervision by DNN to train a mixture of linear models (MLM), aimed at filling the gap between transparent and black-box models. The idea is to treat a DNN model as an approximation to the optimal prediction function based on which augmented data are generated. Clustering methods are used to partition the feature space based on the augmented data, and a linear regression model (or logistic regression for classification) is fit in each region. Although MLMs have existed in various contexts, they have not been widely adopted for high-dimensional data because of the difficulties in generating a good partition. The authors overcame this challenge by exploiting DNN. They also developed methods to help interpret models either by visualization or simple description. Advances in the direction of developing accurate models that are directly interpretable are valuable for emotion recognition. 

\subsubsection{Modeling Under Uncertainty}

Data available for modeling human emotions and behaviors is often suboptimal, leading to various challenges in accurate modeling. For instance, in the case of bodily expressed emotion recognition, the available data is often limited to partial body movements (e.g., only the upper body is visible in the video) and may include occlusions (i.e., certain body parts are blocked from view). In addition, the automated detection of human body landmarks in video frames is not always precise. An interesting research direction will be to establish accurate models based on incomplete and inaccurate data. Furthermore, it is important to quantify uncertainty throughout the machine learning process, given that such uncertainties are present at each step. This is necessary in order to effectively communicate results to users and help them make informed interpretations.

\subsubsection{Learning Paradigms for Ambiguous Concepts}

The utilization of traditional learning frameworks, such as DNN, SVM, and classification and regression trees (CART), may prove to be limited in tackling complex problems, such as modeling not-so-well-defined concepts like emotions and movements. Unlike more concrete objects, such as cars and apples, there may not always be a clear ground truth for expression of emotions in a video clip. The ambiguity of establishing a ground truth can be influenced by subjective interpretations of annotators, leading to different viewpoints and no correct interpretation.

Although these traditional learning frameworks are optimized for well-defined concepts, their straightforward application to emotion recognition may lead to incorrect assumptions, such as assuming a majority annotator holds the ground truth. In the case of BEEU, the creation of a mid-layer of movement concepts between pixel-level information and high-level emotions brings forth another challenge. Many movement classes are qualitative in nature (e.g., dropping weight, rhythmicity, strong, light or smooth movement, and space harmony), further complicating the development of accurate models. 
The need for advanced learning paradigms that can effectively tackle the challenges of modeling complex, ambiguous concepts such as emotions and movements has become apparent.

\subsubsection{AI Fairness and Imbalanced Datasets}

In the collection of data regarding human emotions and behaviors, it is not uncommon for certain demographic or emotion/behavior groups to have smaller sample sizes compared to others. For instance, data about White people is often more abundant than data about Black people, and data about happiness and sadness is typically more abundant than data about esteem, fatigue, and annoyance.  Similarly, data across cultural groups varies significantly. Despite being a current area of research in AI~\cite{mehrabi2021survey}, tackling such inadequacies is particularly complex in emotion understanding due to its unique characteristics.

\subsubsection{Efficiency of AI Models}

For some applications, emotion recognition  algorithms must be fit onto robots or mobile devices that are limited by their on-board computing hardware and battery power. Complex problems that require multiple AI algorithms and models to work in concert can be difficult to address in real-time without high-performance GPU/CPU computing hardware. Therefore, it is crucial to simplify the mathematical models while maintaining their accuracy.

Machine learning researchers have derived techniques to simplify the model, including channel pruning techniques. In an earlier attempt, \citeauthor{ye2018rethinking}~\cite{ye2018rethinking} proposed a method for accelerating the computations of deep CNNs, which directly simplify the channel-to-channel computation graph without the need to perform the computationally difficult and not-always-useful task of making high-dimensional tensors of CNN structure sparse. There have been numerous recent studies on pruning neural networks~\cite{frankle2018lottery,hoefler2021sparsity}.

\subsection{Fundamental AI Methods}\label{sec:ai}

These are some fundamental AI components for creating an effective human-AI interaction and collaboration system that can have a significant impact in relevant domains. Simply increasing the amount of data alone may not be sufficient to counterbalance the deficiencies in these fundamental areas of AI. Below we will explore some of these challenges in greater detail.

\subsubsection{Decision-Making Based on Complex Information}

In real-world applications, such as mental healthcare, we often need to incorporate multiple sources of information (e.g., nonverbal cues, verbal cues, health record information, observations over time), and different people may have different sets of information inputs (e.g., some patients may not have observations over time whereas others may not have detailed health records). For example, when a patient with depression visits a clinic, a combination of behaviors, speech, and health record information can be used to make informed decisions. To effectively connect all relevant information, research is required to develop a comprehensive framework. This fundamental area of AI research has the potential to impact many AI applications.

\subsubsection{Integrative Computational Models and Simulations for Understanding Human Communication and Collaboration}

Current research primarily concentrates on analyzing individuals, such as deducing an individual's emotional state from their behavior. However, emotional expressions play a significant role in interpersonal communication and collaboration. The emotional expression of one individual can have a significant impact on the emotions and behavior of those around them. Thus, there is a need for the development of integrative computational models and simulations to investigate the emotional interactions among individuals. As the problem of understanding individual emotions remains unresolved, a comprehensive approach to study interpersonal emotional interactions must consider the uncertainties present in individual-level emotion recognition.

In addition, when robots are integrated into the interaction, the distinction between human-human and human-robot interactions (HRIs) must be weighed. The external design and behavior of robots can vary greatly and offer a much wider range of possibilities than humans, making it challenging to sample comprehensively the potential space of robot interactions. For example, a robot can take on various forms, such as a human, animal, or a unique entity with its own distinct personality. Similarly, robots can move in ways that go beyond human or animal-like motions. whereas research can be conducted with limitations on specific types of robots, the results may not be applicable to other forms of robots.

\subsubsection{Incorporation of Knowledge and Understanding of Cognitive Processes}

When humans interpret emotions through visual media, they rely on their accumulated knowledge and experience of the world around them. However, the same behavior can be interpreted differently depending on the context or situation. Data-driven emotion recognition approaches require vast amounts of training data to be effective, but the countless possible scenarios and contexts can make obtaining such data challenging. To address this issue, AI researchers have been exploring the development of common-sense knowledge representations~\cite{ji2021survey}. Integrating these advances into an emotion recognition framework is an important area of research. In addition, cognitive scientists have gained valuable insights into human cognitive processes through experiments, and incorporating these findings into the design of a next-generation emotion recognition system could be key to its success.

The expression and interpretation of emotions by humans involve various levels and scales, ranging from basic physiological processes impacting a person's behavior to socio-cultural structures that shape their knowledge and actions. Currently, multilevel and multiscale analyses of emotions are a rare occurrence in AI due to the complexity it entails.

\subsubsection{Prediction of Actions}

Most of the current research in the field focuses on emotion recognition using visual information. However, for certain human-AI interaction applications, it is necessary to not only recognize emotions based on past data but also proactively gather information in real-time and make predictions about future events. For example, in the event of a heated argument between two human workers, a robot may need to move closer to better understand the situation and based on changes in the behavior of the workers, it may need to predict any potential danger and take action to resolve the issue. This might entail alerting others or attempting to redirect the attention of involved parties. Research is required to map emotion recognition with appropriate actions, even while acknowledging the inherent uncertainty involved in the process of emotion understanding. Such process is more nuanced than typical scene understanding.

\subsection{Demographics}\label{sec:demographics}

Unlike many computer vision problems such as object detection, when it comes to emotion, we simply cannot ignore the effect of demographics. Emotional responses can vary greatly among different demographic groups, including gender, age, race, ethnicity, education, socioeconomic status, religion, marital status, sexual orientation, health and disability status, and psychiatric diagnosis. Existing machine learning-based recognition technologies are not equipped to effectively handle such a vast array of demographic factors. In the absence of sound methods for evaluation, our brains tend to resort to shortcuts that may not be dependable, in order to conserve energy and navigate difficult situations where solid judgment is not present. To fill the gap, we often rely on stereotypes, heuristics, experience, and limited understanding to gauge emotions in other demographic groups, which may be unreliable. However, when we design AI systems, such shortcuts are not acceptable as mistakes made by machines can have disproportionate negative consequences for individuals and society as a whole. Addressing the issue of demographics in automated emotion understanding will likely remain a persistent challenge in the field.

\subsection{Disentangling Personality, Function, Emotion, and Style}\label{sec:disentangle}

A person's behavior, captured by imaging or movement sensors, is a combination of several elements, including personality, function, emotion, and style. Even if we can find solutions to the problems we have mentioned earlier, separating these elements so that emotional expression can be properly analyzed will remain challenging. For example, the same punching motion would convey different emotions in a volleyball game versus during an argument between two individuals. This single example highlights the need for AI to first understand the purpose of the movement. For the same function, with the exact sequence of movements, two persons with very different personalities and contexts would likely be expressing different emotions or at least different levels of the same emotion. Without knowing people's personality traits, it will be impossible to pinpoint their emotional state. There is a need to advance technology to differentiate between these factors in movements.

While fine-tuning the learned model to a specific person is possible, it usually requires collecting a substantial amount of annotated data from that person, which may not be feasible in practical applications requiring personalization. Further research is necessary to develop methods for personalizing emotion-related models with minimal additional data collection.

\subsection{Partitioning the Space of Emotion}\label{sec:emotionspace}

Thus far, technology developers have mostly relied on psychological theories of emotion, including the various models we discussed. However, these models have limitations that make them not ideal for AI applications. For example, if a model used in an AI program has too many components, the program may struggle to differentiate among them. At the same time, if the model is too simple with too few components, the AI may not be able to fully grasp the human emotion for the intended application. The VAD model offers a solution to this issue, but it is not suitable for AI applications for which specific emotions need to be identified. A deeper understanding of the emotional spectrum in AI will lead to more effective applications.

In a recent study, \citeauthor{wortman2022} articulated that the strongest models needed robust coverage, which meant defining the minimal core set of emotions from which all others could be derived~\cite{wortman2022}. Using techniques from natural language processing and statistical clustering, these researchers showed that a set of 15 discrete emotion categories could achieve maximum coverage. This finding applies across six major languages--Arabic, Chinese, English, French, Spanish, and Russian--they have tested. 
Categories were identified as {\it affable}, {\it affection}, {\it afraid}, {\it anger}, {\it apathetic}, {\it confused}, {\it happiness}, {\it honest}, {\it playful}, {\it rejected}, {\it sadness}, {\it spiteful}, {\it strange}, {\it surprised}, and {\it unhealthy}. A more refined model with 25 categories was also proposed, which included the addition of {\it accepted}, {\it despondent}, {\it enthusiasm}, {\it exuberant}, {\it fearless}, {\it frustration}, {\it loathed}, {\it reluctant}, {\it sarcastic}, {\it terrific}, and {\it yearning} and the removal of {\it rejected}. Through the analysis of two large-scale emotion recognition datasets, including BoLD, the researchers confirmed the superiority of their models compared to existing models~\cite{wortman2022}.

\subsection{Benchmarking Emotion Recognition}\label{sec:benchmark}

Effective benchmarking has been instrumental in driving advancements in various AI research areas. However, benchmarking for emotion recognition is a challenging task due to its unique nature and the obstacles discussed earlier. Below, we offer insights on how to establish meaningful benchmarks for the field of emotion recognition, with a specific emphasis on the relatively new area of BEEU, where benchmarking is currently lacking.

\subsubsection{Benchmark Task Types}
A suite of tasks should be devised, including basic tasks such as single-data-type recognition of emotion (based on video only, images only, audio only, skeleton only, human mesh only), as well as multimodal recognition (a combination of video, audio, and text). Emotion localization, which involves determining the range of frames in a video that depicts a targeted label, as well as movement recognition or LMA recognition using video, skeleton, or human mesh should also be considered. Furthermore, tasks related to predicting emotion from movement coding and video, or based on interaction can be developed.
With the rich data across various contexts, natural environments, or situations (e.g., celebration, disaster, learning),  
data mining tasks focused on social interactions, including the comparison between age groups or the impact of assistive animals on mood, can be explored. In addition, real-world use-case challenges targeting specific applications can be utilized to assess algorithms' broad applicability and robustness.

\subsubsection{Testing and Evaluation} 
In a benchmarking competition, the performance of participating teams' algorithms or systems can be evaluated using various criteria. Along with  standard prediction accuracy based on shared training and testing datasets and the extent of emotion coverage, a system's performance with limited training data and equitable accuracy across different demographic subgroups, such as gender or ethnicity, can also be considered. The competition host can supply training data of varying sizes and required metadata, allowing participating teams to focus on a specific evaluation criterion and compete against others using the same standard. This competition format promotes diverse scientific exploration among participating teams, and collectively, teams focusing on different standards broaden the scope of models being investigated, effectively fostering a form of free-style community collaboration.

\subsubsection{Verification and Validation}
To validate software packages developed by participating teams in the competition, winning teams should be required to deposit their packages on repositories such as GitHub. The competition host should provide guidelines for verification, including compatibility with common computing environments, comprehensive documentation, and clear feedback on the execution status and reasons for any unexpected termination. To maintain fairness, true labels for test cases should not be disclosed prior to the completion of a competition.

Software packages should undergo thorough verification and validation throughout the entire training and testing pipeline. The competition host should replicate the training and testing process provided by each winning team, and the results should be compared to the claimed results. To streamline the validation process, subsampling of test cases may be employed.

\subsubsection{Risk Management} 
To gauge the robustness of winning teams' software packages, they should be asked to provide results from a set of robustness tests, although during the competition, the comparison standard should be based on a single, focused criterion within a relatively straightforward test framework. Specifically, the impact of variations in factors such as batch randomization, bootstrapped sampling of training images, and training data size should be numerically evaluated.

\subsubsection{Evaluation Metric} 
As a person's emotional state does not fall exclusively into a single type, to provide a fair evaluation of algorithms developed by the competition participants, emotions can be characterized by a distribution over a given set of types, allowing each dataset to be described by different (and multiple) types and users to select the set that works best for their methods.
Both the ground truth and the output of emotion recognition algorithms are formatted as distributions over these types. Suppose there are a total of $K$ emotion types denoted by $e_1$, $e_2$, ..., $e_{K}$. Different from a typical classification problem, there is a more subtle relationship between these types. Each pair of emotion types has a specified distance (or similarity) instead of just being different. For example, the emotions ``happy'' and ``sad'' are farther apart than ``happy'' and ``excited''. As a result, when we compare two emotion distributions over these types, we want to account for the underlying distances between emotion types. These pairwise distances can be estimated from the data based on how two emotion types co-occur. We can then use Wasserstein distance~\cite{zhang2015parallel} to compute the overall distance between the ground truth and the computer prediction.
Other commonly used distances between distributions such as $L_p$ norm or KL divergence cannot factor in the underlying distances between emotion types. Let the distance between $e_i$ and $e_j$ be $c_{i,j}$, $i, j =1, ..., K$. Consider two probability mass functions over $\{e_1, ..., e_{K}\}$: $p=(p_1, ..., p_{K})$ and $q=(q_1, ..., q_{K})$. The Wasserstein distance is defined by an optimal transport problem~\cite{zhang2015parallel}. 
Let $\mathbf{W}=(w_{i,j})_{i,j=1, ..., K}$ be a non-negative matching matrix between $p$ and $q$. The Wasserstein distance is:
\begin{eqnarray*}
D(p,q)&=&\min_{\mathbf{W}}\sum_{i=1}^{K}\sum_{j=1}^{K}w_{i,j}c_{i,j}\\
w_{i,j}&\geq& 0\, , i, j=1, ..., K\\
\sum_{i=1}^{K}w_{i,j}&=&q_j \, , j=1, ..., K\\
\sum_{j=1}^{K}w_{i,j}&=&p_i\, , i=1, ..., K \, .
\end{eqnarray*}

\section{Beyond Emotion: Interaction with Other Domains}\label{sec:connect}

Emotion, as one of the core components of human-to-human communication, can play an essential role in an array of future technological advancements impacting different parts of society. In this section, we provide an overview of how visual emotional understanding can be connected with other research problems, domains, or application areas, including art and design (Sections~\ref{sec:visualart} and~\ref{sec:design}), mental health (Section~\ref{sec:health}), 
robotics, AI agents, autonomous vehicles,
animation, and gaming (Section~\ref{sec:robot}),
information systems (Section~\ref{sec:infosys}), industrial safety (Section~\ref{sec:safety}), and education (Section~\ref{sec:education}). Instead of attempting to provide exhaustive coverage, we aim to highlight key intersections. Because some areas are in their early stages of development, we provide only a brief discussion of their potential.

\subsection{Emotion and Visual Art}\label{sec:visualart}

Art often depicts human emotional expressions, conveys the artist's feelings, or evokes emotional responses in viewers. Except for certain genres in visual art, e.g. Realism, achieving lifelikeness is usually not the primary goal. Dutch Post-Impressionist painter Vincent van Gogh wrote, ``I want to paint what I feel, and feel what I paint.'' Similarly, fine-art photographer Ansel Adams stated, ``A great photograph is one that fully expresses what one feels, in the deepest sense, about what is being photographed.'' It is evident that artists intentionally link visual elements in their works with emotions. However, the relationship between visual elements in art and the emotion they evoke is still largely an enigma.

\begin{figure}[ht!]
\setlength{\tabcolsep}{0.8pt}
\begin{tabular}{cc}
  \includegraphics[width=0.49\linewidth]{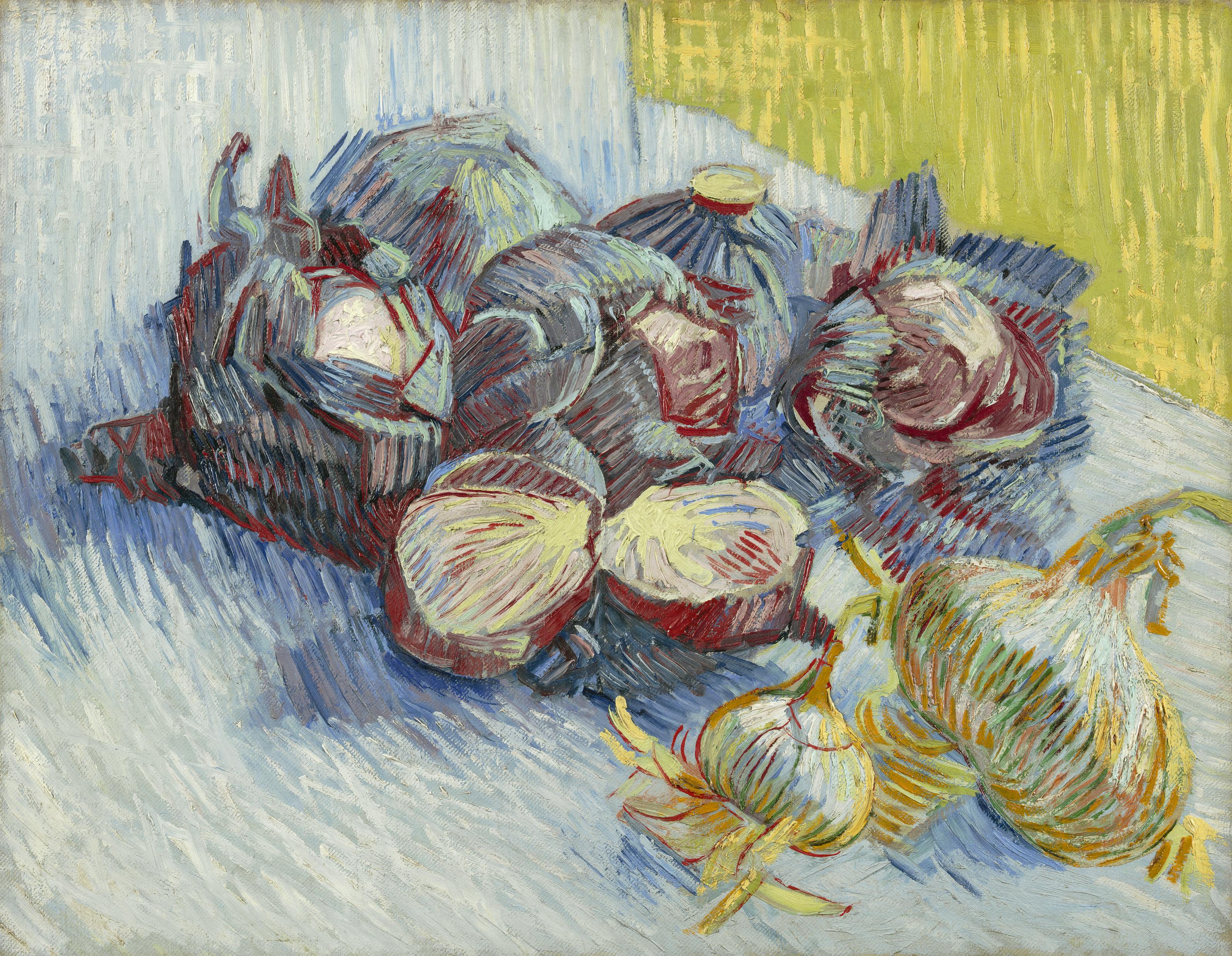}&
  \includegraphics[width=0.485\linewidth,cfbox=lightgray 1pt 1pt]{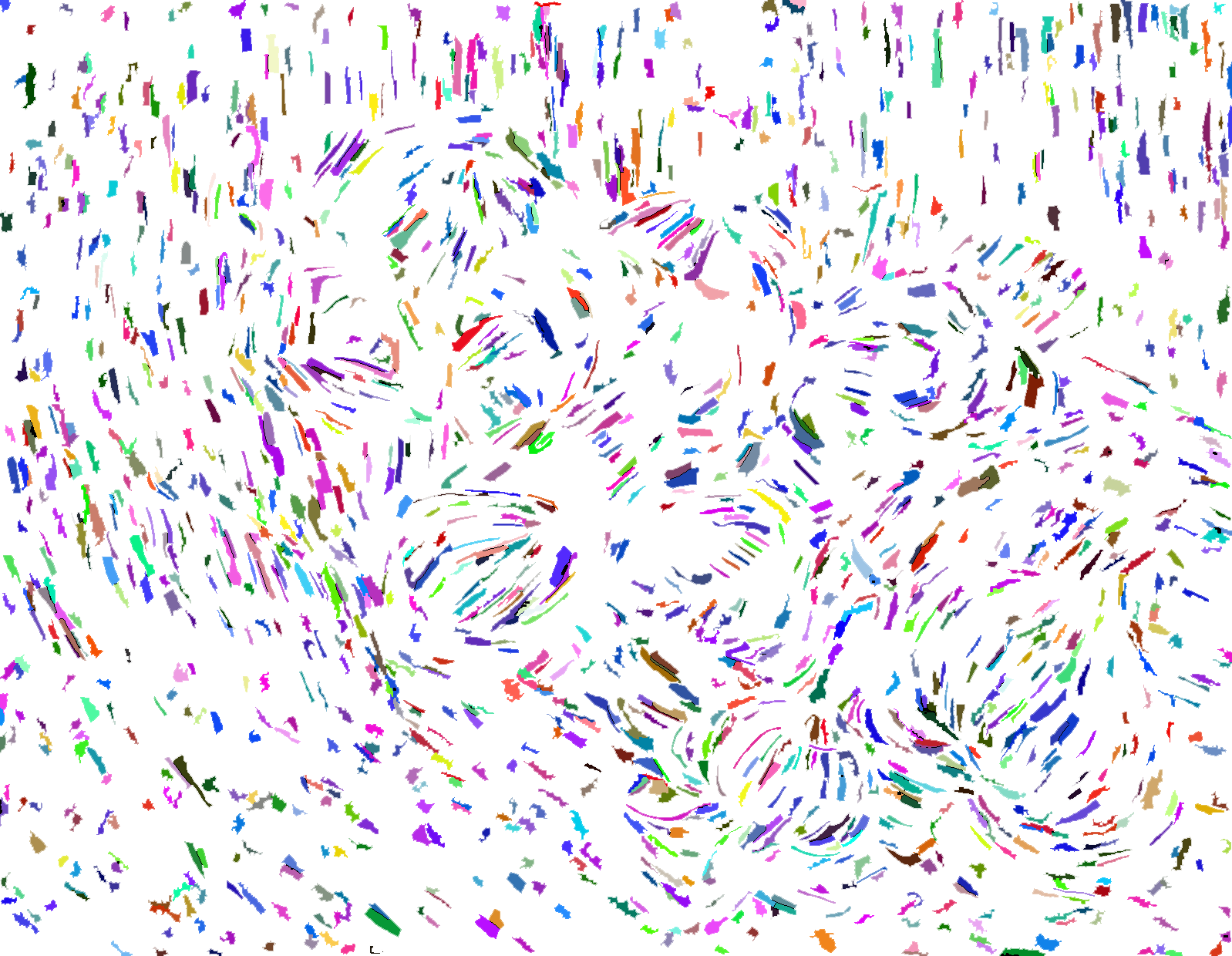}\\
\end{tabular}
\caption{Automatic brushstroke extraction for {\it Red Cabbages and Onions} (Paris, October-November 1887,
oil on canvas, 50.2 cm $\times$ 64.3 cm) by Vincent van Gogh~\cite{li2012rhythmic}. Painting image courtesy of the Van Gogh Museum, Amsterdam (Vincent van Gogh Foundation). The brushstroke map was provided by the James Z.~Wang Research Group (The Pennsylvania State University).}\label{fig:vangogh}
\end{figure}

Because visual artworks are almost always handcrafted by an artist and artists often develop unique styles, artworks are often abstract and difficult to analyze. In 2016, \citeauthor{lu2012shape} extended this research on evoked emotion in photographs~\cite{lu2012shape} to paintings~\cite{lu2016identifying}. They acknowledged that using models developed for photographs on paintings would not be accurate due to different visual characteristics of the two types of images. To address this, they created an adaptive learning algorithm that leveraged labeled photographs and unlabeled paintings to infer the visual appeal of paintings. 

To convey emotion effectively, artists often create and incorporate certain visual elements that are not commonly seen in real-world objects or scenes. An example is van Gogh's highly rhythmic brushstroke style, which has been shown by computer vision researchers \citeauthor{li2012rhythmic} to be one of the key characteristic differences between him and his contemporaries~\cite{li2012rhythmic} (Fig.~\ref{fig:vangogh}). In fact, van Gogh took piano lessons in the period 1883-1885 and in the middle of his painting career. He wrote to his younger brother Theo toward the latter part of his career in 1888, ``......this bloody mistral is a real nuisance for doing brushstrokes that hold together and intertwine well, with feeling, like a piece of music played with emotion.'' 
The study by \citeauthor{li2012rhythmic}~\cite{li2012rhythmic} highlights the importance of designing algorithms specifically to answer the art-historical question at hand rather than using existing computer vision algorithms meant for analyzing real-world scenes. 

Because artwork is the crystallization of the creativity and imagination of artists, studying artwork using computers and modern AI has the potential to reveal new perspectives on the connection between visual characteristics and emotion. 
Artists often incorporate exaggerated visual expressions such as carefully designed color palettes, tonal contrasts, brushstroke texture, and elegant curves. These features have inspired computer scientists to create new algorithms for analyzing visual content. For instance, \citeauthor{yao2012oscar} developed a color triplet analysis algorithm to predict the aesthetic quality of photographs, drawing inspiration from artists' use of limited color palettes~\cite{yao2012oscar}. \citeauthor{li2015photo} created an algorithm for tonal adjustments based on the visual art concept of ``Notan'' which captures the dark and light arrangement of masses~\cite{li2015photo}. Motivated by the use of explicit and implicit triangles in artworks, \citeauthor{he2017discovering} and \citeauthor{zhou2017detecting} developed algorithms to identify triangles in images, which can assist portrait and landscape photographers with composition~\cite{he2017discovering,zhou2017detecting}. 

Current techniques for understanding emotion are not yet capable of analyzing certain aspects of emotion expressed in artwork, particularly at the level of composition or abstraction. For example, when emotions are conveyed through subtle interactions between people, the correspondence between low-level features we can extract and the emotions they represent cannot be easily established. American Impressionist painter Mary Cassatt's work, for example, depicts the love bond between a mother and child through composition, pose, and brushstrokes, rather than through clear facial expressions. Similarly, American Modernist artist Georgia O'Keeffe used dramatic colors, elegant curves, and creative, abstract composition in her paintings of enlarged flowers and landscapes to convey feelings. She stated, ``I had to create an equivalent for what I felt about what I was looking at – not copy it.'' There is still much to be discovered by technology researchers in terms of the systematic connection between visual elements in abstract artwork and the emotions they convey.

\subsection{Emotion and Design}\label{sec:design}

Emotion plays a key role in product design, whether it be for physical or virtual products. Cognitive scientist D. A. Norman was a pioneer in the study of emotional design~\cite{norman2004emotional}. A successful design should evoke positive emotions in users/customers, such as excitement and a sense of pride and identity. In physical products, from a bottle of water to a house, designers carefully select visual elements, such as round corners, simple shapes, and elegant curves, to evoke positive emotions in customers. Similarly, designers of Websites, mobile apps, and other digital products and services use harmonious color schemes, simple and clean layouts, and emotion-provoking photographs to create a positive emotional impact on viewers or users. 

By advancing evoked emotion prediction, future designers can be assisted by computers in multiple ways. First, computers can assess the evoked emotion of a draft design, based on models learned from large, annotated datasets. For example, a Website designer can ask the computer to rate a sample screenshot, identify areas for improvement, and provide advice on how to improve it. To develop this capability, however, researchers need to gain a better understanding of how demographics affect emotion. Certain design elements, e.g. color, may evoke different feelings in different cultures. A system trained with a general population may perform poorly for certain demographic groups. Second, computers can provide designers with design options that not only meet customer needs but also evoke a specific emotion. For example, deep learning and generative adversarial networks (GANs) can already do the former task. If an additional emotion understanding component can be used to assess the design options generated and provide feedback to the system, the resulting designs can then evoke a desired emotion.

\subsection{Emotion and Mental Health}\label{sec:health}

Many mental health disorders can be considered disorders of emotion and emotion regulation~\cite{sheppes2015emotion}. This is because mental health disorders often entail extremes of chronic self-reported distress, sadness, anxiety, or lack of emotions such as flat affect and numbness as well as extremes in fluctuation of emotions~\cite{APA2013}. For example, anxiety disorders are instances of being in an ongoing state of the fight or flight response, often viewing danger where none exists and therefore reacting as though the danger is constantly present and/or overreacting to ambiguous information. Emotion-related symptoms of anxiety disorders include panic attacks, fear, feelings of impending danger, agitation, excessive anxiety, feeling on edge, irritability, and restlessness~\cite{APA2013}. Major depressive disorder has been conceptualized as a disorder of sustained negative affect, particularly sadness, and low levels of positive affect~\cite{joormann2016examining,vanderlind2020understanding}. Emotion-related symptoms of major depression include feeling sad or down most of the day nearly every day for at least two weeks and can also include an abundance of guilt, agitation, excessive crying, irritability, anxiety, apathy, hopelessness, loss of interest or pleasure in activities, mood swings, and feelings of restlessness. Similarly, bipolar disorder can include mood swings, elevated sadness, anger, anxiety, apathy, apprehension, euphoria, general discontent, guilt, hopelessness, loss of interest or pleasure, irritability, aggression, and agitation~\cite{APA2013}. Extreme mood swings are also a prominent feature for some personality disorders such as borderline personality disorder, which also entails intense depressed mood, irritability or anxiety lasting a few hours to a few days, chronic feelings of emptiness, intense or uncontrollable anger, shame, and guilt~\cite{APA2013}. Schizophrenia is associated with symptoms that are associated with mood. Positive symptoms can include delusions or paranoia, feeling anxious, agitated, or tense, and being jumpy or catatonic. Negative symptoms can include lack of interest or enthusiasm, lack of drive, and being emotionally flat~\cite{APA2013}. Because extreme emotions are associated with these disorders researchers have examined ways to identify important distinctive features of them from videos. 

Such studies have examined ways to use machine learning to code videos for nonverbal behaviors from facial expressions, body posture, gestures, voice analysis, and motoric functioning to diagnose mental health problems. In terms of facial expressions, studies have found that people with major depression, bipolar disorder, and schizophrenia demonstrated less facial expressivity compared to individuals without these disorders~\cite{gaebel1992facial,gaebel2004facial,bersani2013facial}. Depressed compared to non-depressed individuals also evidenced shorter durations and lower frequency of smiling behavior, less looking at an interviewer, and less eyebrow movement~\cite{jones1979some,troisi1999gender,sobin1997psychomotor}. Such differences have been used to discriminate depressed from nondepressed individuals~\cite{jan2018artificial,zhu2018automated,kulkarni2018clinical,he2019automatic,song2018human,nasser2020review} Studies have similarly diagnosed differential facial movement features of people with disorders such as autism spectrum disorder~\cite{mengi2021artificial}, posttraumatic stress disorder, and generalized anxiety disorder from healthy controls~\cite{gavrilescu2019predicting}. Similar to facial emotion, studies have used linguistic and voice emotion analysis to detect disorders such as depression, schizophrenia, bipolar disorder, posttraumatic stress disorder, and anxiety disorders~\cite{low2020automated,morales2017cross,parola2020voice}. 

As with facial actions, gestures and body movement have also been examined. In terms of gestures, those with depression showed more self-touching than those without depression~\cite{jones1979some,sobin1997psychomotor,schrijvers2008psychomotor}. Compared to schizophrenic individuals, those with depression tended to make fewer hand gestures~\cite{jones1979some}. Bipolar and depressed people also showed less gross motor activity than those without these disorders~\cite{sobin1997psychomotor}. However, depressed and bipolar individuals showed more gross motor activity than people with schizophrenia~\cite{sobin1997psychomotor}. At the same time, patients with schizophrenia demonstrated fewer hand gestures~\cite{annen2012nonverbal}, fewer small and large head movements, and shorter duration of eye contact compared to those with depression~\cite{jones1979some,sobin1997psychomotor,davison1996facial}. Additional studies have detected attention deficit hyperactivity disorder from gestures and body movements~\cite{mengi2021artificial}. Such differences have been used to diagnose mental health problems~\cite{pampouchidou2017quantitative,kacem2018detecting}. See Table~\ref{tab:mental} for more details about differentiating clinical disorders from
healthy controls.

\begin{table*}[ht!]
    \centering
        \caption{Distinct emotion-related nonverbal behaviors in mental health disorders compared to healthy controls.}
    \label{tab:mental}
\colorbox{lightyellow}{%
\begin{tabular}{|wl{0.46\linewidth} wl{0.48\linewidth}|}
\hline
        & \\[-1.5ex]
\multicolumn{2}{|c|}{\bf Major Depressive Disorder} \\ [0.5ex]
\tabitem Reduced facial expressivity~\cite{gaebel1992facial,gaebel2004facial} & \tabitem Reduced variability of head movements~\cite{girard2014nonverbal,scherer2013audiovisual} \\
\tabitem Less eyebrow movements~\cite{jh2012verbal,jones1979some,schelde1998major,segrin2000social} & 
\tabitem More nonspecific gaze patterns~\cite{schelde1994ethology}\\
\tabitem Looking-down behaviors~\cite{schelde1994ethology} & \tabitem Less eye contact with another person~\cite{jones1979some}\\
\tabitem Reduced hand gestures~\cite{jh2012verbal,segrin2000social} &
\tabitem Less smiling~\cite{jh2012verbal,jones1979some,schelde1998major,scherer2013automatic,segrin2000social,sobin1997psychomotor,troisi1999gender}\\
\tabitem More self-touching~\cite{jones1979some,scherer2013automatic,segrin2000social,sobin1997psychomotor} &
 \tabitem Slower voice~\cite{cummins2015review} \\ 
 \tabitem Reduced rate of speech~\cite{cummins2015review} & \tabitem More monotonic voice~\cite{horwitz2013relative,kiss2017mono,low2020automated,quatieri2012vocal} \\ 
 \tabitem Reduced speech~\cite{cummins2015review} & \tabitem Reduced pitch range~\cite{cummins2015review}\\
   \tabitem Slower movements or abnormal postures~\cite{buyukdura2011psychomotor,schrijvers2008psychomotor} &
\tabitem Reduced gross motor activity~\cite{jh2012verbal,parker1990classifying,sobin1997psychomotor} \\
\tabitem Reduced stride length and upward lifting motion of legs~\cite{lemke2000spatiotemporal,sloman1982gait}&
\tabitem Slower Gait speed~\cite{hausdorff2004gait,lemke2000spatiotemporal,michalak2009embodiment} \\ 
\tabitem Arm swing and vertical head movements while walking~\cite{michalak2009embodiment}
& \tabitem Lateral upper body sway while walking~\cite{michalak2009embodiment}
\\ 
\tabitem Slumped posture~\cite{michalak2014sitting,michalak2009embodiment,wilkes2017upright}
& \tabitem Forward inclination of head and shoulders~\cite{canales2017investigation,rosario2014differences}
\\
\tabitem Balance difficulties during motor and cognitive tasks~\cite{deschamps2015balance,doumas2012dual,michalak2009embodiment,nakano2014physical,radovanovic2014gait}
& \\
\tabitem Impaired balance and lower gait velocity~\cite{lemke2000spatiotemporal,michalak2014sitting,michalak2009embodiment,sanders2010gait}
&
\tabitem Difficulty recognizing emotions~\cite{bora2016theory,van2021deficiencies}  \\
        & \\[-1.5ex]
\hline
        & \\[-1.5ex]
\multicolumn{2}{|c|}{\bf Bipolar Disorder}\\ [0.5ex]
 \tabitem Reduced levels of facial expressivity~\cite{aghevli2003expression} &
 \tabitem Greater speech tonality~\cite{guidi2017features,guidi2015analysis,zhang2018analysis}
\\ 
\tabitem Less gross motor activity~\cite{sobin1997psychomotor} &\\
 \tabitem More frequent and longer speech pauses when in depressive states~\cite{guidi2017features,maxhuni2016classification} &
\\
\tabitem More postural sway~\cite{bolbecker2011postural}
& \tabitem Difficulty recognizing emotions~\cite{baez2013contextual,donohoe2012social}\\
        & \\[-1.5ex]
\hline
        & \\[-1.5ex]
\multicolumn{2}{|c|}{\bf Schizophrenia}\\[0.5ex] 
 \tabitem Reduced facial expressivity~\cite{berenbaum1992emotional,gaebel1992facial,gaebel2004facial,kring1999stability}
& \tabitem Less upper facial movement expressing positive emotion~\cite{gaebel2004facial,juckel1998fine,krause1989facial}
\\ \tabitem Less smiling~\cite{jones1979some,steimer1990interaction,troisi2007facial} &\\
 \tabitem Reduced smiling eye gaze and head tilting associated with negative symptoms~\cite{lavelle2013nonverbal,troisi1999ethological}&
\\ \tabitem Fewer hand gestures when speaking~\cite{annen2012nonverbal,brune2008nonverbal,lavelle2013nonverbal,troisi1998non,troisi1999ethological}
& \tabitem Fewer gestures and poses~\cite{brune2008nonverbal}
\\ \tabitem Less head nodding~\cite{annen2012nonverbal,lavelle2013nonverbal}
& \tabitem Less head movement~\cite{davison1996facial,jones1979some}
\\ \tabitem Lower total time talking~\cite{kliper2015prosodic,kliper2010evidence,parola2020voice}
& \tabitem Slower rate of speech~\cite{perlini2012linguistic,tahir2019non}
\\ \tabitem Longer speech pauses~\cite{kliper2015prosodic,kliper2010evidence,parola2020voice,rapcan2010acoustic}
& \tabitem More pauses~\cite{parola2020voice}
\\ \tabitem Flat affect~\cite{parola2020voice}
& \tabitem Forward head posture and lower spine curvature~\cite{cristiano2017postural}
\\ \tabitem Balance difficulties and increased postural sway, 
~\cite{kent2012motor,marvel2004quantitative,matsuura2015standing,teng2016postural} 
& \tabitem Difficulty walking in a straight line~\cite{jeon2007quantitative,lallart2014gait}
\\ \tabitem Slower velocity of walking and shorter strides~\cite{putzhammer2005gait}
& \tabitem Difficulty recognizing emotions~\cite{baez2013contextual,donohoe2012social,sparks2010social}\\
        & \\[-1.5ex]
\hline
        & \\[-1.5ex]
\multicolumn{2}{|c|}{\bf Anxiety Disorders}\\ [0.5ex]
\tabitem Less eye contact~\cite{gilbert1991physiological,wenzel2005communication,wiens1980personality}
& \tabitem Instability of gaze direction~\cite{laretzaki2011threat}
\\ \tabitem Grimacing~\cite{gilbert1991physiological}
& \tabitem Nonsymmetrical lip deformations~\cite{metaxas2004image}
\\ \tabitem Strained face~\cite{hamilton1959assessment}
& \tabitem Eyelid twitching~\cite{hamilton1959assessment} 
\\ \tabitem Smiled less,~\cite{wenzel2005communication}
& \tabitem More frequent and faster head movements~\cite{dinges2005optical,hadar1983head,liao2005real}
\\ \tabitem More and faster blinking~\cite{dinges2005optical,giannakakis2017stress,harrigan1996detecting,harris1966blink}
& \tabitem Nodded less~\cite{wenzel2005communication}
\\ \tabitem Small rapid head movements~\cite{giannakakis2017stress}
& \tabitem Made fewer gestures~\cite{wenzel2005communication}
\\ \tabitem More physical movements indicative of nervousness (e.g., bouncing their knees, fidgety, reposuring~\cite{ekman1974detecting,heerey2007interpersonal,jurich1974correlations,lecompte1981ecology,shechter2022man,wenzel2005communication} &
\\ \tabitem Self touching~\cite{ekman1974detecting}
& \tabitem Speech errors~\cite{harrigan1994role}
\\ \tabitem Speech dysfluency~\cite{gilbert1991physiological}
& \tabitem More jittery voice~\cite{ozseven2018voice,silber2016social}
\\ \tabitem Slow gait velocity associated with fear of falling~\cite{reelick2009influence,staab2013threat,wynaden2016recognising}
& \tabitem Balance dysfunction~\cite{balaban2002neural,bart2009balance,bolmont2002mood,feldman2019gait,hainaut2011role}
\\ \tabitem Slower speed walking~\cite{feldman2019gait}
& \tabitem Shorter steps~\cite{feldman2019gait}
\\ \tabitem Enhanced recognition of anxiety~\cite{surcinelli2006facial,zainal2018worry} & \\
        & \\[-1.5ex]
\hline
        & \\[-1.5ex]
\multicolumn{2}{|c|}{\bf Posttraumatic Stress Disorder}\\ [0.5ex]
\tabitem Monotonous slower flatter speech~\cite{low2020automated,marmar2019speech,scherer2015self,xu2012voice}
& \tabitem Reduced facial emotion~\cite{stratou2013automatic}
\\ \tabitem More anger, aggression, hostility, less joy~\cite{kirsch2007facial,stratou2013automatic} &\\
        & \\[-1.5ex]
 \hline
         & \\[-1.5ex]
\multicolumn{2}{|c|}{\bf Autism Spectrum Disorder}\\ [0.5ex]
 \tabitem Distinctions in gait~\cite{biffi2018gait}
& \tabitem Difficulty recognizing emotions~\cite{baron2000theory}
\\  [1ex]
\hline
    \end{tabular}
    }% end of color box
\end{table*}

Gait, balance, and posture have also been used to identify mental health problems. For example, one meta-analysis summarized 33 studies of gait and balance in depression~\cite{murri2020instrumental}. Depressed individuals had worse and more slumped posture~\cite{rosario2014differences,wilkes2017upright,canales2010posture,michalak2009embodiment} and more postural instability and control~\cite{doumas2012dual}. In terms of gait, compared to healthy controls, those with depression took shorter strides, lifted their legs in an upward as opposed to a forward motion~\cite{sloman1982gait, lemke2000spatiotemporal}, had more body sway, and walked more slowly, possibly to maintain their balance~\cite{michalak2009embodiment,lemke2000spatiotemporal,radovanovic2014gait,hausdorff2004gait}. These results are consistent with psychomotor retardation, a common symptom of depression. In terms of anxiety disorders, a study showed that these individuals walked more slowly, took shorter steps, and demonstrated problems with balance and mobility~\cite{feldman2019gait,feldman2020gait}. Studies have also used gait to identify bipolar disorder~\cite{kang2018motor}, autism spectrum disorders~\cite{mengi2021artificial}, and attention deficit hyperactivity disorder~\cite{mengi2021artificial}. 

In addition to mental health problems being disorders of emotional expression and experience, mental health problems can also be considered to be disorders of emotion recognition and understanding leading to social deficits. Understanding one's own and others' emotion has been termed theory of mind. Tests for theory of mind can include either identifying emotions by looking at images of faces with various emotional expressions (sometimes with parts of the faces obstructed), or watching a video of an interpersonal interaction and answering questions about various people's emotions and intentions. Having difficulties of theory of mind has been associated with depression~\cite{bora2016theory,van2021deficiencies}, social anxiety disorder~\cite{hezel2014theory}, obsessive-compulsive disorder~\cite{yazici2019decreased}, schizophrenia~\cite{van2021deficiencies,brune2005theory}, bipolar disorder~\cite{van2021deficiencies,montag2010theory}, and autism spectrum disorders~\cite{baron2000theory}. For example, both schizophrenic and bipolar patients showed emotional reactivity that was discordant to emotional videos~\cite{bersani2013facial}. Studies have also examined videotapes of facial emotional reactions to emotionally evocative videos as a means of diagnosing mental health problems. For example, using this technique, those with autism spectrum disorders demonstrated impairment in their ability to recognize emotions from body gestures~\cite{atkinson2009impaired,jarraya2021comparative}. Thus, emotion regulation, emotional understanding, and emotional reactivity can be impaired in those with mental health problems. Such impairment, however, can be used to create systems to automatically detect the presence of these emotional disorders. See Table~\ref{tab:mental} for more details.

\subsection{Emotion and Robotics, AI Agents, Autonomous Vehicles, Animation, and Gaming}\label{sec:robot}

A natural application domain for emotion understanding is robotics and AI. 
In science fiction films, robots and AI agents are often depicted as having a high level of EQ, such as R2-D2, T-800, and Wall-E. They are able to understand human emotion, effectively communicate their own emotional feelings, engage in emotional exchanges with other robots or humans, take appropriate actions in challenging situations or conflicts, and so on. The idea of empowering robots and AI with this level of EQ is widely seen as a desirable and ultimate goal, or a ``Holy Grail.'' 

Some recent surveys studied the field of robotics and emotion~\cite{savery2021robots,cavallo2018emotion,marcos2021emotional}, covering topics such as advanced sensors, latest modeling methods, and techniques for performing emotional actions. Research in the fields of HRI, human-machine interaction (HMI), and human-AI interaction is highly relevant. However, because BEEU is in its infancy and is considered a bottleneck technology, we have yet to see its applications in robotics.

If we can effectively model human emotions through both facial and bodily expressions, robots can work more effectively with human counterparts. Humans would be able to communicate with robots in a way similar to how emotions are used in human-to-human communication. For example, when human workers in a warehouse want to stop a fast-moving robot, they could wave their hands swiftly to signal distress. Similarly,  pedestrians could wave their hands to signal to a self-driving vehicle on a highway that there is an accident ahead, and cars should slow down to avoid a collision. In such situations, traditional forms of communication such as speech and facial expressions may not effectively convey a sense of urgency.

Effective emotional communication can help us understand the intention of robots or AI. For instance, emotionless robots can be perceived as unfriendly or unsympathetic. In certain robotic or AI applications, such as companion robots or assistive robots, it is desirable to project a compassionate and supportive image to establish trust and cooperation between the device and humans interacting with it. Researchers have begun to investigate the relationship among robotics, personality, and motion~\cite{agnihotri2020distinguishing}.

In animated films, robots can display emotional behaviors, but these are often created by recording the movements of human actors through MoCap. That is, the animated characters mimic the movements of the actors behind the scenes. However, the capacity for computers to comprehend emotions akin to human perception could enable animated characters to use an emotion synthesis engine to autonomously generate authentic emotional behaviors. Advancements in computer graphics, virtual reality, and deep learning techniques including GANs, transformers, diffusion models, and contrastive learning have facilitated the creation of increasingly realistic and dynamic visual content. These technologies potentially enable the synthesis of complex and nuanced emotional displays.

{Emotion understanding can substantially enhance the gaming experience by making games more emotionally responsive and immersive, as well as by providing personalized feedback to players. Game designers can make design decisions that enhance a player's experience based on the player's frustration level. If players are feeling sad, the game could offer them a story-based scenario that is more emotionally uplifting. By providing meaningful feedback, players are more likely to stay engaged with the game, improving their overall experience.}

\subsection{Emotion and Information Systems}\label{sec:infosys}

Emotion understanding can play a pivotal role in advancing information systems. Currently, when searching for visual content in online collections, we primarily rely on keyword-based metadata. Whereas recent developments in deep learning have enabled information systems to search using machine-generated annotations, these annotations are typically limited to identifying objects and relationships (e.g. a boy in a yellow shirt playing soccer). 

\begin{figure}[ht!]
\setlength{\tabcolsep}{2pt}
    \centering
    \begin{tabular}{cc}
    \includegraphics[trim=0 0 0 0,clip,width=0.485\linewidth]{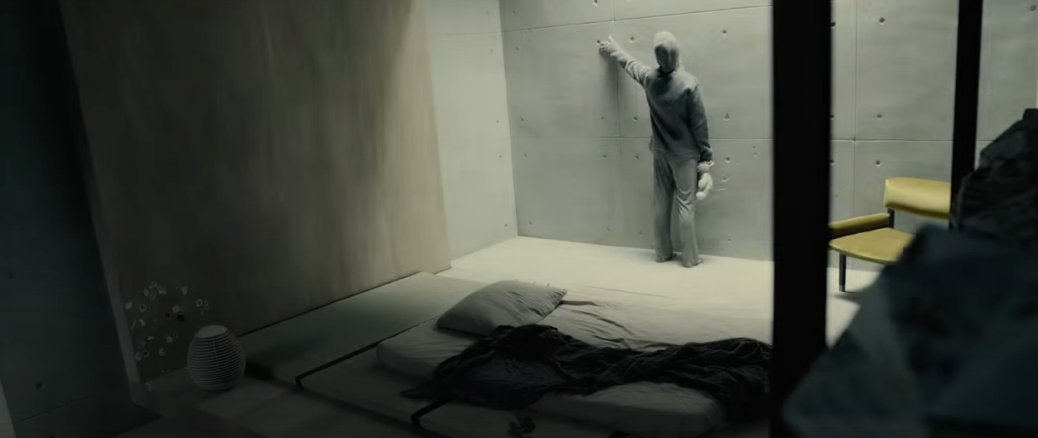} &
        \includegraphics[trim=0 0 0 0,clip,width=0.485\linewidth]{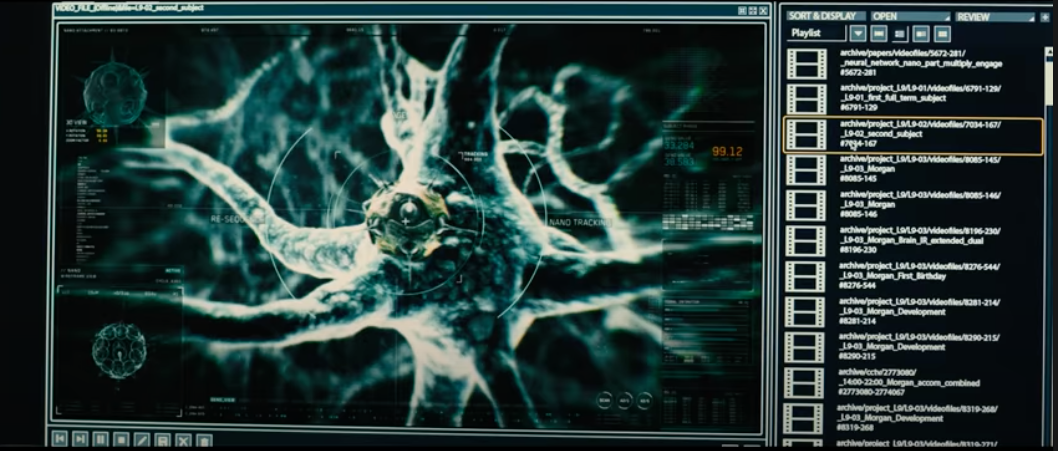} 
    \end{tabular}
    \caption{Two scenes included in the computer-generated trailer for the sci-fi thriller ``Morgan''~\cite{smith2017harnessing}. The computer program considered these scenes most suited to give viewers a sense of horror or thrill. 
    The copyright of these movie frames belongs to 20th Century Fox. They are used here for illustrating a scientific concept only.}\label{fig:morgan}
    \end{figure}

IBM scientists demonstrated that computers with the ability to understand emotions could aid in sorting through a large amount of visual content and composing emotion-stimulating visual summaries~\cite{smith2017harnessing}. In 1996, they created the first computer system for generating movie trailers. The trailer it produced for the 20th Century Fox sci-fi thriller ``Morgan'' was released as the official trailer. The system identified the top ten moments for inclusion in the trailer (Fig.~\ref{fig:morgan}). This work represents a significant milestone in understanding evoked emotions.

IBM's program is likely to be the starting point for a surge of emotion-based information systems. We can expect to see new applications such as evoked emotion assessment systems, emotion-based recommender systems, emotion-driven photo/video editing software, emotion-based media summarization/synopsizing services, and so on.

\subsection{Emotion and Industrial Safety}\label{sec:safety}

Emotion understanding can be useful in promoting safety in workplaces such as factories and warehouses. {It can provide early warnings of potential safety risks, such as worker fatigue or stress, allowing managers to take proactive measures to address the situation. Such capabilities can also provide personalized support and resources to workers who are experiencing emotional distress, improving overall emotional well-being and contributing to a safer work environment.} The National Safety Council estimates that fatigue costs employers over \$130 billion annually in lost productivity, and over 70 million Americans have sleep disorders~\cite{fatigue_cost_nodate}.
Existing research on fatigue detection typically involves  specialized sensors or vision systems that monitor the face or eyes~\cite{baghdadi_monitoring_2019, dow_driver_2013, 8923100, 8482470, 8095346, yamada2017fatigue, yamada_detecting_2018}. However, sensor-based approaches have limitations such as the need to wear them, size, cost, and reliability. Thus, there is a need to develop recognition systems that use body movement to enhance such systems~\cite{baghdadi_monitoring_2019, noakes_fatigue_2012}. 

\subsection{Emotion and Education}\label{sec:education}

Emotion recognition technology can help create a more engaging and effective learning experience for online education. Many universities have been offering online courses for years, but the COVID-19 pandemic led to widespread adoption of online teaching using video conferencing platforms in 2020 and 2021. Even as in-person instruction has resumed, many educational institutions continue to conduct some of their teaching activities online. For example, instructors may be allowed to teach a portion of their classes online for pedagogical or emergency reasons or to hold office hours online. In a traditional classroom setting, instructors can gauge students' attentiveness and emotional states by observing their facial and bodily expressions. Such feedback can help instructors better understand and respond to students' needs, e.g., by adjusting the pace of instruction or covering alternative materials. However, in an online teaching environment, instructors often can only see the faces of a small number of students, which does not provide real-time feedback on the instruction. Potentially, if an online teaching platform could dynamically monitor the students and provide aggregated feedback (e.g., the percentage of students with high attentiveness, the overall mood of the class) to the instructor, the quality of online learning could be improved. To protect students' privacy, the monitoring process should only produce overall statistics.

\section{Emotion and Society: Ethics}\label{sec:ethics}

New technologies often bring new ethical concerns. In the field of emotion understanding, we have begun to witness the potential misuse of these technologies. In this section, we will discuss some of the general ethical issues surrounding the development and deployment of these technologies.

\subsubsection{Generalizability} 
Because the emotion space is complex, it is important to recognize that there will always be outliers or unusual situations (e.g., an otherworldly scene, or an eccentric behavior) that may not be captured by our models. Without proper consideration of demographic differences and individual variations, these technologies may only provide a broad overview of the general population. To be truly beneficial, the system must be carefully tailored to specific needs of an individual. Likewise, diversity of representation in sample datasets is critical to ensure that algorithms emerging from them are inclusive. 

\subsubsection{Verification of Accuracy or Performance} 
It is important for researchers to keep in mind that there is {\it almost always} a lack of ground truth in emotion understanding. We have discussed the impact on data collection, modeling, and evaluation/benchmarking earlier (Sections~\ref{sec:reliability},~\ref{sec:stat}, and~\ref{sec:benchmark}). For AI models, it is imperative that both the output, design, and training processes are transparent and auditable. Black-box models can become uncontrollable if not properly monitored.
		
\subsubsection{Privacy -- Data Collection} 
The collection of human behavior data, including facial, body, and vocal information, raises privacy concerns. Research involving sensitive populations, such as patients in psychological clinics, must be conducted with utmost care. Further, almost all emotion-related annotations must be collected from humans. To protect human subjects, research protocols must be carefully designed to collect only necessary information, de-identify before distribution, and protect the data with proper access control and encryption. All protocols must be reviewed by an Institutional Review Board.

\subsubsection{Privacy -- Use of Technology.}
In today's automated world, people are losing their privacy to whoever controls data: Governments and companies are collecting data about where we are at any given moment; our financial transactions are followed and verified; companies are collecting data about our purchases, preferences, and social networks; most public places are constantly videotaped; and so on. People must sacrifice privacy to live a normal life because everything is computerized. We are being followed, and
``Big brother'' knows all about us. The only thing we can still keep to ourselves is our thoughts and emotions. 

Once AI reaches high-accuracy automatic emotion recognition, our emotions will not be private anymore, and videos of our movements could be used against us by authorities or whoever will have videos of our movements. This situation could become very frightening. Moreover, if we want to hide our emotions, we will have to move in a way that will not reveal them, like using a ``Poker face'' to hide facial expressions. However, because specific movements not only express associated emotions but also enhance those emotions~\cite{shafir2016emotion}, moving in ways that flatten emotional expressions can also flatten the felt emotions, and living in such a way can lead to depression or other mental health problems. 
%Thus, like many other technological advancements these days, the development of automated emotion recognition should be accompanied by worldwide laws and regulations that will guarantee proper and ethical use of this technology. 

\subsubsection{Synthesized Affective Behavior} As much as its potential use in entertainment, success in emotion modeling could inevitably lead to even more lifelike deepfakes and similar abuses. As a society, instead of being fearful of the negative impact of new, beneficial technologies, we need to take on the challenge of detecting fakes, much as we recognize scammers, and mitigating the harm. 

\subsubsection{Lower the Risks.} To mitigate the risks of misuse, proactive measures must be taken. It is essential that laws and regulations are established to keep pace with the rapid development of AEI technologies. As researchers, we have a responsibility to involve affected communities, particularly those that are traditionally marginalized such as minority groups, elderly individuals, and mental health patients, in design, development, and deployment processes to ensure that these individuals' perspectives and needs are recognized and valued.

\subsubsection{Performance Criteria} To promote a responsible and ethical expansion of the field, it is crucial to establish reliable mechanisms for comparing the predictions of different algorithms and learning procedures. As discussed in Section~\ref{sec:benchmark}, a thorough evaluation of algorithms should not only consider accuracy and speed but also factors such as interpretability, demographic representation, context coverage, emotion space coverage, and personalization capabilities.

\section{Conclusion}

We provided an overview of the stimulating and exponentially growing field of visual emotion understanding. Adopting a multidisciplinary approach, we discussed the foundational principles guiding technological progress, reviewed recent innovations and system development, identified open challenges, and highlighted potential intersections with other fields. Our objective was to provide a comprehensive introduction to this vast field sufficient to intrigue researchers across related IEEE subcommunities and to inspire continued research and development towards realizing the field's immense potential. 

Given the multidisciplinary nature of this field, which encompasses multiple technical fields, psychology, and art, the barrier to entry can be considerable. Our aim is to provide researchers and developers with the essential knowledge required to tackle the numerous attractive open problems in the field. Interested readers are encouraged to delve deeper into the cited references for a more profound understanding of the topics discussed. 

As active researchers in this domain, we strongly recommend that those interested in pursuing this research topic collaborate with others possessing complementary expertise. Although we anticipate the development and sharing of more large-scale datasets and continuous incremental progress, transformative solutions will not arise solely from straight application of data-driven approaches. Instead, sustained collaboration among researchers in computational, social and behavioral, and machine learning and statistical modeling fields will likely lead to lasting contributions to this intricate research field.

\section*{Acknowledgment}

The authors would like to acknowledge the valuable contributions of several colleagues and advisees to this work. In particular, we would like to thank Jia Li for contributing valuable ideas and some writing related to explainable machine learning and benchmarking. They also extend appreciation to Hanjoo Kim, Amy LaViers, Xin Lu, Yu Luo, Yimu Pan, Nora Weibin Wang, Benjamin Wortman, Jianbo Ye, Sitao Zhang, and Lizhen Zhu for their research collaboration or discussions. They also thank Bj\"oern W. Schuller and Matti Pietik\"ainen for organizing this special issue on a timely and impactful topic. In addition, they would also like to express the gratitude to the anonymous reviewers for their valuable insights and constructive feedback, which contributed to the improvement of the manuscript. 
J. Z. Wang and C. Wu used the Extreme Science and Engineering Discovery Environment, which was supported by NSF under Grant ACI-1548562, and the Advanced Cyberinfrastructure Coordination Ecosystem: Services \& Support (ACCESS) Program supported by NSF under Grant OAC-2138259, Grant OAC-2138286, Grant OAC-2138307, Grant OAC-2137603, and Grant OAC-2138296.
J. Z. Wang is grateful for the support and encouragement received from Adam Fineberg, Donald Geman, Robert M. Gray, Dennis A. Hejhal, Yelin Kim, Tatiana D. Korelsky, Edward H. Shortliffe, Juan P. Wachs, Gio Wiederhold, and Jie Yang throughout the years.

\bibliographystyle{IEEEtranN}

\renewcommand*{\bibfont}{\scriptsize}
\bibliography{main}

\begin{IEEEbiography}[{\includegraphics[width=1in,height=1.25in,clip,keepaspectratio,trim=1 12 0 0]{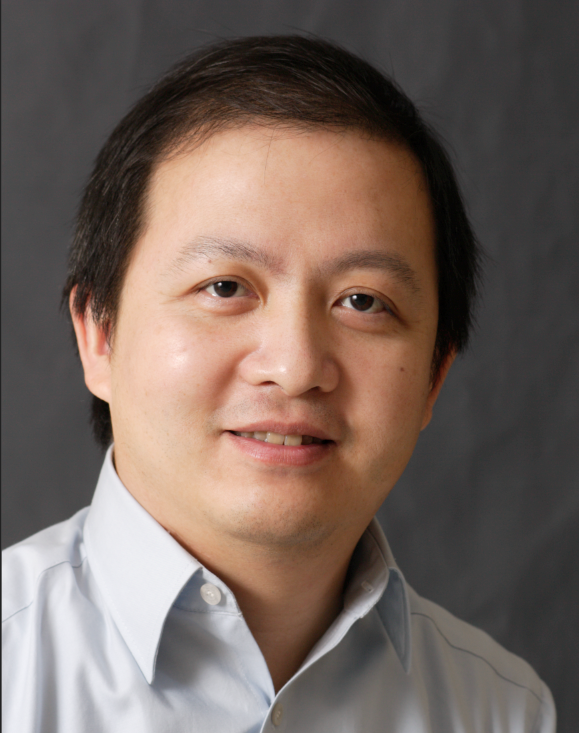}}]
{James Z. Wang} (Senior Member, IEEE)
is a Distinguished Professor of the Data Science and Artificial Intelligence area and the Human-Computer Interaction area of the College of Information Sciences and Technology at The Pennsylvania State University. He received a bachelor's degree in mathematics {\it summa cum laude} from the University of Minnesota (1994), and an MS degree in mathematics (1997), an MS degree in computer science (1997), and a PhD in medical information sciences (2000), all from Stanford University. His research interests include affective computing, image analysis, image modeling, image retrieval, and their applications. He was a visiting professor at the Robotics Institute at Carnegie Mellon University (2007-2008), a lead special section guest editor of the IEEE Transactions on Pattern Analysis and Machine Intelligence (2008), and a program manager at the Office of the Director of the National Science Foundation (2011-2012).
\end{IEEEbiography}

\begin{IEEEbiography}[{\includegraphics[width=1in,height=1.25in,clip,keepaspectratio,trim=40 60 30 0]{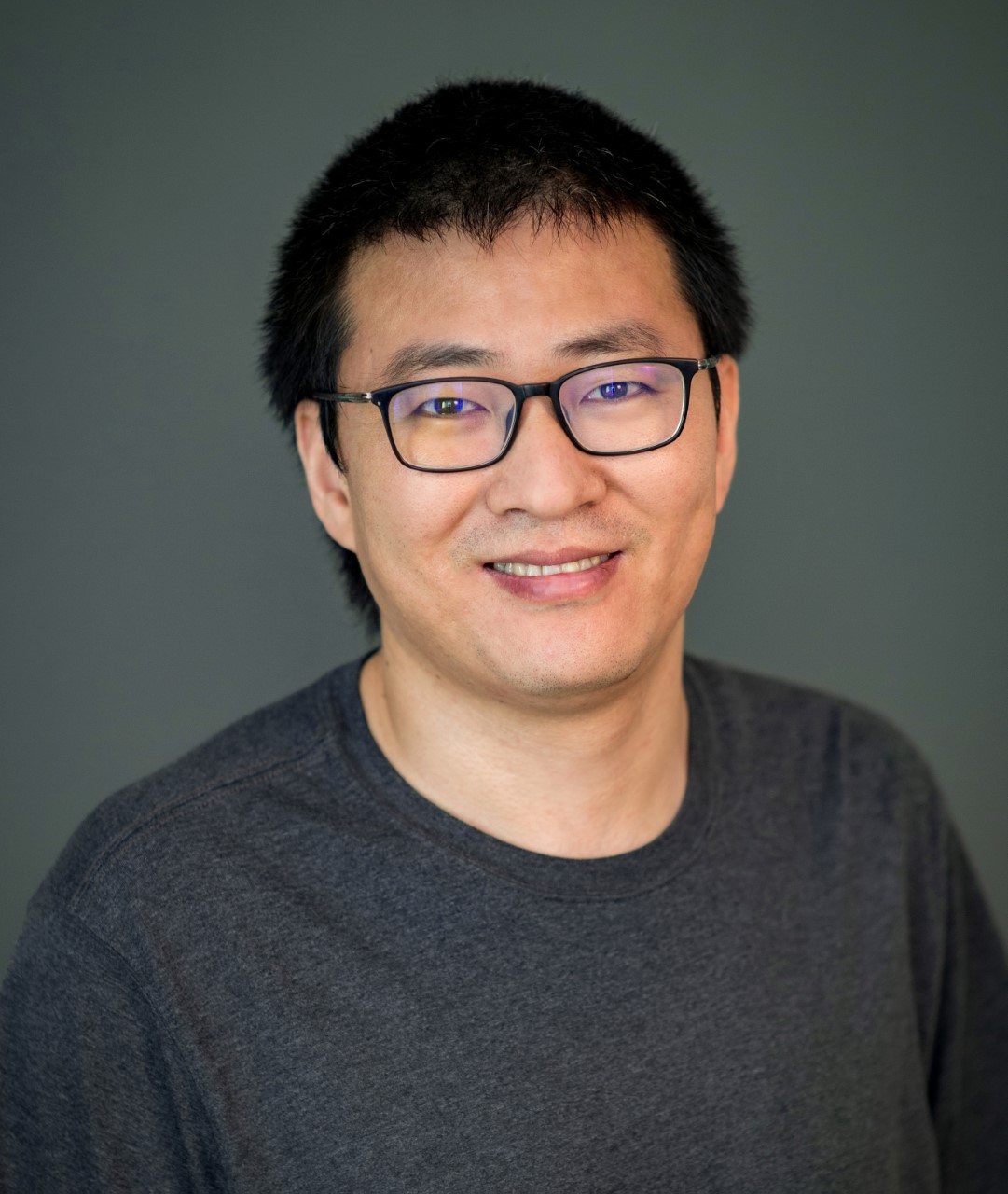}}]
{Sicheng Zhao}
(Senior Member, IEEE) is currently a Research Associate Professor at Tsinghua University. He received a PhD from Harbin Institute of Technology in 2016. He was a Visiting Scholar at National University of Singapore from 2013 to 2014, a Research Fellow at Tsinghua University from 2016 to 2017, a Postdoc Research Fellow at University of California, Berkeley from 2017 to 2020, and a Postdoc Research Scientist at Columbia University from 2020 to 2022. His research interests include affective computing, multimedia, and computer vision.
\end{IEEEbiography}

\begin{IEEEbiography}[{\includegraphics[width=1in,height=1.25in,trim={0 0 20 10},clip,keepaspectratio]{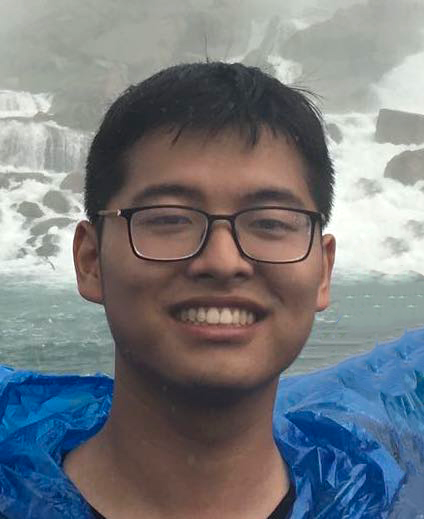}}]{Chenyan Wu} is a doctoral candidate in the Informatics program of the College of Information Sciences and Technology at The Pennsylvania State University. He earned a B.E. degree in Electronic Information Engineering from the School of the Gifted Young, University of Science and Technology of China in 2018. He has worked as an intern at Amazon Lab126 (Bellevue, WA), Microsoft Research Asia (Beijing, China), and SenseTime Research (Shenzhen, China). His research interests are affective computing, computer vision, and machine learning.
\end{IEEEbiography}

\begin{IEEEbiography}[{\includegraphics[width=1.2in,height=1.3in, trim={0 20 40 0}, clip,keepaspectratio]{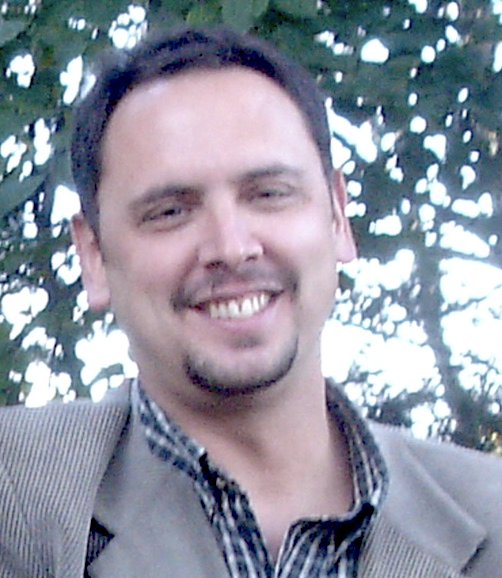}}]{Reginald B. Adams, Jr.} is a Professor of Psychology at The Pennsylvania State University. He received his PhD in social psychology from Dartmouth College in 2002. He is interested in how we extract social and emotional meaning from nonverbal cues, particularly via the face. His work addresses how multiple social messages ({\it e.g.}, emotion, gender, race, age, etc.) combine across multiple modalities and interact to form unified representations that guide our impressions of and responses to others. Although his questions are social psychological in origin, his research draws upon visual cognition and affective neuroscience to address social perception at the functional and neuroanatomical levels. Before joining Penn State, he was awarded a National Research Service Award (NRSA) from the National Institute of Mental Health to train as a postdoctoral fellow at Harvard and Tufts Universities. His continuing research efforts have been funded through NSF and the National Institute on Aging and the National Institute of Mental Health of the National Institutes of Health.
\end{IEEEbiography}

\begin{IEEEbiography}[{\includegraphics[width=1in,height=1.25in,trim={0 80 0 0},clip,keepaspectratio]{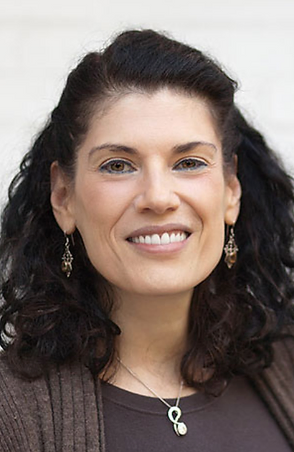}}]{Michelle G. Newman} is a Professor of Psychology and Psychiatry, and Director of the Center for the Treatment of Anxiety and Depression at the Pennsylvania State University. She received her PhD in clinical psychology from the University of Stony Brook in 1992 and completed a postdoctoral Fellowship at Stanford University in 1994. Dr. Newman has conducted basic and applied research on anxiety disorders and depression and has published over 200 papers on these topics. She is the past editor of Behavior Therapy and is currently Associate Editor of Journal of Anxiety Disorders. She is also the recipient of the APA Division 12 Turner Award for distinguished contribution to clinical research, APA Division 29 Award for Distinguished Publication of Psychotherapy Research, ABCT Outstanding Service Award, APA Division 12 Toy Caldwell-Colbert Award for Distinguished Educator in Clinical Psychology, and Raymond Lombra Award for Distinction in the Social or Life Sciences.  She is also a Fellow of American Psychological Association Divisions 29 and 12, the Association for Behavioral and Cognitive Therapies, and American Psychological Society.
\end{IEEEbiography}

\newpage

\begin{IEEEbiography}[{\includegraphics[width=1in,height=1.25in,clip,keepaspectratio]{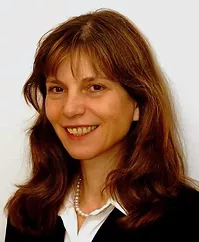}}]{Tal Shafir} graduated from law school at the Hebrew University of Jerusalem.  
She then studied dance-movement therapy at the University of Haifa and completed her PhD in neurophysiology of motor control
and two postdoctoral fellowships: in brain-behavior interactions in infants, and in affective neuroscience, all at the University of Michigan. 
She developed research on movement-emotion interaction and its underlying brain mechanisms, behavioral expressions, and therapeutic applications.
Shafir, certified also in Laban Movement Analysis, was the main editor of The Academic Journal of Creative Arts Therapies, and Frontiers in Psychology research topic: `The state of the art in creative arts therapies.' She has been serving on The American Dance Therapy Association (ADTA) research committee since 2016 and was the recipient of the ADTA 2020 Innovation Award. 
\end{IEEEbiography}

\begin{IEEEbiography}[{\includegraphics[width=1in,height=1.25in,trim={40 0 40 0},clip,keepaspectratio]{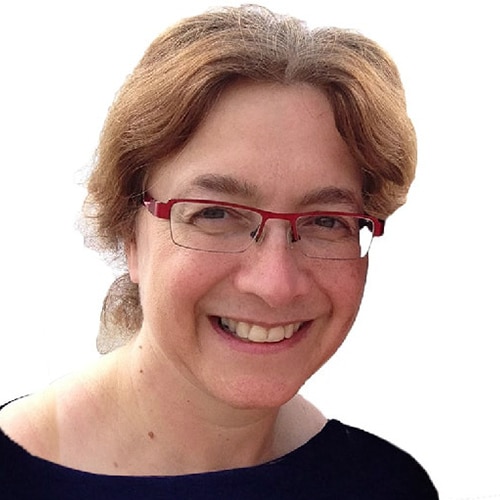}}]{Rachelle Tsachor} is an Associate Professor of Movement at the University of Illinois at Chicago. She is certified in Mind-Body Medicine (CMBM), Movement Analysis (CMA), and Somatic Movement Therapy (RSMT-ISMETA). Her research investigates body movement to bring a human, experiential understanding to how movement affects our lives. Tsachor analyzes patterns in moving bodies in diverse projects, researching movement's effects on our brains, emotions, health, and learning. She is co-PI on an NSF-Funded project STAGE and the UI Presidential Initiative for the Young People's Science Theater: CPS and UIC Students Creating Performances for Social Change. Both initiatives bring Mind/Body methods into Chicago Public Schools' that educate primarily students of color to support learning in embodied ways, so students view themselves as potential contributors of creative solutions.
\end{IEEEbiography}

\vfill
\newpage

\end{document}